%% file: ACCA_Paper1_v60.tex
\documentclass[11pt]{article}
\usepackage[T1]{fontenc}
\usepackage{lmodern}
\usepackage[margin=1in]{geometry}
\usepackage{amsmath,amssymb,booktabs,tabularx,array,microtype,hyperref,xcolor}
\usepackage{enumitem}
\usepackage{needspace}
\usepackage{ragged2e}
\usepackage{graphicx}
\input{preamble_add.tex}
\input{fig_refs.tex}
\usepackage[numbers,sort&compress]{natbib}
\hypersetup{
  colorlinks=true,
  linkcolor=blue,
  citecolor=blue,
  urlcolor=blue,
  pdftitle={Assurance-Conditioned Continuous Authorization for Autonomous AI Agents},
  pdfauthor={Mehdi Shajari}
}
\newcommand{\framework}{\textsc{ACCA}}
\newcommand{\dgp}{\textsc{DGP}}
\title{Assurance-Conditioned Continuous Authorization for Autonomous AI Agents}
\date{}

\begin{document}

\author{Mehdi Shajari\\\textit{Independent Researcher, Canada}\\\texttt{mehdi.shajari@totaltrust.ai}}

\maketitle

\begin{abstract}
Autonomous software agents can act through tools, delegate work, and change their models
or execution environments while earlier authorizations remain in flight. We present
Assurance-Conditioned Continuous Authorization (ACCA), an approach designed to be harness-independent,
for maintaining operational authority according to currently justified assurance. It
distinguishes evidence validity from contextual applicability, derives assurance claims
from eligible evidence, and resolves a multidimensional effective-authority envelope
within a principal-declared ceiling and applicable policy. A typed dependency graph
supports selective re-evaluation of authorization-relevant changes. An authority epoch
changes when effective authority changes; a distinct optional authorization-basis version supports
revalidation of action-specific authorization conditions. Protected effect-time enforcement
can reject stale actions before a consequential resource change, subject to the available
commit mechanism. A compositional Domain Governance Profile expresses domain-specific
semantics without granting authority. Controlled scenario tests implemented as a regression suite using a trading profile
exercise default-deny rules, evidence-conditioned contraction and recovery, policy
divergence, selective non-response, and stale-action rejection at a local gateway before the simulated-broker boundary.
The study establishes implementation-level evidence in one domain, not a cross-domain
guarantee of harmful-behavior prevention.
\end{abstract}

\section{Introduction}

Agentic AI changes the authorization problem in ways that go beyond tool
invocation. A conventional application typically executes a relatively stable
set of operations through predefined workflows and interfaces. An autonomous
or semi-autonomous agent can instead select among actions, formulate
intermediate plans, adapt its behavior as observations arrive, delegate work,
interact with external systems, and continue acting as its environment
changes. Consequently, authorization conditions that were appropriate when a
task began may cease to be appropriate during execution. The agent itself may
change---for example, through a model, prompt, tool, configuration, persistent
memory, learned skill, routing, subagent, or harness update---but authority may
also become inappropriate without any change to the agent. A dependency may degrade, an external environment may enter conditions
outside those under which the agent was evaluated, a security control may
fail, or a stakeholder's risk posture may change. Static permission
assignments are therefore poorly matched to autonomous systems whose
operational context, dependencies, risk, and assurance basis can change while
they act.

This problem is not merely technical. Traditional organizational and service
workflows routinely delegate substantial authority to trusted insiders, but
that authority is embedded within a broader institutional control system.
Human discretion is constrained by professional judgment, procedural controls,
separation of duties, approval requirements, supervision, organizational
accountability, performance incentives, professional obligations, and the
prospect of financial, employment, legal, or reputational consequences for
misconduct. Together, these mechanisms create a balance between granting
individuals sufficient discretion to perform useful work and constraining the
risks created by that discretion. Many of these safeguards operate beyond
technical access control: an employee may be technically capable of performing
an action and formally authorized to do so, yet refrain because the action is
inappropriate under the circumstances, requires escalation or additional
approval, violates a professional obligation, or could create unacceptable
consequences.

As service activities are increasingly delegated to autonomous agents, this
institutional control model does not transfer directly to software. Some
controls can be translated into machine-enforceable mechanisms---for example,
separation of duties can become independent approval, procedural controls can
be expressed as policy, and supervisory intervention can become
human-in-the-loop authorization. Other safeguards do not have straightforward
equivalents. Agents do not inherently possess professional responsibility,
internalize organizational norms, respond to career or reputational
incentives, or face penalties in the manner of human decision makers. Nor can
organizations assume that an agent will exercise the contextual and prudential
judgment through which trusted insiders recognize exceptional circumstances
and voluntarily limit their own actions. Agentic automation therefore does not
merely replace human labor; it changes the institutional mechanisms through
which delegated authority has traditionally been made tolerable.

This distinction exposes a fundamental governance problem. Technical
capability and access permission are insufficient bases for delegation. An
agent may be capable of performing an action and possess legitimate access to
the required resources while the available assurance evidence, current
operating conditions, or applicable risk and governance policy no longer justify allowing
it to exercise that authority. The relevant question is therefore not merely
whether an agent \emph{can} perform an action, or even whether an identity has
permission to invoke an interface, but what the agent should currently be
\emph{allowed} to do, under what constraints, and with what degree of
autonomy. The distinction between \emph{capability} and \emph{justified
authority} consequently becomes a central governance concern.

The economic motivation for solving this problem is substantial. Agentic
software may automate not only information-technology tasks but significant
portions of professional and operational services, moving software from
supporting workflows toward performing services themselves. This
``service-as-software'' trajectory increases the potential value of autonomy
while simultaneously increasing the consequences of inappropriate authority.
The objective is therefore not to eliminate autonomy, but to enable useful
autonomy while continuously ensuring that the authority exercised by an agent
remains justified by current evidence, operating conditions, organizational
policy, and stakeholder risk tolerance.

This implies a bidirectional authority lifecycle. Deteriorating or non-applicable
assurance may justify contraction, while newly established or restored assurance
may justify expansion or recovery within the principal-declared authority ceiling. Authority is therefore neither a static entitlement nor a monotonically
increasing measure of agent maturity; it is a reversible governance state whose
current value should remain justified by current evidence and policy.

This problem becomes sharper for agents that learn or self-modify during
deployment. Continual Harness permits an agent to refine its own prompt,
subagents, skills, and memory online; Harness Continual Learning (HCL) formalizes
continual adaptation of model-external harness state and identifies
harness-level forgetting; HarnessDev evaluates iterative harness creation and
evolution; EvolveNet studies collaborative composition of harness adaptations
across deployments; and Harness-Zero studies a complementary transition in which
behavior induced by an optimized training-time harness is distilled into model
parameters and retained under a fixed target harness after the specialized
harness is removed
\cite{karten2026continualharness,kang2026hcl,wu2026harnessdev,nie2026evolvenet,ye2026harnesszero}.
These systems illustrate a form of \emph{dynamic autonomy}: the system's future
capability and execution behavior can change from experience, including by moving
capability across the model--harness boundary, even when its identity and standing
credentials do not. The governance question is therefore
not only whether a changed system can still perform prior tasks, but whether the
operational authority justified for the pre-change system should survive the
learning transition. \framework{} treats a learning-induced system transition as
a potentially authorization-relevant state transition, not as an independent source of authority. When such a transition is authorization-relevant, prior effective
authority is not presumed to survive from actor identity alone; its
continuation depends on the applicable post-change assurance and
authorization basis.

Throughout this paper, $d$ denotes an application domain, $u$ an authorizing
principal, and $g$ a governed actor. We write $\mathsf{Act}_d$ for the action
vocabulary governed in domain $d$, $A_{d,u,g}$ for the principal-declared
authority ceiling for actor $g$, and $A_{\mathrm{eff},u,g}^{(t)}$ for the
currently justified effective authority at time $t$. The structured,
multidimensional representation of $A_{\mathrm{eff},u,g}^{(t)}$ is referred to
as the effective-authority envelope. These are different kinds of objects:
$\mathsf{Act}_d$ is a schema-level vocabulary, whereas $A_{d,u,g}$ is a
deployment-specific governance decision expressed over that vocabulary. We define
$\preceq_d$ as the domain-specific authority-containment relation: for two
authority states or ceilings $A$ and $B$, $A\preceq_d B$ means that $A$ is no
more permissive than $B$ in any governed dimension, which implies
\[
\{q : q\models A\}\subseteq\{q : q\models B\}.
\]
Here $q$ ranges over well-formed consequential action requests and
$q\models A$ means that the request satisfies every applicable dimension of
authority state or ceiling $A$.
On canonical authority states (or equivalence classes of states with identical
authorization semantics), $\preceq_d$ is a partial order, not necessarily a total
order: $A\parallel_d B$ denotes incomparable authority states. A contraction
or expansion corresponds to a strict comparison when the states are comparable;
an incomparable change is a lateral reconfiguration that may narrow some
dimensions while broadening others. The strict relations $\prec_d$ and
$\succ_d$ denote contraction and expansion under this order. For any envelope $A$, $\operatorname{Allow}(A)$ denotes its
explicitly permitted-action projection, a subset of $\mathsf{Act}_d$; actions
outside that projection are denied under the closed-world semantics used here when they are governed consequential actions. Informational actions are not automatically authorized: their governance and enforcement requirements depend on whether they produce relevant effects, including disclosure or resource consumption.
Formal definitions and invariants retain the $u$ and $g$ indices rather than
using partially indexed authority symbols.

Dynamic authorization itself is not new. Usage Control (UCON) introduced
continuity of authorization, mutable attributes, obligations, and ongoing
policy decisions \cite{park2004ucon}. XACML formalized policy administration,
decision, and enforcement roles \cite{oasis2013xacml}; ABAC and
risk/context-aware access-control approaches established context-sensitive
authorization \cite{nist2014abac}; and Zero Trust architectures normalized
continuous posture-informed access decisions \cite{nist2020zta}.
These mechanisms establish important foundations for
the present work but are not themselves contributions of this paper.

Recent agent-specific research extends these foundations to delegated identity,
least-privilege tool access, runtime policy enforcement, contextual and
semantic authorization, behavioral governance, and authorization profiles or
envelopes
\cite{south2025delegation,progent2025,agentspec2026,conseca2025,
ietfagentauth2026,aae2026,casa2026,agentbound2026,calboreanu2026lattice,forge2026}.
These approaches substantially advance the control of autonomous agents.
However, a gap remains between \emph{assurance} and \emph{authority}. Existing
mechanisms generally derive authorization directly from identity, delegation,
attributes, observed context, task semantics, policy constraints, or
action-level proofs. They do not explicitly maintain the \emph{assurance
provenance} of currently justified operational authority: which evidence supports the
claims that justify a particular level of authority; which assumptions and
dependencies determine the continued applicability of that evidence; how
authorization-relevant changes affect those dependencies and claims; and how the
resulting assurance state should alter the effective authority allocated to an agent under
an applicable risk and governance policy.

We propose \framework{} (Assurance-Conditioned Continuous Authorization), a
harness-independent-by-design governance and authorization control plane for continuously
maintaining the correspondence between an agent's effective operational authority
and the conditions that justify that authority. Through a compositional Domain
Governance Profile (\dgp{}), it combines core cross-domain governance semantics with domain-specific semantics while interoperating with heterogeneous
identity, telemetry, attestation, evaluation, behavioral-monitoring, policy, and
enforcement systems. Rather than treating authorization as a static entitlement,
\framework{} treats authority as a continuously maintained governance state and
maintains or references the decision and authority lineage needed to explain how
that state and resulting consequential-action decisions were derived and enforced.

\framework{} is designed as a domain-independent framework for governing the authority
of consequential agent actions. A validated \dgp{} supplies the semantics of a
domain's requirements, whether security, safety, privacy, financial, organizational,
social, or ethical. The profile grants no authority: the principal-declared ceiling
and applicable policy remain separate inputs. A requirement that can be expressed as
an operating constraint, assurance requirement, obligation, or human-approval
requirement can shape effective authority, which never exceeds the ceiling. A
requirement that cannot be reliably observed or enforced does not become a guaranteed
protection by being encoded in a profile; many social and ethical requirements fall
into this category. This paper evaluates a single trading-domain instantiation. It
does not test other domains, and it does not show that \framework{} prevents social
or ethical harms.

To preview the lifecycle, let $X_t$ denote the
authority-relevant system and operating state, $E_t^*$ the evidence currently
valid and applicable to that state, $C_t$ the assurance claims currently
supported by that evidence, and $P_u$ the applicable risk and governance policy for principal $u$,
which may combine authenticated stakeholder, organizational, governed-resource-owner,
regulatory, or other authorized policy inputs. Here $A_{d,u,g}$ supplies the
principal-declared authority ceiling, and $\mathrm{Epoch}_t$ is a monotonically
ordered version value associated with the current effective-authority state. We use $\mathcal{B}^{\mathrm{auth}}_t$ for the logical current
\emph{authorization basis}: the current set or versioned representation of
inputs and intermediate governance state that can affect either derivation of
effective authority or authorization of a particular consequential action. It
can include $A_{d,u,g}$, $X_t$, $E_t^*$, $C_t$, $P_u$, profile state,
approvals, obligations, target or resource state, execution-path state, and
other configured authorization inputs. It need not be stored as one monolithic
object. This notation is distinct from the behavioral-state component $B_t^d$
introduced later as part of $X_t^d$.

For a reauthorization event occurring at time $t$, let $t^{-}$ and
$t^{+}$ denote the instants immediately before and after the transition,
respectively. Throughout Eq.~\ref{eq:canonical-lifecycle}, $\Delta$ denotes a detected
or derived change in the indexed quantity over the transition at time $t$,
rather than numerical subtraction. In particular, $\Delta E_t^*$ may include evidence entering or
leaving the applicable evidence basis, and $\Delta C_t$ may include the
establishment, withdrawal, or changed support status of an assurance claim.
For fixed $P_u$ and $A_{d,u,g}$, one common evidence-mediated
propagation path is

\begin{equation}
\boxed{
\begin{aligned}
\Delta X_t
&\leadsto
\Delta E_t^*
\leadsto
\Delta C_t
\\
&\xrightarrow[\text{bounded by }A_{d,u,g}]
              {\text{policy }P_u}
\Delta A_{\mathrm{eff},u,g}^{(t)}
\Longrightarrow
\mathrm{Epoch}_{t^{+}}
:=
\mathrm{Epoch}_{t^{-}}+1
\end{aligned}}
\label{eq:canonical-lifecycle}
\end{equation}
Figure~\ref{fig:lifecycle} summarizes the assurance-to-authority derivation and protected effect-time authorization path.\par
\input{fig1_lifecycle.tex}

Here $\leadsto$ denotes possible propagation, which stops when reevaluation
leaves the next downstream quantity unchanged. We call a propagation path
\emph{evidence-mediated} when an authorization-relevant change affects
authorization through a change in evidence validity or applicability and the
resulting assurance-claim support. Evidence mediation characterizes the
propagation path rather than an exclusive type of source event: the same
underlying change can also affect a directly evaluated policy predicate or other
authorization-basis state. This branch is non-exhaustive:
direct evaluation of $X_t$, or an authorized change to $P_u$ or $A_{d,u,g}$,
may trigger rederivation without changing $E_t^*$ or $C_t$.

The final implication holds only if
$A_{\mathrm{eff},u,g}^{(t^{+})}
\neq A_{\mathrm{eff},u,g}^{(t^{-})}$.
In that case, the $+1$ advances only the authority epoch, which is the version
counter of the effective-authority state; it does not represent an increase in authority or an elapsed unit of time. If
rederivation leaves effective authority unchanged, then
$\mathrm{Epoch}_{t^{+}}=\mathrm{Epoch}_{t^{-}}$. Where convenient, we abbreviate
$A_{\mathrm{eff},u,g}^{(t^{+})}\neq A_{\mathrm{eff},u,g}^{(t^{-})}$ as
$\Delta A_{\mathrm{eff}}\neq 0$; this denotes a changed envelope, not numerical
subtraction.

An artifact $e$ contributes to $E_t^*$ exactly when
$\mathrm{Valid}(e,t)\land\mathrm{Applicable}(e,X_t)$ holds, separating
evidence validity from contextual applicability. Authorization relevance is not equated with observation of a change or
graph reachability alone. A change is authorization-relevant when its semantic
effect can alter evidence eligibility, assurance-claim support, satisfaction
of an applicable-policy predicate, an action-specific authorization condition
or the continued validity of a previously issued action authorization, a
configured adaptation-activation condition, or one or more dimensions of
effective authority. A change is authority-material only when semantic
reevaluation actually changes the effective-authority envelope. One or more policy
enforcement points (PEPs) then consume the resulting authority state and epoch
when mediating consequential actions. A PEP may be hosted in a harness or tool
gateway, a queue worker, a resource adapter, or the governed system itself;
strict stale-effect prevention depends on the PEP that protects the relevant
effect boundary, not on the name of the component that hosts it.

Here $\Delta X_t$ may originate in the agent, model, harness, tools,
configuration, dependencies, operating environment, stakeholder context, or
assurance-relevant observed behavior. Along an evidence-mediated path, the
framework identifies evidence whose assumptions or applicability conditions
may be affected, reevaluates only the relevant evidence and claims, and
applies applicable policy to determine whether one or more dimensions of
effective authority change. A context change that affects only operational
feasibility need not change authority. Conversely, a context or harness change
can be authorization-relevant when it invalidates evidence supporting current
authority or changes a directly evaluated policy predicate.

This formulation permits authority to change even when the agent itself is
unchanged. For example, a trading agent may have been evaluated and authorized
for autonomous execution under specified assumptions about market volatility,
liquidity, data quality, and execution conditions. A severe market disruption
may invalidate those assumptions without changing the model, strategy, tools,
or credentials of the agent. \framework{} can treat the environmental change
as authorization-relevant, determine which assurance evidence and claims depended on the
violated assumptions, and contract the agent's authority---for example, from
autonomous live execution to reduced position limits, mandatory human
approval, or recommendation-only operation. Conversely, newly established
assurance evidence may justify controlled expansion of authority.

The temporal separation between authorization and effect makes this distinction
operational rather than terminological. For example, an order may be authorized
under epoch 41 and placed in a queue; before a worker submits it, newly applicable
evidence may contract the agent's trading limit and advance the current epoch to
42. A worker that treats the queued item as already approved can still execute an
obsolete authorization. An epoch-aware enforcement point instead rejects the
epoch-41 action at the consequential boundary and, if reconsideration is allowed,
requires complete authorization under epoch 42. No policy was edited in this example. A hypothetical authorization mechanism whose revocation triggers are limited to policy-definition or static-permission changes would not respond to the assurance-dependent transition in this example and could continue to accept the epoch-41 order; this is not a characterization of capable event-driven or continuous-authorization systems. The \framework{} authority epoch instead advances because the authority derived under the unchanged policy has contracted.

The paper makes six contributions, building on established foundations in continuous
usage control, access control, and runtime enforcement:
\begin{enumerate}[leftmargin=*]
\item \textbf{Assurance-conditioned authority.} The model explicitly separates evidence validity from
contextual applicability, derives assurance claims from currently eligible evidence, and resolves
effective authority under applicable policy within a principal-declared ceiling. Direct policy and
state dependencies are also supported.
\item \textbf{Selective change propagation.} Declared dependencies identify potentially affected authorization paths, while semantic reevaluation distinguishes authorization-relevant changes from the narrower class of authority-material changes that alter the effective-authority envelope. These changes may originate in models, harnesses, operating environments, or observed behavior.
\item \textbf{Multidimensional governed authority.} An inspectable assurance-to-policy derivation
supports independently constrained permissions, resources, limits, autonomy, delegation, approvals,
environments, and obligations; capability and operational feasibility remain distinct from authority.
\item \textbf{Versioned consequential authorization.} Authority epochs advance when effective authority
changes; optional authorization-basis versions require action-specific revalidation without falsely
implying an authority-envelope change. The evaluated local gateway rejects epoch-stale actions before the simulated-broker boundary; stronger protected effect-time guarantees depend on separately specified commit and fencing contracts and were not evaluated.
\item \textbf{Compositional domain governance.} A validated Domain Governance Profile composes core and domain-specific governance semantics without granting authority. Applicable policy and principal-declared
ceilings remain separate inputs.
\item \textbf{Prototype and evaluation.} A trading-domain prototype tests closed-world authority,
evidence-conditioned contraction and recovery, selective non-response, policy divergence, and
stale-action rejection. The results support this implementation, not a general cross-domain guarantee.
\end{enumerate}

Conceptually, \framework{} extends the continuous-control lineage represented
by UCON. Whereas continuous usage control asks whether an ongoing use of a
resource remains authorized as relevant attributes and conditions change,
\framework{} asks a broader question for autonomous systems: \emph{given the
evidence and conditions that hold now, how much operational agency remains
justified?}

\section{Background and Previous Work}

\subsection{Continuous Usage Control}

UCON established that authorization need not be a one-time decision made
before access. Its continuity and attribute-mutability concepts permit
decisions to be reevaluated while usage is in progress
\cite{park2004ucon}. U-XACML and subsequent UCON implementations
operationalized these concepts in policy engines and distributed systems
\cite{colombo2010uxacml,lazouski2016cloud}. UCON+ further broadened
monitoring, policy administration, and delegation \cite{uconplus2023}.

Our earlier UCON architecture for cloud environments is an architectural precedent for continuous, context-aware usage control \cite{tavizi2012uconcloud}. \framework{} retains this continuous-control principle but changes the governed object and derivation structure: rather than asking only whether an ongoing resource usage may continue, it asks what degree of operational agency remains justified by the currently applicable assurance basis and policy.

\subsection{Context-, Risk-, and Trust-Adaptive Authorization}

ABAC, context-aware access control, risk-adaptive access control, and Zero
Trust demonstrate that changing context and risk can legitimately change
authorization decisions \cite{nist2014abac,nist2020zta}. Accordingly, neither
context-sensitive policy nor dynamic reevaluation is claimed as novel here.
\framework{} instead adds an explicit assurance layer between
authority-relevant state and operational authority: a state change can alter
whether evidence is applicable, a change in applicability can alter which
assurance claims are supported, and policy maps those claims to justified authority.

\subsection{Uncertainty, Distribution Shift, and Evidence Applicability}
\label{sec:uncertainty}

Improving model capability does not eliminate uncertainty at the level at
which consequential authorization decisions must be made. Aggregate benchmark
performance does not establish uniform reliability for every input, context,
or operating condition. Work on personalized uncertainty quantification
emphasizes the distinction between population-level performance and
instance-specific reliability \cite{chakraborti2025puq}.

For autonomous agents, uncertainty may propagate from interpretation and
reasoning through planning, tool selection, and consequential action. The
governance question is therefore not simply whether a model is ``confident,''
but whether the evidence supporting a given degree of operational authority
remains applicable to the current situation. Uncertainty quantification,
conformal methods, drift and out-of-distribution signals, behavioral monitors,
and evaluation results can provide assurance evidence, but should not
themselves be treated as authorization decisions.

Let $X_t$ denote the authority-relevant system and operating context at time
$t$, including relevant agent, model, dependency, environmental, behavioral,
and stakeholder state. Let an evidence artifact $e$ have a time-indexed
validity predicate $\mathrm{Valid}(e,t)$ and an applicability predicate
$\mathrm{Applicable}(e,X_t)\equiv\phi_e(X_t)$, where $\phi_e(X_t)$ indicates
whether the current context satisfies the conditions and assumptions under
which $e$ can support an assurance claim. We write this evidence-eligibility condition as
\begin{equation}
\chi_e(t)=\mathrm{Valid}(e,t)\land\mathrm{Applicable}(e,X_t).
\end{equation}

The predicate $\chi_e(t)$ separates whether an evidence artifact remains valid from whether it remains applicable to the agent's current operating conditions. A result obtained from a prior agent or model evaluation, for example, need not become invalid merely because the operating environment changes. Instead, a change in $X_t$ may cause one or more assumptions underlying that prior evaluation to cease to hold, making the resulting evidence non-applicable to the current context even though the evidence itself remains valid. Such a change can propagate through the assurance and authorization process as 
\begin{equation}
\Delta X_t
\leadsto
\Delta\phi_e(X_t)
\leadsto
\Delta E_t^*
\leadsto
\Delta C_t
\xrightarrow{\,P_u\,}
\Delta A_{\mathrm{eff},u,g}^{(t)},
\end{equation}
where $C_t$ denotes the assurance claims supported under the current evidence
state and $P_u$ denotes the applicable risk and governance policy.

Consequently, an unchanged agent may receive a different effective authority
when its operating context moves outside the conditions supported by its
assurance evidence. This does not require concluding that the agent or model
is incorrect or unsafe in general. It is sufficient to determine that the
currently applicable evidence no longer justifies the previous level of
consequential autonomy. \framework{} can therefore respond conservatively to
uncertainty arising from changes in operating conditions without treating
uncertainty itself as proof of failure.

\subsection{Agent Identity, Delegation, and Least Privilege}

Recent work treats AI agents as first-class delegated actors. Authenticated
Delegation and Authorized AI Agents extends identity and delegation mechanisms
for agent-mediated action \cite{south2025delegation}. IsolateGPT uses
execution isolation and permissions for LLM applications
\cite{isolategpt2025}. Progent provides programmable fine-grained privilege
control for tool-calling agents \cite{progent2025}; AgentSpec provides
customizable runtime enforcement \cite{agentspec2026}; MiniScope reconstructs
least-privilege permission scopes \cite{miniscope2025}; and Conseca generates
contextual policies for agent tasks \cite{conseca2025}. These works establish
strong mechanisms for restricting agent actions, but authority is principally
derived from permission policy, task context, or user intent rather than from
the continuing applicability of assurance evidence.

Risk-adaptive work moves closer to our concern by incorporating uncertainty
and operational risk into agent authorization \cite{riskadaptive2025}. This
establishes that model uncertainty may be an authorization input.
\framework{} generalizes the relationship by representing the evidence and
assumptions supporting an assurance claim and propagating authorization-relevant
changes through those dependencies before policy derives effective authority.

CapScope demonstrates a complementary harness-level approach: it derives a
task-wide authority ceiling from trusted input, assigns typed capabilities to
individual agents outside model context, and checks each tool call against the
issuing agent's capabilities \cite{bouras2026capscope}. This establishes scoped,
externally maintained capability enforcement and sub-agent separation. It does
not by itself determine how changes in the applicability of assurance evidence
should alter the authority represented by those capabilities.

\subsection{Emerging Agent Authorization and Runtime Governance}

Protocol and infrastructure mechanisms supply important complementary controls.
The Model Context Protocol (MCP) authorization specification describes
OAuth-based transport-level authorization for restricted HTTP MCP servers
\cite{mcpAuthorization2026}; it does not assess whether the assurance supporting
an agent's authority remains contextually applicable. The Continuous Access
Evaluation Profile (CAEP) supplies interoperable security-event signals that
cooperating services can use to attenuate access~\cite{caep2023}.
Zanzibar provides a scalable authorization model with causal consistency across
access-control updates and object contents~\cite{pang2020zanzibar}.
These are important precedents for authorization, signaling and consistency,
not mechanisms ACCA claims to replace.

NIST's 2026 concept paper explicitly identifies dynamic agent identity and
authorization as an emerging standards problem \cite{nistagents2026}. The
IETF AI Agent Authentication and Authorization draft addresses agent identity,
delegated context, observability, remediation, and dynamic privilege
attenuation \cite{ietfagentauth2026}. The Agent Authorization Envelope (AAE)
defines a machine-evaluable representation of an agent mandate and constraints
\cite{aae2026}, while the World Economic Forum's ACAP work separates
capability and authorization at deployment level \cite{wefacap2026}. We
therefore do not claim the concept of an authorization envelope itself as
novel.

The Agent Action Decision Protocol (AADP) further separates standing capability
from whether a particular action, with particular argument values, may proceed
under mutable budgets, reservations, approval state, obligations, and kill
switches \cite{saha2026aadp}. AADP is an individual Internet-Draft and is cited
as work in progress rather than as an IETF standard. Its PDP--PEP protocol is a
natural consumer of an authority state maintained by \framework{}; its governed
object is the individual action decision rather than the assurance basis from
which standing operational authority is derived.

Independent predictive guardrails provide a complementary source of runtime
assurance rather than an authority-state mechanism. LawZero's Scientist AI
\cite{bengio2026scientistai} proposes a non-agentic Predictor trained toward consequence-invariant probabilistic
prediction and deployable as a guardrail around agentic scaffolding; we return to
this monitor-to-authority composition in
Section~\ref{sec:behavioral-monitoring}.

Other recent work addresses Continuous Agent Semantic Authorization (CASA)
\cite{casa2026}, verifiable behavioral governance (AgentBound)
\cite{agentbound2026}, governance-first runtime architecture (LATTICE)
\cite{calboreanu2026lattice}, Datalog-based formal policy enforcement with a runtime reference monitor
(FORGE; named PCAS in the first arXiv version) \cite{forge2026}, capability-versus-allowed-autonomy distinctions
\cite{aclaal2026}, and cryptographic or proof-derived authorization
\cite{cva2026,dtf2026}. Collectively, these works supply many of the building
blocks required for trustworthy agent authorization.

Recent work further sharpens the comparison boundary. OpenKedge's Taming Cognitive Transactions (TCT) uses typed dependency tokens, versioned executable state,
current dependency vectors, and commit-time guards to preserve derivation-relevant
state through consequential commit \cite{he2026tct}. Accentrust's root-scoped
quiescence work formalizes revocation for long-running and delegated agent work using
root epochs, local fences, and provider-frontier certificates \cite{zhu2026quiescence}.
These results independently demonstrate that versioned authorization state,
dependency-aware currentness, fencing, and late-effect suppression are not unique to
\framework{}. Emerging standards and products likewise provide strong surrounding
mechanisms: the IETF AIMS draft describes authorization-relevant change signals and
access attenuation or policy re-evaluation \cite{ietfaims2026}; OpenID AuthZEN
standardizes PDP--PEP authorization exchange and develops agent-era approval and
protocol-mapping profiles \cite{authzen2026,authzenaaarp2026}; AWS AgentCore Policy
uses Cedar or Dogwood for deterministic, per-request control of agent interactions at the gateway
\cite{awsagentcorepolicy2026}; and SPIFFE/SPIRE provides portable workload identity
and attestation primitives \cite{spiffe2026}. These mechanisms narrow the novelty
claim: the contribution is not policy epochs, dependency tokens, runtime signals,
workload identity, declarative policy evaluation, revocation, or effect-time fencing
in isolation, but the explicit derivation of a principal-bounded multidimensional
effective-authority state from validity- and applicability-qualified assurance
evidence and claims, joined to version-bound action authorization and protected
consequential-effect enforcement.

Several concurrent works further address the effect-time and evidence-currentness
halves of this problem. Commit-time authorization for LLM agents requires the witness
that licensed a durable effect to remain fresh, bound, and eligible at commit time
\cite{santosgrueiro2026commit}. PORTICO uses temporary, revocable resource-and-effect
capabilities whose authority is bounded by the task phase that justifies them
\cite{santosgrueiro2026portico}. AID-Guard revalidates an approved request and provider
state at commit while preserving effect lineage across retry and recovery
\cite{tong2026aidguard}. Proof-or-Stop admits lifecycle transitions only when fresh,
source-state-bound, mechanically verifiable evidence satisfies the relevant gate
\cite{huang2026proofstop}. FAVA checks proposed actions against an evidence-backed
permission graph using deterministic lowering and SMT authorization before effectful
execution \cite{zhang2026fava}. The Evidence Qualification Receipt Internet-Draft
binds a proposition to a versioned evidence-requirements profile and explicitly
separates evidence qualification from authorization \cite{watts2026eqr}. These works
reinforce that fresh evidence, current authority, and commit-time checks are not
unique to \framework{}. The distinct question addressed here is how the current
\emph{degree} of operational authority is derived from evidence whose validity and
contextual applicability are represented separately, through explicit assurance
claims and applicable policy, before that authority is supplied to an
action-level or commit-time enforcement mechanism.

Architectural-governance work provides a complementary governance-level
perspective on execution-bearing systems. ISDAIRE formalizes ex-ante action-class
specification and admissibility; ARETABA formalizes the construction of runtime
authority objects; and OTANIS and related mechanisms formalize execution-boundary,
composition, traceability, accountability, refusal, and escalation requirements
\cite{otani2026isdaire,otani2026aretaba,otani2026otanis,otani2026mgag,otani2026usg}.
The materials reviewed are primarily formal architectural frameworks and worked
examples rather than an empirical implementation campaign comparable to the ACCA
prototype evaluated here. They are nevertheless important neighboring work and make
clear that runtime authority objects, execution-boundary mediation, refusal,
escalation, provenance, traceability, accountability, and compositional authority
preservation are not individual novelties of \framework{}. ACCA's narrower focus is
the upstream maintenance of the assurance-conditioned \emph{degree} of operational
authority that an action-level or execution-governance mechanism may consume.

\subsection{Agentic Security Guidance and Secure Inter-Agent Interaction}
\label{sec:agentic-security-guidance}

Recent security guidance further narrows the contribution boundary. The OWASP
Top 10 for Agentic Applications 2026 identifies goal hijacking, tool misuse,
identity and privilege abuse, agentic supply-chain compromise, unexpected code
execution, memory and context poisoning, insecure inter-agent communication,
cascading failures, human-agent trust exploitation, and rogue-agent behavior
as major agentic-security risks \cite{owaspagentic2026}. Its recommended
mitigations include least privilege, per-action authorization, runtime policy
enforcement, short-lived or just-in-time privileges, provenance, behavioral
monitoring, adaptive adjustment of autonomy and human oversight, and
independent enforcement of consequential actions. These mechanisms are
important foundations and are not claimed as contributions of \framework{}.
In particular, \framework{} does not claim to have introduced per-action
authorization, adaptive autonomy, human escalation, runtime policy gates, or
behavior-triggered revocation. Its distinct contribution is to make the assurance
basis for current authority explicit, derive and maintain authority from the
evidence that remains applicable under current conditions, and expose that
authority and its provenance to heterogeneous enforcement and audit interfaces.

OWASP also identifies a closely related TOCTOU failure mode that it calls
\emph{Workflow Authorization Drift}: a procurement agent obtains authorization
for a purchase workflow, the user's spending limit is subsequently reduced,
yet the workflow proceeds using the earlier authorization token
\cite{owaspagentic2026}. OWASP recommends per-action re-authorization and
task-scoped, time-bounded permissions for this class of failure. \framework{}
addresses the stale-authority aspect by binding an
authorized request to the authority-state epoch under which it was admitted
and rejecting execution when that epoch no longer matches the current
authority state, as exercised in the S10 stale-authorization scenario in
Section~\ref{sec:experimental-evaluation}. The novelty claim is therefore not
the identification of authorization drift, TOCTOU risk, re-verification, or
time-bounded authorization itself, but the joined chain from validity- and
applicability-qualified assurance evidence through principal-bounded effective
authority to version-bound consequential execution. This rejection property
remains conditional on authorization-relevant changes
being observed by the governance plane and reflected in the current authority
state, as discussed in
Section~\ref{sec:governance-plane-integrity}.

SAGA addresses a complementary problem at the secure inter-agent interaction
layer~\cite{syros2026saga}. It provides user-controlled agent lifecycle
management, agent registration and discovery, policy-controlled inter-agent
contact, and cryptographically derived access-control tokens for authenticated
agent-to-agent communication. \framework{} does not replace these identity,
discovery, secure-channel, or token mechanisms. However, inter-agent
communication may itself constitute a consequential action governed by
\framework{} when a DGP represents communication actions, recipients,
information classes, delegation, or related interaction constraints as
dimensions of operational authority. \framework{} can therefore condition,
restrict, or revoke an agent's authority to communicate with another agent as
the applicable assurance basis changes, while SAGA provides mechanisms for
securely authenticating and enforcing permitted inter-agent interactions.
SAGA is consequently treated as complementary infrastructure rather than as a
head-to-head authority-state baseline. Governing individual inter-agent
interactions does not, however, by itself solve compositional multi-agent risk,
in which individually authorized actions or communications may combine across
agents into an unsafe consequence; such collusion- and composition-aware
authority reasoning remains future work.

\subsection{Agent Harness Engineering, Runtime Governance, and ACCA}
\label{sec:harness-engineering}
An agent harness provides execution loops, tools, memory, orchestration,
observability, and governance functions. Recent surveys and system studies
make these roles explicit~\cite{li2026harnesssurvey,barbaste2026harness}.
Contemporary runtime controls already support action gating, monitoring, revocation,
permission adaptation, and protected execution boundaries; examples include
MI9~\cite{wang2025mi9}, LATTICE~\cite{calboreanu2026lattice},
Aegis~\cite{mazzocchetti2026aegis}, HarnessAudit~\cite{liu2026harnessaudit},
Microsoft ACS~\cite{microsoftagt2026}, and
OpenAgentFlow~\cite{chen2026openagentflow}.
These techniques are not claimed as novel in themselves.

The contribution addressed here is the explicit assurance-to-authority lifecycle:
evidence can remain authentic while becoming contextually inapplicable following
a model, prompt, memory, tool, harness, or environment change. Typed dependencies
identify affected evidence and claims; applicable policy resolves justified
effective authority within a separately declared ceiling. The trusted
control plane may be implemented outside a harness, in protected harness
components, or across distributed services; agent-controlled proposal and
adaptation logic must not unilaterally rewrite protected policy, evidence
decisions, or authority state. A policy enforcement point is a logical protected
mediator rather than a synonym for the harness.

Early tool-gateway authorization alone cannot guarantee safety of delayed
execution. A final effect-producing path must check the required current version
values and use a commit contract capable of preventing a newly stale operation
from producing the governed effect; if no such path exists, the security claim is
correspondingly limited. The formal model and enforcement assumptions follow.

\subsection{Four Distinct Control Questions}
\label{sec:four-control-questions}

Request-time action authorization, effective-authority continuity,
authorization-basis continuity, and effect-time commitment are related but
distinct control questions. Conflating them can make an architecture appear
continuously authorized even though a favorable decision can be queued, cached,
retried, or delegated and later committed after either the actor-level authority
or an action-specific supporting condition has changed. Table~\ref{tab:four-control-questions}
separates the four questions.

\begin{table}[t]
\centering
\caption{Distinct control questions in a consequential agent workflow.}
\label{tab:four-control-questions}
\small
\setlength{\tabcolsep}{4pt}
\renewcommand{\arraystretch}{1.12}

\begin{tabularx}{\textwidth}{@{}
  >{\RaggedRight\arraybackslash}p{0.20\textwidth}
  >{\RaggedRight\arraybackslash}p{0.31\textwidth}
  >{\RaggedRight\arraybackslash}X
  @{}}
\toprule
\textbf{Control question}
& \textbf{Primary object}
& \textbf{Representative mechanism} \\
\midrule

Request-time action authorization
& Whether a particular submitted request, including its
  material parameters, is permitted under current effective
  authority and applicable action-specific conditions
& A policy decision point (PDP), which may be implemented
  using a policy engine such as Open Policy Agent (OPA),
  evaluates the request; a distinct policy enforcement
  point (PEP) enforces the resulting decision at the
  protected action or effect path. RBAC, ABAC, UCON,
  XACML, or AADP-like protocols may supply policy
  semantics or exchange patterns. \\

\addlinespace[3pt]

Effective-authority continuity
& Whether the effective-authority generation under which
  a consequential action was authorized remains current
& \framework{} effective authority, authority epoch,
  and an epoch-bound action authorization. \\

\addlinespace[3pt]

Authorization-basis continuity
& Whether a previously issued action authorization remains
  acceptable after a configured change to supporting
  action-specific authorization state that leaves effective
  authority unchanged
& Optional authorization-basis version carried or
  referenced by the action-authorization artifact and
  checked when required. \\

\addlinespace[3pt]

Effect-time commitment
& Whether required version comparisons and
  resource-changing commit are protected against a
  concurrent authorization transition
& Transaction, conditional write, provider-local fence,
  protected execution intent, or another explicitly
  stated commitment protocol. \\

\bottomrule
\end{tabularx}
\end{table}

The principal research contribution of \framework{} remains the derivation and
maintenance of effective authority from a currently applicable assurance basis,
together with explicit version interfaces that allow an enforcement substrate to
consume that state. Per-action authorization and transaction/fencing primitives
are established mechanisms and are not claimed here as independently novel. The
optional authorization-basis version closes a distinct continuity gap: an issued
action can require complete reauthorization because its supporting action-specific
basis changed even when the actor-level effective-authority envelope did not.
Without an adequate effect-boundary mechanism, however, either version can identify
stale authorization without guaranteeing that a stale effect is prevented.

Contemporary policy-as-code systems illustrate the request-time substrate shown
in the first row of Table~\ref{tab:four-control-questions}. For
example, the documented EnforceAuth architecture places a customer-operated PEP
in an application, gateway, or middleware, sends a synchronous decision request
to a PDP built on OPA, and builds and distributes policy bundles while collecting
decision logs and status telemetry outside the runtime request path
\cite{enforceauth2026pep,enforceauth2026deployment}. OPA is a general-purpose policy engine that serves as a policy decision point
(PDP): its declarative Rego policies evaluate structured request input against
policy and data \cite{openpolicyagent2026}. OPA is not, by that role, the
ACCA-aware policy enforcement point (PEP) and is not the ACCA authority control
plane. In an ACCA integration, an OPA-based PDP can participate in applicable-policy
mediation, action-specific authorization, or both, using effective-authority state
and associated version information made available as authenticated or
integrity-protected policy data. A distinct ACCA-aware PEP consumes the decision
and enforces it at the protected action or effect path.
Evidence applicability, assurance-claim derivation, authority-epoch advancement,
and the authoritative effect-time comparison remain responsibilities of the
ACCA control plane and required enforcement boundary unless a deployment
explicitly assigns protected implementations of those functions elsewhere.
Because replicated or cached policy data can lag authoritative authorization
state, an earlier favorable PDP decision does not by itself establish that the
authorization remains current at effect time. The documented request-time
pattern does not by itself establish an assurance-derived authority epoch, an
action authorization bound to that epoch, or a comparison-and-commit guarantee
at a downstream provider. Those properties must be supplied by the ACCA
integration and the governed-resource adapter. This is an architectural
comparison to the cited documentation and an unevaluated integration pattern,
not a claim that Rego, OPA, or a particular product is incapable of being
extended to implement them.

\subsection{Stateful Authorization, Commitment, Revocation, and Continuity}

Recent work separates several correctness questions that are easy to conflate.
MasuGate (named Provenact in the first arXiv version) identifies stale authorization under concurrent mutable policy state
and defines policy-state serializability, requiring committed effects to remain
explainable against the policy state immediately preceding the effect
\cite{provenact2026}. Taming Cognitive Transactions (TCT) extends the commit
boundary across data, evidence, policy, and authority dependencies, distinguishing
strict cognitive serializability from effect-compatible recertification after
dependency drift \cite{he2026tct}. These mechanisms address whether a proposed
effect remains justified through commitment. \framework{} addresses the earlier
derivation question of what authority should be supplied to such an authorization
or commit mechanism, given the assurance currently supported by applicable
evidence.

Revocation research addresses a different part of the lifecycle. The
\emph{Bureaucracy of Speed} treats multi-agent revocation as a coherence problem
and bounds residual unauthorized operations under execution-count-directed
coherence \cite{parakhin2026bureaucracy}. Root-scoped quiescence goes further for
long-running delegated and asynchronous work: a root-epoch cut, protected-sink
fences, carrier coverage, exact channel accounting, and composable evidence
establish that a retired authorization root no longer has a path to a new
protected commitment \cite{zhu2026quiescence}. These results sharpen the runtime
obligations behind epoch changes, delegation containment, and in-flight recovery;
they do not derive the assurance-conditioned authority state whose contraction or
revocation initiates those obligations.

The closest structural parallel is Earned Authority
\cite{zhang2026earnedauthority}. Zhang and Zhang fix an immutable user-issued
effect ceiling, permit free contraction below it, require specified evidence for
expansion or incomparable change, and prove a non-amplification result under
complete mediation and related assumptions. Accordingly, neither the containment
relation $A_{\mathrm{eff},u,g}^{(t)}\preceq_d A_{d,u,g}$ nor evidence-gated ascent beneath a
fixed ceiling is claimed here as an independent novelty. \framework{} differs by
making evidence validity and contextual applicability separate objects, deriving
explicit assurance claims from the currently applicable evidence basis, mediating
those claims through applicable policy $P_u$, operationalizing domain semantics through
\dgp{}s, and propagating authorization-relevant change through explicit dependencies.

Earned Authority also separates effect containment from a \emph{transition
envelope}. In the Earned Authority framework, an issued grant can be suspended
because the authorized subject or context has mutated beyond that grant's transition
envelope even when the requested effect remains below the ceiling. Grant continuity
and grant survival are therefore concepts attributed to Earned Authority. The present
\framework{} model has functional overlap through authorization-relevant change
propagation and conditional epoch advancement after an effective-authority change,
but ACCA does not introduce a separate ``grant'' authorization object or a
grant-survival predicate. It instead rederives
$A_{\mathrm{eff},u,g}^{(t)}$ and invalidates stale action authorizations when
required.

Table~\ref{tab:closest-work-comparison} summarizes the most closely related prior work by the object each approach governs, its primary mechanism, and its boundary relative to the assurance-to-authority lifecycle studied by \framework{}.

\begin{table}[t]
\centering
\caption{Closest work and its relationship to the assurance-to-authority
lifecycle studied by \framework{}. The comparison emphasizes different governed
objects and correctness boundaries rather than scoring prior work against
\framework{}-specific constructs.}
\label{tab:closest-work-comparison}

\scriptsize
\setlength{\tabcolsep}{3pt}
\renewcommand{\arraystretch}{1.16}

\begin{tabularx}{\textwidth}{@{}
  >{\RaggedRight\arraybackslash}p{0.135\textwidth}
  >{\RaggedRight\arraybackslash}p{0.17\textwidth}
  >{\RaggedRight\arraybackslash}p{0.255\textwidth}
  >{\RaggedRight\arraybackslash}X
  @{}}
\toprule
\textbf{Work}
& \textbf{Governed object}
& \textbf{Primary mechanism}
& \textbf{Boundary relative to \framework{}} \\
\midrule

Earned Authority \cite{zhang2026earnedauthority}
& Grant continuity and active effects
& Transition envelope; immutable effect ceiling;
  evidence-gated ascent; formal non-amplification
& Establishes the closest ceiling/allocation structure.
  \framework{} adds the explicit
  validity--applicability--claims--applicable-policy
  derivation and \dgp{} operationalization. \\

\addlinespace[3pt]

CapScope / AAE \cite{bouras2026capscope,aae2026}
& Capability scope and authorization representation
& External typed capabilities; machine-evaluable
  mandate and constraints
& Can represent or constrain authority;
  \framework{} derives why a particular authority
  state remains justified. \\

\addlinespace[3pt]

AADP \cite{saha2026aadp}
& Individual consequential actions
& Stateful per-action decisions using budgets,
  approvals, reservations, obligations, and
  kill switches
& Supports per-action authorization using mutable
  decision state; \framework{} additionally derives
  and maintains actor-level effective authority
  from currently applicable assurance evidence
  and policy. \\

\addlinespace[3pt]

MI9 / ACS / OpenAgentFlow
\cite{wang2025mi9,microsoftagt2026,chen2026openagentflow}
& Runtime agent behavior, policy decisions,
  and pending actions
& Continuous permission adaptation;
  deterministic policy decisions;
  shared pre-execution enforcement
& Provide runtime governance and enforcement
  substrates around \framework{}'s
  assurance-conditioned authority state. \\

\addlinespace[3pt]

MasuGate / TCT \cite{provenact2026,he2026tct}
& Committed effects and derivation dependencies
& Policy-state or cognitive serializability;
  commit-time validation and recertification
& Protect effect-time correctness;
  \framework{} determines the
  assurance-grounded authority supplied
  to authorization. \\

\addlinespace[3pt]

Revocation coherence / root-scoped quiescence
\cite{parakhin2026bureaucracy,zhu2026quiescence}
& Distributed revocation and residual effect paths
& Coherence bounds; root cuts, sink fences,
  carrier closure, and cutset evidence
& Addresses revocation propagation and residual-effect
  closure under stated system and enforcement
  assumptions; complementary to
  effective-authority derivation. \\

\addlinespace[3pt]

\framework{}
& Effective operational authority
& Principal-declared ceiling, current state,
  and applicable evidence $\rightarrow$
  assurance claims and policy-mediated
  effective-authority derivation,
  with dependency-aware rederivation
& Determines what portion of declared authority
  remains justified now and exposes that state
  to representation, decision, commit,
  and enforcement layers. \\

\bottomrule
\end{tabularx}
\end{table}

Recent complementary work distinguishes version changes from actual decision
conflicts in selective revalidation \cite{lyu2026selective}, and cryptographically
separates a learning agent's behavior from its operator-signed authority through
semantic-effect gating \cite{qin2026governed}. These address, respectively, pending-action
condition freshness and learning-resistant confinement; ACCA instead maintains the
assurance-conditioned actor-level authority state they may constrain or consume.

\subsection{Research Gap}

Taken together, prior work provides substantial mechanisms for continuous and
contextual authorization, scoped capabilities, delegated authority, immutable
ceilings, evidence-gated allocation below a ceiling, per-action decisions,
runtime policy enforcement, stale-state protection, transaction validation,
revocation propagation, action-boundary mediation, provenance, audit, and
authorization representation \cite{zhang2026earnedauthority,bouras2026capscope,
saha2026aadp,wang2025mi9,microsoftagt2026,chen2026openagentflow,
provenact2026,he2026tct,parakhin2026bureaucracy,zhu2026quiescence}. We therefore
do not claim these mechanisms individually as contributions of \framework{}.

The remaining problem addressed in this paper is the explicit maintenance of
the \emph{justification for operational authority}. A policy state may be
current while the assurance evidence that justified an agent's present degree
of authority is no longer applicable. Conversely, an observed change need not
alter authority when it does not affect the assurance basis. \framework{} makes
that basis explicit:
\[
\boxed{\text{Evidence}
\xrightarrow{\mathrm{validity+applicability}}
\text{Assurance Claims}
\xrightarrow{\mathrm{applicable\ policy}}
\text{Effective Authority}}
\]
and propagates authorization-relevant change through the same evidence-mediated
assurance path, with authority materiality determined by whether the resulting
effective authority changes:
\[
\boxed{\begin{gathered}
\text{Authorization-Relevant Change}
\leadsto \text{Affected Evidence}\\
\leadsto \text{Affected Assurance Claims}
\leadsto \text{Affected Effective Authority},\\
\Delta A_{\mathrm{eff}}\neq 0
\Rightarrow \text{Authority-Material Change}
\end{gathered}}
\]

This evidence-mediated path is not exhaustive: the full derivation also
consumes $X_t$, $P_u$, and $A_{d,u,g}$ directly, so directly evaluated state
predicates and authorized changes to applicable policy or declared authority can
trigger reauthorization without first changing $E_t^*$ or $C_t$. We use
\emph{assurance basis} for the evidence, claims, assumptions, and provenance that
support assurance, whereas the broader \emph{authorization basis}
$\mathcal{B}^{\mathrm{auth}}_t$ also includes the other current inputs and
intermediate governance state relevant to effective-authority derivation or a
particular action authorization. This distinction becomes important when an
action-specific approval, target state, obligation, or enforcement-path property
changes without changing $A_{\mathrm{eff},u,g}^{(t)}$.

\enlargethispage{\baselineskip}
The scoped contribution is this assurance-conditioned authority lifecycle,
together with its separation from domain semantics supplied by a \dgp{} and its
integration with authority-state epochs. Continual-learning and self-evolving
harness work sharpens one instance of the same gap: after an
authorization-relevant learning transition, the system for which a prior
authority state was justified may no longer satisfy the assumptions under which
the supporting evidence was obtained. The cited works primarily investigate learning, retention, harness evolution, and adaptation composition rather than the continued justification of operational authority across learning-induced system changes \cite{karten2026continualharness,kang2026hcl,wu2026harnessdev,nie2026evolvenet,ye2026harnesszero}.

\Needspace{4\baselineskip}
The corresponding research question is therefore: \emph{what portion of the
principal-declared operational authority remains justified now, given the
assurance that can currently be supported by applicable evidence?} For a
learning-induced transition, this includes the narrower question of what prior
authority, if any, remains justified for the evolved system. Earned
Authority-style grant continuity, effect-time serializability, and distributed
revocation closure remain distinct complementary decisions rather than ACCA
authorization objects.

\section{Assurance-Conditioned Continuous Authorization}

\subsection{System State and Runtime Projections}

Let the active, validated profile for domain $d$ be the composition
$\dgp{}_d=\mathcal{G}_{\mathrm{core}}\oplus\mathcal{L}_d$, formalized in
Section~\ref{sec:tgp}. The profile determines the schema and semantics under
which runtime state is interpreted for governance. Different, potentially
overlapping projections of that state may be relevant to effective-authority
derivation, action-specific authorization, operational-feasibility evaluation,
or more than one of these mechanisms. For any active profile, the unified
runtime state retains the same top-level structure:

\begin{equation}
X_t^d =
\langle
S_t^d,
D_t^d,
O_t^d,
H_t^d,
B_t^d
\rangle
\in \operatorname{Val}\!\left(\Sigma_X^{(d)}\right),
\qquad
\Sigma_X^{(d)}
=
\Sigma_{X,\mathrm{core}}\oplus\Sigma_{X,d},
\label{eq:profile-relative-state}
\end{equation}

Here $\operatorname{Val}(\Sigma)$ denotes the set of type-correct state values
admitted by schema $\Sigma$. The schema $\Sigma_{X,\mathrm{core}}$ contains
core state fields and $\Sigma_{X,d}$ contains fields supplied by the domain
layer. The operator $\oplus$ denotes the same validated, type-aware composition
used for the profile itself; it does not create two independent runtime states.
Thus $X_t^d$ is one unified runtime-state value whose fields retain their core
or domain provenance. It is not implied that every field is authority-relevant:
the active DGP, applicable policy, and dependency semantics determine which
portions participate in effective-authority derivation or action-specific
authorization. For operational feasibility we use the projection
\[
X_t^{d,\mathrm{op}} := \pi_{\mathrm{op}}^d(X_t^d),
\]
where $\pi_{\mathrm{op}}^d$ selects the current platform, account, resource,
infrastructure, dependency, and environmental state relevant to technical
executability. This is a view over fields already represented in $X_t^d$, not
a sixth primitive state component. The projections may overlap: a field can be
authority-relevant, action-authorization-relevant, feasibility-relevant, or
relevant to more than one mechanism. When the active profile is clear, we
write $X_t$ for $X_t^d$.

$S_t^d$ denotes the governed system's identity, configuration, and
versioned adaptive state. It includes agent and model type, model
version and digest, prompts, tools, harness and orchestration
versions, runtime configuration, and authority-relevant adaptive
state such as persistent memory, learned skills, capability maps,
routing rules, subagent definitions, generated code or tool
compositions, model adapters, and online-updated parameters,
when present. It also records the identity, version, digest,
and configuration reference of the active DGP and applicable policy. These references identify
the active governance configuration; they do not replace the
governance inputs themselves.

A candidate profile is compiled and validated before activation.
Its activated identity and digest are then recorded in $S_t^d$
and determine which profile semantics govern the interpretation
of subsequent state and evidence. An active-profile transition
or applicable-policy version change is therefore a potentially
authorization-relevant system-state change requiring assessment
under the active dependency and policy semantics. Such a change
is authority-material only if rederivation changes effective
authority. The profile and policy references in $S_t^d$ should
not be confused with $P_u$, which denotes the applicable risk
and governance policy used in authority derivation for
principal $u$.

$D_t^d$ denotes technical and informational dependency state,
including the identity, configuration, availability, and health
of external data sources, services, infrastructure, monitoring
mechanisms, and security controls on which the governed system
depends.

$O_t^d$ denotes current external operating conditions,
including market conditions, account state, available resources,
platform operating conditions, and other domain-specific
environmental conditions where relevant. Unlike $S_t^d$,
which characterizes the governed system and its configuration,
$O_t^d$ characterizes the circumstances in which that system
currently operates. The configuration or health of an external
service is represented as dependency state in $D_t^d$;
conditions of the operating environment or governed resources
are represented in $O_t^d$ according to the composed DGP schema.

$H_t^d$ denotes relevant human, organizational, and stakeholder
context, including modeled approval, supervision, organizational,
or stakeholder conditions where applicable.

These components are typed state categories distinguished by
their primary governance meaning, rather than mutually
exclusive decision roles. Each state field has a defined
location in the composed DGP schema, although related
information may span categories. A field may participate in
effective-authority derivation, action-specific authorization,
operational-feasibility evaluation, or several of these
functions without requiring duplicate state representations.

$B_t^d$ denotes normalized, assurance-relevant behavioral
state derived from observable events over an explicit
observation window, for example
\begin{equation}
B_t^d =
\mathcal{N}_B\!\left(
\mathcal{Z}_{[t-\tau_B,t]}
\right),
\label{eq:behavioral-state-window}
\end{equation}
where $\mathcal{Z}_{[t-\tau_B,t]}$ denotes raw behavioral
telemetry and $\mathcal{N}_B$ is a declared normalization
or feature-extraction procedure. $B_t^d$ does not represent
the agent's complete internal cognitive state. Examples
include normalized properties concerning context integrity,
instruction consistency, task focus, policy adherence,
decision stability, tool-use reliability, and delegation
behavior. Normalized behavioral state can influence evidence
applicability or other authorization-relevant conditions,
but does not itself establish an assurance claim. Raw
observations and assessment artifacts remain evidence
candidates; they are not silently conflated with the
state derived from them.

Let $E_t$ be the set of candidate assurance-evidence artifacts
available at time $t$. The subset eligible to support claims
under the active profile is
\begin{equation}
E_t^*
=
\left\{
e\in E_t
\;\middle|\;
\mathrm{Valid}(e,t)
\land
\mathrm{Applicable}(e,X_t^d)
\right\},
\qquad
\mathrm{Applicable}(e,X_t^d)\equiv\phi_e(X_t^d).
\label{eq:eligible-evidence-set}
\end{equation}
Eligibility here concerns evidence validity and contextual
applicability. An eligible artifact may contribute to assurance
claims according to the active profile's support semantics;
eligibility alone does not establish that a claim is supported.
An additional action- and authority-relative notion of
evidence representativeness is discussed as an extension in
Section~\ref{sec:representativeness}.

Let
$C_t=\operatorname{Support}(E_t^*;\dgp{}_d)$
denote the assurance claims supported by currently eligible
evidence under the active DGP. A profile may also attach a graded status
$\operatorname{status}_t(c)$ to a claim, for example,
$\mathrm{supported}$, $\mathrm{degraded}$, or $\mathrm{unassured}$.
These status labels are profile-defined assessments, not authorization
outcomes. Membership $c\in C_t$ denotes that the claim meets its
configured support criterion; a degraded status does not, by itself,
establish that criterion. Policy defines any corresponding partial-authority
response. The appendix illustrates status labels but does not define a
complete status-aggregation algorithm.

A principal $u$ is associated
with applicable risk and governance policy $P_u$ and declares
the authority ceiling $A_{d,u,g}$ for governed actor $g$.
The policy $P_u$ may comprise stakeholder-specific,
organizational, governed-resource-owner, regulatory, or
other authorized policy inputs. The principal-declared
authority ceiling remains a separate governance input.

We define effective authority as
\begin{equation}
A_{\mathrm{eff},u,g}^{(t)}
=
F(A_{d,u,g},X_t^d,E_t^*,C_t,P_u).
\label{eq:effective-authority-derivation}
\end{equation}

Effective authority is always bounded by the authority
declared by the principal:
\begin{equation}
A_{\mathrm{eff},u,g}^{(t)}
\preceq_d A_{d,u,g}.
\label{eq:acca-authority-containment}
\end{equation}

Accordingly, authority expansion in \framework{} means
an increase in effective authority relative to a prior
effective state within $A_{d,u,g}$; it does not mean that
\framework{} enlarges the principal-declared authority
ceiling. Expansion beyond the existing ceiling requires
an explicit change to declared authority by the principal
or another authorized governance process.

Crucially, availability in $E_t$ does not imply membership
in $E_t^*$, and neither $E_t^*$ nor $C_t$ is presumed
unchanged merely because the agent implementation remains
unchanged. A change in authority-relevant elements of
$X_t^d$ may affect evidence validity or contextual
applicability, alter assurance-claim support, and thereby
change the effective authority justified under applicable
policy. Other state changes may affect directly evaluated
policy conditions or action-specific authorization conditions
without changing evidence or claims. Conversely, a change
confined to $X_t^{d,\mathrm{op}}$ can alter technical executability
without affecting assurance or authorization unless DGP semantics or
applicable policy make that condition authorization-relevant. Thus, not
every change in $X_t^d$ requires an effective-authority transition.

\subsection{Authorization Basis and Action-Specific Authorization}
\label{sec:authorization-basis}

We define the current authorization basis
$\mathcal{B}_t^{\mathrm{auth}}$ as a logical set or
versioned representation of current inputs and
intermediate governance state that can affect either
effective-authority derivation or authorization of a
particular consequential action. It may include
$A_{d,u,g}$, $X_t^d$, $E_t^*$, $C_t$, $P_u$,
active profile state, approvals, obligations, target
or resource state, execution-environment state,
enforcement-path state, and other configured
authorization inputs.

Approvals, target or resource state,
execution-environment state, and enforcement-path
state are not additional top-level inputs to
effective-authority derivation beyond those
represented in Equation~\ref{eq:effective-authority-derivation}.
Where relevant, they are modeled as fields of
$X_t^d$ under the active DGP (for example, within
$H_t^d$, $O_t^d$, or $D_t^d$) and may be obtained
from protected external services. The DGP supplies
the semantics for these state fields, while
applicable policy $P_u$ specifies or selects the
requirements under which they influence effective
authority or action-specific authorization.

The authorization basis is a logical object rather
than a requirement for one monolithic store:
a deployment may maintain its elements in
protected registries, policy services, approval
systems, resource-state stores, and enforcement
adapters. The notation
$\mathcal{B}_t^{\mathrm{auth}}$ is distinct from
$B_t^d$, which denotes the behavioral-state
component of $X_t^d$.

Different decisions use different, potentially
overlapping projections of this basis. Let
$\pi_A(\mathcal{B}_t^{\mathrm{auth}})$ denote the
basis elements whose dependency paths can affect
one or more dimensions of effective authority,
and let $\pi_q(\mathcal{B}_t^{\mathrm{auth}})$
denote the basis elements relevant to authorizing
a particular consequential request $q$.

These are dependency-defined projections rather
than fixed data categories. They may share state
elements, but their membership is determined by
the dependencies relevant to effective-authority
derivation or authorization of the particular
request, respectively. An approval, operating-regime
value, resource state, or enforcement-path property
may affect effective authority, an action-specific
authorization decision, or both, depending on the
active DGP and applicable policy.

Equation~\ref{eq:effective-authority-derivation}
can therefore also be written in compact form as
\begin{equation}
A_{\mathrm{eff},u,g}^{(t)}
=
F_A\!\left(
\pi_A(\mathcal{B}_t^{\mathrm{auth}})
\right),
\label{eq:effective-authority-basis-projection}
\end{equation}
where the expanded inputs remain
$A_{d,u,g}$, $X_t^d$, $E_t^*$, $C_t$, and $P_u$.

A proposed action request $q$ and its material
parameters are not treated as standing
authorization-basis state merely because they
are submitted. They are candidate inputs evaluated
against the current effective-authority envelope
and the action-specific projection:
\begin{equation}
\operatorname{Authorize}(q,t)
=
F_{\mathrm{auth}}\!\left(
q,
A_{\mathrm{eff},u,g}^{(t)},
\pi_q(\mathcal{B}_t^{\mathrm{auth}})
\right).
\label{eq:action-authorization-basis}
\end{equation}

The function $F_{\mathrm{auth}}$ is an architectural
interface for action-specific authorization. It
evaluates the proposed request against the effective
authority already derived by the ACCA authority
resolver and the applicable action-specific
conditions. It does not itself derive or allocate
effective authority, and it may be implemented
using an existing policy or authorization substrate.
Per-action authorization is therefore not claimed
as a new mechanism introduced by \framework{}.

The function may return $\textsf{Permit}$,
$\textsf{Deny}$, $\textsf{Hold}$,
$\textsf{Escalate}$, or a structured equivalent.
Here, \(\textsf{Hold}\) indicates that authorization is pending resolution of a required condition, whereas \(\textsf{Escalate}\) indicates that an additional authorized decision or intervention is required, such as human approval. Neither outcome permits execution; a consequential action may proceed only after a subsequent \(\textsf{Permit}\) decision and satisfaction of the required effect-time enforcement conditions. For compactness, we write $\mathrm{Permit}(q,t)$
for the predicate
$\operatorname{Authorize}(q,t)=\textsf{Permit}$.
This predicate denotes a current request-level
authorization outcome under the applicable
authorization basis, not a declaration or allocation
of effective authority.

The separation between $F_A$ and $F_{\mathrm{auth}}$
distinguishes actor-level effective-authority
derivation from authorization of a particular
consequential request.

This separation explains why a previously issued action authorization can
become stale without a change to effective authority. For example, a trading
agent may retain standing permission to \texttt{place\_order} with a
\$10{,}000 per-order limit. A particular \$8{,}000 order may initially
receive \textsf{Permit} after satisfying a required order-specific compliance
pre-clearance imposed by an illustrative applicable policy independently of
the \$20{,}000 value-triggered approval threshold in Appendix~\ref{app:tgp-spec}. If that pre-clearance is subsequently withdrawn, the same illustrative policy
is configured so that reevaluation of the queued order returns \textsf{Deny}, even
though the actor-level effective-authority envelope remains unchanged.

Where withdrawal of the approval is configured to invalidate previously
issued action authorizations, the optional authorization-basis version
advances while the authority epoch remains unchanged. The effect-time PEP
must then reject an action authorization bound to the superseded required
basis version. This optional basis-version enforcement is not implemented
in the present prototype.

A prospective evaluator may assess a particular
proposed consequential action before authorization
and return a structured assessment record containing,
for example, an assessment identifier, evaluator
identity and version, an assessed action or parameter
digest, estimated risk or harm, confidence or
uncertainty, scope, provenance, and validity
information. Such an assessment is treated as
assurance evidence or authorization-basis state
rather than as an authorization decision.
Applicable policy determines how the assessment
contributes to action authorization or effective
authority, and an action-authorization artifact
can bind or reference the assessment record so
that effect-time enforcement can require the
assessment to correspond to the same operation
and remain current under the applicable
authorization basis. The present prototype does
not implement this prospective-assessment
binding mechanism.

Let $\mathrm{BasisVer}_t$ denote an optional
authorization-basis version used when a deployment
requires issued action authorizations to be
revalidated after selected basis changes. The
version is distinct from the authority epoch.
The resulting versioning logic is:
\begin{equation}
\begin{aligned}
\Delta A_{\mathrm{eff}}\neq 0
&\Rightarrow \text{advance the authority epoch},\\
\Delta A_{\mathrm{eff}}=0\ \land\
\substack{\text{configured action}\\\text{revalidation required}}
&\Rightarrow
\begin{gathered}
\text{retain the authority epoch;}\\
\text{advance }\mathrm{BasisVer},
\end{gathered}\\
\Delta A_{\mathrm{eff}}=0\ \land\
\substack{\text{no configured action}\\\text{revalidation}}
&\Rightarrow \text{retain both values.}
\end{aligned}
\label{eq:authority-basis-versioning-cases}
\end{equation}

A monotonic implementation may encode
$\mathrm{BasisVer}_t$ as a basis epoch. The basis
version need not advance for every mutation of
$\mathcal{B}_t^{\mathrm{auth}}$; it advances only
for changes configured to make one or more
previously issued action authorizations require
revalidation. This avoids turning either version
into an undifferentiated ``something changed''
counter.

When a transition both changes effective authority
and changes configured action-specific supporting
state, the authority epoch advances; any separately
maintained authorization-basis version may also
advance when required by the deployment's
versioning policy.

Effect-time protection applies to every
authorization-state version value whose currentness
is required for the particular action. The authority
epoch is always the effective-authority-generation
value. An authorization-basis version is additional
and conditional: it is checked only when the action
or enforcement path requires freshness of supporting
authorization state that can change without changing
effective authority.

Accordingly, a concurrent authorization-state
transition can be an authority-epoch transition,
a required authorization-basis version transition,
or another required version transition that would
render the issued action authorization stale before
the resource-changing effect occurs.

\subsection{Operational Feasibility}

Action authorization should be distinguished from \emph{operational
feasibility}. Using the operational-feasibility projection
$X_t^{d,\mathrm{op}}=\pi_{\mathrm{op}}^d(X_t^d)$ defined above, let
$\mathrm{Feasible}(q,X_t^{d,\mathrm{op}})$ state that request $q$ is
technically executable under the current platform, account, resource,
infrastructure, dependency, and environmental constraints represented in that
projection. The predicate is derived from runtime state; feasibility is not an
additional authority state or a sixth component of $X_t^d$.
Operational feasibility does not itself confer authority or authorization. The set
of requests eligible to proceed to effect-time enforcement is constrained by both
a current action-specific authorization outcome and current operational feasibility:
\begin{equation}
\mathsf{ExecEligible}_{u,g}^{(t)}
=
\left\{q\ \middle|\ \operatorname{Authorize}(q,t)=\textsf{Permit}
\land \mathrm{Feasible}(q,X_t^{d,\mathrm{op}})\right\}.
\label{eq:authority-feasibility}
\end{equation}
Here, $\textsf{Permit}$ is the current action-specific authorization outcome
defined above; it does not confer effective authority. Actual resource-changing
effect remains subject to the required effect-time PEP validation and any other
enforcement conditions applicable to the action.

Thus a request can lie within the actor-level effective-authority envelope yet fail
a current action-specific authorization condition, while a request with a current
$\textsf{Permit}$ outcome can remain temporarily non-executable because an external
technical constraint is not satisfied. Neither case should be conflated with the
other. Operational feasibility does not create authority or authorization: an action
that an external platform is technically capable of executing remains prohibited
unless it is currently authorized. Conversely, a purely feasibility-related change
need not alter assurance, effective authority, or action authorization unless DGP
semantics identify the condition as authority- or authorization-relevant, or
applicable policy makes authorization depend upon it. This preserves a four-way
distinction among system capability, current operational feasibility, actor-level
effective authority, and action-specific authorization.

\subsection{Assurance and Authorization Dependency Graph}
\label{sec:dependency-graph}

We model assurance provenance and authorization
dependencies as a directed graph
\begin{equation}
G_A=(V_A,E_A),
\end{equation}
whose nodes may represent system components,
configurations, dependency states, environmental
assumptions, operating conditions, tests and
evaluations, evidence artifacts, assurance claims,
policies, effective-authority dimensions,
action-specific authorization conditions, action
decisions, evidence producers, and outcomes.

An edge $x\rightarrow y$ indicates that the validity,
applicability, justification, or authorization result
represented by $y$ may depend on $x$ under the
active DGP semantics. The graph therefore represents
dependencies with different governance meanings,
including evidence applicability, claim support,
policy-mediated authority derivation, and
action-specific authorization.

The graph $G_A$ supports both
$\pi_A(\mathcal{B}^{\mathrm{auth}}_t)$ and
$\pi_q(\mathcal{B}^{\mathrm{auth}}_t)$ without
requiring every authorization-relevant condition
to become a dimension of the effective-authority
envelope. Dependency reachability identifies
potentially affected paths; whether an observed
change is authorization-relevant or authority-material
depends on subsequent semantic reevaluation, as
defined in the following subsection.

For example, a model-specific assurance path may
be represented as
\begin{equation}
M_1
\rightarrow
e_{17}
\rightarrow
C_{\mathrm{model\mbox{-}high}}
\rightarrow
A_{\mathrm{live}}.
\end{equation}
Here, evaluation evidence $e_{17}$ is scoped to
model $M_1$. Replacing $M_1$ with an unassessed
model $M_2$ can make $e_{17}$ non-applicable
to the modified system, potentially withdrawing
support for $C_{\mathrm{model\mbox{-}high}}$ (the model-assurance claim, represented as \texttt{MODEL\_ASSURANCE} in the TGP).
Whether this changes live-execution authority
depends on the remaining eligible evidence,
supported claims, and applicable policy.

Dependencies need not originate in the agent
itself. An environmental assumption may similarly
participate in an assurance path:
\begin{equation}
\mathrm{Volatility}<\sigma_{\max}
\rightarrow
e_{23}
\rightarrow
C_{\mathrm{strategy\mbox{-}safe}}
\rightarrow
A_{\mathrm{live}}.
\end{equation}
If market volatility exceeds the conditions
under which $e_{23}$ was obtained, the evidence
may cease to be applicable to the current context.
The associated assurance claim may then lose
support, even though the model, strategy, tools,
and credentials remain unchanged. Any resulting
authority change is determined by applicable
policy rather than by graph reachability alone.

The evidence artifacts participating in these dependency paths may
originate from heterogeneous \emph{Assurance Evidence Producers} (AEPs).
An AEP is a governed mechanism that observes or evaluates an agent,
proposed consequential action, system component, dependency, operating
environment, or other relevant condition and produces candidate assurance
evidence. AEPs may be deterministic, statistical, human-operated, or
AI-based, including model-integrity checks, behavioral evaluators,
security monitors, environmental monitors, policy checkers, and
independently governed safety evaluators. For example, a market-regime
monitor in the Trading Governance Profile (TGP) may serve as an
environmental AEP by observing market conditions rather than the trading
agent.

\framework{} is evidence-source agnostic. An AEP is logically distinct
from the authority resolver and does not itself grant effective authority
or issue action-authorization decisions. Its operational independence and
trustworthiness depend on deployment, provenance, integrity protections,
and evidence-quality controls. AEP output is candidate evidence subject
to the validity, contextual-applicability, and claim-support semantics of
the active \dgp{}. Applicable policy determines how the resulting
assurance state contributes to effective authority or action-specific
authorization.

The model-specific and environmental dependency paths illustrate the role
of the TGP in supplying trading-domain governance semantics within the
compositional DGP model developed later in the paper. The TGP combines
core governance semantics with trading-specific specializations.

Core governance semantics may include model identity and validation, evidence
provenance and freshness, security and integrity, behavioral evaluation, autonomy,
delegation, human approval, execution environments, obligations, and temporal
validity. Trading-specific extensions specialize evidence classes, applicability
conditions, assurance dependencies, and authority semantics for market regimes,
liquidity, strategy validation, instrument classes, position exposure, and trading
limits.

Applicable policy $P_u$ is instantiated using
these composed DGP semantics and may incorporate
stakeholder-specific, organizational,
governed-resource-owner, regulatory, or other
authorized requirements.

The DGP defines the governance vocabulary,
evidence classes, validity and applicability
semantics, claim-support relationships, dependency
interpretations, and action-specific authorization
conditions. Applicable policy determines how
the currently supported claims and relevant
state affect effective authority and individual
action decisions within the declared ceiling.

Consequently, different principals may use
the same DGP and observe the same applicable
assurance evidence while deriving different
effective-authority envelopes or action-specific
decisions under different applicable policies.
The ACCA mechanism remains domain-independent:
the composed DGP supplies governance semantics,
$P_u$ supplies applicable policy, and ACCA
maintains the justified effective-authority
and action-authorization state.

\subsection{Authorization-Relevant Change and Authority-Material Change}

Not every observed change should trigger complete reauthorization.
Let $M$ denote an \emph{authorization change classifier}, a logical
abstraction of dependency analysis and semantic reevaluation that
assesses the governance significance of an authorization-basis change:

\begin{equation}
\begin{aligned}
M(\Delta\mathcal{B}^{\mathrm{auth}}_t,G_A,P_u,
A_{\mathrm{eff},u,g}^{(t)})
\in \{&
\mathrm{non\mbox{-}authorization\mbox{-}relevant},\\
&
\mathrm{authorization\mbox{-}relevant/
non\mbox{-}authority\mbox{-}material},\\
&
\mathrm{authority\mbox{-}material}\}.
\end{aligned}
\end{equation}

Here, $\Delta\mathcal{B}^{\mathrm{auth}}_t$ represents a transition
from the pre-change authorization basis to a candidate post-change
basis, not simply a notification that a change was observed.
Classification compares relevant pre-transition and candidate
post-transition state through semantic reevaluation. In particular,
authority materiality requires comparing the effective-authority
envelopes before and after reevaluation rather than relying on
dependency reachability or the magnitude of a state change.
More generally, $M$ may return a graded assessment. It is a logical
classification function, not a separately required runtime component
or a general-purpose detector of unknown changes.

We define an \emph{authorization-relevant change}
as a change in any relevant element of the current
authorization basis that can alter (i) the
applicability or validity of assurance evidence,
(ii) support for an assurance claim,
(iii) satisfaction of a policy predicate,
(iv) an action-specific authorization condition
or the continued validity of a previously issued
action authorization,
(v) an adaptation-activation condition, or
(vi) one or more dimensions of effective authority.

The narrower term \emph{authority-material change}
refers to an authorization-relevant change for which
reevaluation alters one or more dimensions of
effective authority itself. Authorization relevance and authority materiality depend on the change's governance effects, not on whether the change originates within the agent or its external environment. A change to the agent itself may be non-authorization-relevant, authorization-relevant without altering effective authority, or authority-material, depending on the affected dependencies, applicable policy, and outcome of reevaluation.
A change that alters evidence validity or
applicability, claim support, or another
authorization input can require reevaluation while
leaving effective authority unchanged.

Distribution shift, out-of-distribution conditions,
and entry into an insufficiently evaluated operating
regime are potential authorization-relevant changes
when authority or action authorization depends on
evidence conditioned on those regimes; they are
authority-material only when the resulting
reevaluation changes effective authority.

Learning-induced changes---including online model
updates, persistent-memory revisions, learned
skills, routing or subagent changes, generated tool
compositions, or harness self-evolution---are likewise
potential authorization-relevant changes when they
alter a dependency, applicability condition, policy
predicate, action-specific authorization condition,
or authority-relevant behavior. Their origin in the
agent's own learning process neither makes them
automatically authorization-relevant nor exempts
them from authorization-relevance and
authority-material-change analysis.

Authorization relevance and authority-material
change are therefore relational rather than
intrinsic. A large external change need not be
authorization-relevant when no authorization
dependency or condition depends on it, while a
seemingly small configuration or dependency change
can be authorization-relevant when it affects the
current authorization basis.

Such a change is authority-material only when
semantic reevaluation produces an effective-authority
envelope different from the pre-transition envelope,
whether through contraction, expansion, or an
incomparable reconfiguration. This is consistent with
the broader UCON principle that mutable subject,
object, and environment attributes can affect
ongoing authorization
\cite{park2004ucon,tavizi2012uconcloud}.

Operationally, the prototype treats dependency
reachability as a candidate-impact filter rather
than as authorization relevance or authority-material
change itself. For a changed basis element,
the dependency graph can identify affected
evidence-mediated paths and direct policy,
authority, or action-decision paths.

On an evidence-mediated path, the system updates
relevant context, reevaluates validity or
applicability for affected evidence, recomputes
supported claims, and applies applicable policy.

On a direct policy or action-specific path, the
relevant conditions are reevaluated without requiring
an intervening change to evidence eligibility or
assurance-claim support. Such a change may alter an
action-authorization outcome while leaving the
effective-authority envelope unchanged.

A change
can be authorization-relevant without being
authority-material when it affects only an
action-specific condition or other configured
supporting basis for previously issued action
authorizations; in that case,
$\mathrm{BasisVer}_t$ may advance while the
authority epoch remains unchanged.

The operational classification can be summarized as
\begin{equation}
\begin{gathered}
\text{Dependency-identified candidate changes}\\[3pt]
\xrightarrow{\text{semantic reevaluation}}\\[3pt]
\left\{
\begin{array}{l}
\text{Non-authorization-relevant changes},\\
\text{Authorization-relevant changes}
\end{array}
\right.\\[5pt]
\text{Authority-material subset when }\Delta A_{\mathrm{eff}}\neq 0.
\end{gathered}
\label{eq:materiality-operational}
\end{equation}

Thus authorization relevance does not necessarily imply a changed
action-authorization decision, and authority materiality is
restricted to changes that alter the effective-authority envelope.

The current prototype principally evaluates
authority-material transitions and selective
non-response to non-authorization-relevant changes.
Production enforcement of authorization-basis-version
invalidation for action-specific changes that leave
effective authority unchanged remains an
architectural extension requiring future evaluation.

This operational definition is deliberately narrower
than a general solution to authorization-relevance
and authority-material-change detection; learning
incomplete dependencies and detecting unknown
changes remain research problems.

The presence of a change in $X_t$ or another
element of the authorization basis is therefore
not sufficient to classify the change as
authorization-relevant. For example, a change
in account balance, buying power, available
compute, service capacity, or another external
resource may alter
$\mathrm{Feasible}(q,X_t^{d,\mathrm{op}})$
without changing evidence validity or
applicability, supported claims, applicable
policy, an action-specific authorization
condition, or
$A_{\mathrm{eff},u,g}^{(t)}$.

By contrast, if the changed condition is an
explicit assumption or dependency of evidence
supporting current authority, the same contextual
change can enter the assurance-reevaluation path.
The effect of a contextual change is therefore
determined by its dependency and policy
relationships rather than by the mere fact
that observable state changed.

\subsection{Action-Centric Authority Vector / Envelope}

Authority is multidimensional, but its primary operational unit is the
\emph{consequential action}: an externally relevant operation that can change
the state of a governed resource or produce a material effect in the
environment. Examples include placing an order, transferring funds, deploying
software, modifying a care plan, or invoking a privileged external service.

\Needspace{10\baselineskip}
For principal $u$ and governed actor $g$, we represent the effective authority
envelope at time $t$ as
\begin{equation}
A_{\mathrm{eff},u,g}^{(t)}
=
\langle
\Pi,R,S,V,L,D,H,E,T,O,\tau
\rangle,
\end{equation}
where $\Pi:\mathsf{Act}_d\rightarrow
\{\mathrm{permit},\mathrm{prohibit},\mathrm{unspecified}\}$ is the
standing action-permission component of the effective-authority envelope;
$R$ denotes governed resources; $S$ action scope; $V$ value or risk limits;
$L$ autonomy level; $D$ delegation rights; $H$ human-approval requirements;
$E$ permitted execution environments; $T$ operating constraints;
$O$ obligations; and $\tau$ temporal validity.

The tuple is a structured representation, not an assumption that
authority dimensions are independent. Constraints may span several
dimensions; for example, an order's maximum permitted value may depend
on the financial instrument or execution autonomy level.

The listed dimensions are not exhaustive. The authority-envelope symbols
$R,S,V,L,D,H,E,T,$ and $O$ in this subsection are local authority-dimension
labels. They should not be confused with profile-level objects such as
$\mathcal{R}_d$ or $\mathcal{L}_d$, graph-node notation $V_A$, the state
components $S_t^d,D_t^d,H_t^d,O_t^d$, or the evidence set $E_t$ defined
earlier.

A deployment governing continual adaptation may additionally represent
adaptation authority and learning-zone restrictions as independently
constrainable authority dimensions, governed action families, obligations,
or other profile-defined constraints.

Accordingly,
\[
\operatorname{Allow}(A_{\mathrm{eff},u,g}^{(t)})
=
\{a\in\mathsf{Act}_d\mid\Pi(a)=\mathrm{permit}\}.
\]
Prohibited and unspecified consequential actions are not in the
permitted-action projection. Membership in that projection denotes
standing permission, not complete authorization of particular requests.

For example, consider the trading agent introduced in
Section~\ref{sec:authorization-basis}. Its effective-authority
envelope may permit \texttt{place\_order} with a maximum order
value of $\$10{,}000$. A proposed $\$8{,}000$ order may satisfy
all applicable envelope constraints yet receive
$\operatorname{Authorize}(q,t)=\textsf{Hold}$ because a required
order-specific compliance pre-clearance has not yet been obtained,
or $\textsf{Deny}$ if previously obtained pre-clearance has been withdrawn
under the illustrative policy. Thus, standing permission and satisfaction of the
envelope's constraints do not, by themselves, establish authorization
of a particular request.

This separation is intentional. The effective-authority envelope
represents standing, assurance-justified operational authority,
whereas action-specific authorization evaluates a particular request,
target, or execution context. Encoding all such conditions within
the envelope could unnecessarily couple request-specific changes to
authority rederivation and invalidate unrelated authorizations. The
active DGP and applicable policy determine which conditions affect
effective authority, individual action authorization, or both. Where
a configured action-specific change leaves effective authority
unchanged, an optional authorization-basis version can require
reauthorization without advancing the authority epoch.

Profile-defined assurance requirement-selection rules
$\mathrm{Req}_d$ are inputs to authority derivation rather than, by
default, an additional blanket predicate evaluated independently
at every PEP. A policy may designate a particular assurance claim
as a hard prerequisite for an action, but in the general case the
effects of assurance state are already reflected in the derived
effective authority.

Examples of conditions that may remain action-specific include
target-specific approval, current resource or beneficiary state,
obligations, freshness requirements, and enforcement-path properties.
Changes to these conditions can require new action authorization
without changing the actor-level effective-authority envelope.

Actions are primary in the sense that the remaining authority dimensions
constrain the circumstances under which permitted action types may be
exercised. The effective-authority envelope specifies these standing
constraints, while $F_{\mathrm{auth}}$ evaluates each proposed consequential
request against that envelope and the applicable action-specific authorization
conditions.

This representation permits selective, multidimensional authority
changes. An agent authorized for autonomous live execution may,
after an assurance downgrade, retain analysis and recommendation
authority while losing autonomous execution authority. Similarly,
a value limit may contract without affecting unrelated analytical
actions. Conversely, newly established assurance evidence may
justify selective expansion of particular authority dimensions
without changing unrelated permissions.

Some changes may be incomparable under $\preceq_d$, narrowing
certain dimensions while broadening others; these are lateral
authority reconfigurations rather than pure contractions or
expansions. Any resulting change to the effective-authority
envelope advances the authority epoch.

The framework does not require continuous interpretation of
every internal reasoning step or generated token as an
authorization event. Instead, authorization-relevant state changes trigger appropriate
reevaluation, and effective authority is updated only when rederivation
changes the envelope. Consequential actions are mediated by protected
enforcement points against current effective authority and any other
required authorization-state versions at the effect boundary.

\subsection{Applicable Risk and Governance Policy}

The same assurance state need not imply the same effective authority
for every principal or deployment. Recall that $P_u$ denotes the
applicable risk and governance policy for principal $u$; it may
combine authenticated and authorized stakeholder-specific,
organizational, governed-resource-owner, regulatory, or other
policy inputs.

Different principals may use the same validated DGP and observe
the same currently applicable assurance evidence while applying
different governance requirements. For example, given the same
less-assessed trading model, one principal's policy may prohibit
order placement entirely, while another may permit constrained
paper trading under value limits and human confirmation.
Organizational or governed-resource-owner policies may also impose
constraints independently of an individual principal's preferences.

\Needspace{4\baselineskip}
\framework{} therefore separates assurance, applicable policy,
and declared authority. Assurance characterizes which claims are
currently supported by valid and applicable evidence. Applicable
policy determines how those claims, together with relevant current
state and other authorized inputs, affect the operational authority
justified for a particular principal and context. The
principal-declared ceiling $A_{d,u,g}$ remains a separate bound:
policy-mediated derivation cannot expand effective authority
beyond that ceiling.

\subsection{Capability, Behavioral Role, and Operational Authority}

For general-purpose and LLM-based agents, technical capability,
assigned behavioral role, and operational authority are distinct
concepts.

\Needspace{10\baselineskip}
Let $K_g$ denote the set of actions or behaviors that agent $g$
is technically capable of producing given its model, tools,
credentials, and execution environment. Let $R_g$ denote its
assigned functional or behavioral role, and let $A_{d,u,g}$
denote the principal-declared authority ceiling established by
principal $u$. The effective authority
$A_{\mathrm{eff},u,g}^{(t)}$ is derived within that ceiling
under the applicable assurance and governance conditions.

Neither technical capability nor an assigned behavioral role
establishes or expands the principal-declared authority ceiling
or the effective authority derived within it.

A general-purpose LLM instructed to act as a research analyst
may nevertheless be capable of constructing an order, requesting
execution, invoking another tool, or proposing actions outside
its assigned behavioral role. Role prompts and agent instructions
may provide useful behavioral steering but do not, by themselves,
establish enforceable authorization boundaries. Authenticated role
assignments may contribute to applicable policy or authorization
decisions when explicitly represented and enforced by the
governance system.

Similarly, possession of an execution-capable tool does not
imply permission to exercise that capability. A research agent
may technically be able to construct a valid broker request
while its effective-authority envelope prohibits
\texttt{place\_order}.

Technical capability is also distinct from request-specific
operational feasibility. An agent may possess the tools and
competence needed to perform an action while current resource
or environmental conditions make a particular request
infeasible. Neither capability nor feasibility, by itself,
confers authorization.

Accordingly, \framework{} separates behavioral alignment from
operational authorization. The model and its instructions
influence what an agent may generate or propose; the governance
system determines which consequential actions are authorized
under the applicable authority and policy constraints.

Runtime enforcement therefore occurs at protected consequential
action boundaries rather than relying on the model to remain
voluntarily within its assigned behavioral role.

\subsection{Continual Learning, Self-Evolution, and Authority Continuity}
\label{sec:continual-learning-authority}

For a continually adapting agent, let $\Lambda_g^{(t)}$ denote the versioned
\emph{adaptive state} capable of changing future action selection or execution as
a result of experience. Depending on the system, $\Lambda_g^{(t)}$ may include
model parameters or adapters, agent-controlled prompts, persistent memory,
learned skills, a capability map, routing rules, subagent definitions, generated
code, tool compositions, or harness/orchestration configuration. Protected
applicable policy, principal-declared authority ceiling, assurance records, authority
snapshots, epochs, and enforcement state are not part of this agent-modifiable
adaptive state.

A learning transition
\begin{equation}
\Lambda_g^{(t^-)}\rightarrow\Lambda_g^{(t^+)}
\end{equation}
may increase, decrease, or qualitatively change technical capability $K_g$.
Neither the learning event nor the resulting capability change
independently modifies the principal-declared authority ceiling or
establishes a change in effective authority. A ceiling change requires
a separately authorized governance operation.

When an adaptive-state transition is authorization-relevant,
effective authority is rederived from the post-change authorization
basis rather than inherited from actor identity:
\begin{equation}
A_{\mathrm{eff},u,g}^{(t^+)}
=
F(A_{d,u,g},X_{t^+}^d,E_{t^+}^*,C_{t^+},P_u).
\label{eq:learning-authority-rederivation}
\end{equation}
The notation $t^-$ and $t^+$ identifies pre-transition and
post-transition states. Not every learning-induced change is
authorization-relevant, and rederivation need not change the envelope.

Harness distillation creates an additional adaptive-state case. A system may
transfer behavior previously induced by a specialized harness into model
parameters and then operate under a smaller target harness. The post-distillation
system can preserve or improve task capability even though the location
and mechanism of that capability have changed. Under \framework{}, this
model--harness migration is treated as an adaptive-state transition: evidence
whose scope depends on the former model--harness composition is reconsidered for
applicability to the new composition, and learned behavioral imitation does not
replace a protected enforcement control when such a control is required by the
applicable \dgp{} or applicable policy. Harness-Zero provides a concrete example
of this capability migration pattern \cite{ye2026harnesszero}.

This makes the validity--applicability distinction especially important. An
evaluation performed on harness version $h_{17}$ can remain authentic,
unexpired, and therefore valid after the agent evolves to $h_{18}$ while ceasing
to apply to the modified system:
\begin{equation}
\mathrm{Valid}(e_{H17},t^+)=\mathrm{true},
\qquad
\mathrm{Applicable}(e_{H17},X_{t^+})=\mathrm{false},
\end{equation}
when the evidence scope or assumptions bind it to properties changed by the learning transition. The transition need not invalidate the original evaluation; rather,
it may make the evaluation inapplicable to the modified system.
Conversely, evidence whose declared scope and assumptions cover the
relevant adaptation class may remain applicable. Authority
continuity is therefore an outcome of evidence and policy rederivation, not an
identity-based presumption.

Depending on policy and evidence supporting individual authority
dimensions, a deployment may adopt or combine the following responses
to an authorization-relevant learning transition:
\begin{itemize}[leftmargin=*]
\item \textbf{fail-closed requalification}, in which selected consequential
authority is temporarily reduced to a sandbox, recommendation, or approval-only
state until new evidence is established;
\item \textbf{selective continuity}, in which only authority dimensions whose
supporting evidence remains applicable are retained; or
\item \textbf{evidence-scoped continuity}, in which prior authority is retained
when existing evidence explicitly covers the post-change adaptive state and
applicable policy continues to accept it.
\end{itemize}
An adaptive-state version change does not itself advance the authority
epoch. If rederivation yields the same effective-authority envelope,
the authority epoch remains unchanged under Eq.~\ref{eq:epoch-transition-rule}; when the
adaptive transition nevertheless changes supporting action-authorization state in
a manner configured to require revalidation, $\mathrm{BasisVer}_t$ may advance
separately under Eq.~\ref{eq:authority-basis-versioning-cases}. If effective
authority changes, the authority epoch advances and previously issued actions
bound to the superseded epoch are rejected at required enforcement boundaries.

Thus, continuity of actor identity or adaptive state does not establish
continuity of assurance or effective authority. Retention, restriction,
or expansion of authority after an authorization-relevant learning
transition depends on post-change evidence, supported claims, relevant
state, policy, and the principal-declared ceiling.

\subsection{Behavioral Assurance}
\label{sec:behavioral-assurance}

Agentic systems can exhibit unsafe, unauthorized, or otherwise
undesirable behavior even when their underlying model, configuration,
credentials, tools, and relevant external operating conditions remain
unchanged. Runtime behavior may reveal problems not captured by prior
evaluations, including unreliable behavior, context confusion, instruction
conflict, corrupted or adversarial context, task
drift, unstable planning, erroneous tool use, and unexpected interactions
among agents.

Examples discussed in agent-safety and agentic-security guidance include
goal hijacking, context and memory poisoning, unsafe tool use, task or
instruction drift, and rogue-agent behavior \cite{owaspagentic2026}, together
with confusion, distraction, instruction clash, and related forms of
behavioral degradation. \framework{} does not require these phenomena to be
modeled as psychological states of the agent. Instead, they are treated as potential sources of observable evidence relevant
to behavioral assurance claims, without assuming that an unexpected behavior
necessarily constitutes a failure.

For one configured profile, the behavioral-state component $B_t^d$ of
the authority-relevant state $X_t^d$ may be represented as

\begin{equation}
B_t^d =
\langle
b_c,b_i,b_f,b_p,b_d,b_u,b_g
\rangle,
\end{equation}

where $b_c$ denotes context integrity, $b_i$ instruction consistency,
$b_f$ task focus, $b_p$ policy adherence, $b_d$ decision stability,
$b_u$ tool-use reliability, and $b_g$ delegation reliability.
These components are normalized, potentially uncertain assessments
derived from observable evidence rather than necessarily direct intrinsic
properties of the model. Their interpretation and aggregation are defined
by the active \dgp{}. Domain profiles or implementations may introduce
additional properties; the listed components are illustrative rather than mandatory.

Corresponding assurance claims may include

\begin{equation}
\begin{aligned}
C_B=\{&
\mathrm{CONTEXT\_INTEGRITY},
\mathrm{INSTRUCTION\_CONSISTENCY},
\mathrm{TASK\_FOCUS},\\
&
\mathrm{POLICY\_ADHERENCE},
\mathrm{DECISION\_STABILITY},
\mathrm{TOOL\_USE\_RELIABILITY},\\
&
\mathrm{DELEGATION\_RELIABILITY}
\}.
\end{aligned}
\end{equation}

Behavioral claims are supported by observable evidence rather than assumed
from the model's assigned role or prompt. Such evidence may be supplied by
context-provenance controls, policy checkers, tool-call validators, task-scope
monitors, execution monitors, cross-agent consistency checks, human
supervision, or specialized behavioral evaluators. A behavioral observation
may produce candidate evidence, but it does not automatically establish,
degrade, or withdraw an assurance claim. Claim support depends on the
validity, applicability, and interpretation of evidence under the active
\dgp{}.

\paragraph{Behavioral evidence sources.}
Behavioral assurance uses the general AEP model introduced in
Section~\ref{sec:dependency-graph}; the sources listed above are
examples. Their observations produce candidate evidence rather than
authorization decisions. Evidence eligibility and claim support
remain subject to the active \dgp{}, and applicable policy
determines whether supported behavioral claims affect effective
authority, action-specific authorization, or neither.

AI-generated behavioral assessments require sufficient
provenance, scope, freshness, confidence or uncertainty, and
applicability information for evaluation under the active
\dgp{}. Evidence about an evaluator's own integrity and
reliability may itself be necessary to establish sufficient
confidence in such assessments. Greater evaluator sophistication
does not, by itself, establish assurance or confer authority.

When behavioral evidence is valid, applicable, and relevant to
active governance requirements, it can enter the same
assurance-dependency mechanism used for system and environmental changes.
Conceptually, relevant dependencies may be expressed as
\begin{equation}
\begin{gathered}
\text{Behavioral observation}
\longrightarrow \text{Candidate behavioral evidence}\\
\longrightarrow \text{Supported assurance claims}
\overset{P_u}{\longrightarrow}
A_{\mathrm{eff},u,g}^{(t)}.
\end{gathered}
\end{equation}
The arrows represent potential dependencies rather than automatic
propagation or a guaranteed authority change. Behavioral evidence
may also affect action-specific authorization without altering
the effective-authority envelope.

For example, evidence that retrieved untrusted content has begun
influencing an agent contrary to higher-priority policy may reduce
support for \texttt{CONTEXT\_INTEGRITY}. Under one illustrative
trading policy, sufficiently supported degradation of context
integrity may permit continued analysis while restricting autonomous
consequential execution and requiring human approval for affected
orders. These constraints depend on the deployment's autonomy
levels and applicable policy; they are not universal ACCA responses.

Similarly, repeated contradictory decisions or inconsistent tool arguments
may weaken evidence supporting \texttt{DECISION\_STABILITY} or
\texttt{TOOL\_USE\_RELIABILITY}. The resulting authority response need not
disable the agent entirely; policy may instead contract monetary limits,
reduce autonomy, prohibit selected consequential actions, require human
approval, refresh the agent's context, or terminate only the affected task.

Importantly, \framework{} does not assume that a single anomalous observation
proves a behavioral failure. An agent may legitimately revise a decision,
explore alternatives, disagree with another agent, or investigate information
outside an initial hypothesis. Behavioral assurance therefore should support
graded evidence, persistence thresholds, severity, provenance, and multiple
sources of evidence.

A deployment may record and investigate an isolated low-severity
anomaly, require persistence or corroboration before changing claim
support, or apply immediate intervention when sufficiently credible
high-severity behavior satisfies an applicable policy condition.
These are illustrative policy-dependent responses, not universal
escalation thresholds. A single severe observation may justify
intervention, while repeated benign observations need not justify
an assurance downgrade.

The contribution of the present work is not a new anomaly-detection
algorithm for context poisoning, confusion, distraction, instruction
clash, tool misuse, or related agent failures. Rather, \framework{}
specifies how evidence from such detectors or evaluators can be
assessed for validity and applicability, incorporated into supported
assurance claims, and used under applicable policy to reevaluate
effective authority or action-specific authorization. Evaluator
accuracy, coverage, calibration, and robustness remain independent
assurance concerns, and developing and evaluating specialized
detectors for these conditions remains future work.

\subsection{Authority Epochs, Authorization-Basis Versions, and Stale-Authorization Rejection}

Let $\mathrm{Epoch}_t$ be a monotonically ordered value that versions the
current effective-authority generation. The epoch is associated with the
effective-authority state of a particular principal--actor pair; its
indices are suppressed in $\mathrm{Epoch}_t$ for readability. For exposition,
Eq.~\ref{eq:epoch-transition-rule} uses a monotonically increasing integer
encoding; other ordered encodings can implement the same
new-generation-versus-retained-generation invariant. Equality and inequality
of effective-authority envelopes here refer to canonical authority states
(or equal authorization semantics), not representation-only differences. For a reauthorization
event at time $t$, the integer encoding advances if and only if the
post-transition effective authority differs from the pre-transition effective
authority:

\begin{equation}
\mathrm{Epoch}_{t^+}
:=
\begin{cases}
\mathrm{Epoch}_{t^-}+1,
&
A_{\mathrm{eff},u,g}^{(t^+)}
\neq
A_{\mathrm{eff},u,g}^{(t^-)},\\[3pt]
\mathrm{Epoch}_{t^-},
&
A_{\mathrm{eff},u,g}^{(t^+)}
=
A_{\mathrm{eff},u,g}^{(t^-)}.
\end{cases}
\label{eq:epoch-transition-rule}
\end{equation}

Thus an authorization-relevant change may cause reevaluation or rederivation
without advancing the authority epoch when the resulting effective-authority
envelope is unchanged. Conversely, an
authorized change to $P_u$ or $A_{d,u,g}$ advances the epoch whenever it
changes effective authority, even if $E_t^*$ and $C_t$ are unchanged. The
$+1$ records a new authority-state generation; it denotes neither the
direction nor the magnitude of the authority change, nor elapsed time.

Separately, when a change to $\mathcal{B}^{\mathrm{auth}}_t$ leaves the
effective-authority envelope unchanged but is configured to require revalidation
of one or more previously issued action authorizations, the deployment may advance
$\mathrm{BasisVer}_t$ while retaining $\mathrm{Epoch}_t$. This can occur, for
example, after withdrawal of an action-specific approval, a target-state change,
or loss of an enforcement-path property required for a previously authorized
action. A monotonic encoding of $\mathrm{BasisVer}_t$ is a basis epoch. The basis
version is not required to change after every basis mutation; it is a configured
freshness generation for issued action authorizations whose supporting basis must
be reconsidered.

Let $q$ denote a consequential action request submitted for enforcement and
let $\epsilon_q$ be the authority epoch under which it was authorized. A PEP
rejects $q$ if its carried epoch no longer matches the current authority epoch
at enforcement time. Formally,

\begin{equation}
\epsilon_q \neq \mathrm{Epoch}_t
\quad\Rightarrow\quad
\operatorname{Deny}(q).
\label{eq:stale-epoch-deny}
\end{equation}

Here, $\operatorname{Deny}(q)$ denotes rejection of the submitted action
authorization at the enforcement boundary, not a permanent denial of the
operation. A new authorization decision may be requested under the current
authorization basis.

In a deployment, the carried authorization can be represented by an
integrity-protected record or artifact $\alpha_q$ binding the actor, action,
resource, material action parameters or their digest, validity interval, the
authority epoch $\epsilon_q$ under which it was issued, and any required authorization-basis version, approval, delegation, nonce,
or idempotency reference. The required version components can therefore be
represented as a version vector. At effect time, the PEP additionally validates
each artifact-carried condition required for the action, which can include
expiration, replay protection, validity of an approval reference, validity of a
delegation chain, satisfaction of an obligation, and consistency of required
epoch or version values with corresponding current authoritative state. A PEP
rejects the artifact when $\epsilon_q$ does not match the current authority epoch
or when another required version value is stale, mismatched, or unverifiable.
Such a mismatch does not refresh $\alpha_q$ in place. If policy permits
reconsideration, the operation undergoes complete authorization against the
current authorization basis and receives a new artifact only if
Eq.~\ref{eq:action-authorization-basis} succeeds; otherwise it remains denied.

In a digest-based implementation, let $m_q=\operatorname{Material}(q)$
denote the digest-binding input selected from the authorized request
$q=\langle g,a,r,x,\eta\rangle$. It includes the action $a$, target
resource $r$, material parameters $x$, and execution environment $\eta$; consequential identifiers are included in $x$ unless separately and securely bound to the authorization artifact. The authorization service computes
\[
d_q=\operatorname{Hash}(\operatorname{Canonicalize}(m_q)).
\]
At the effect boundary, the PEP computes the corresponding digest from the
material parameters presented for execution and rejects the operation when the
digests differ. Thus an authorization for one materially specified operation
cannot be reused after a consequential parameter has been modified.

The evaluated prototype carries the authority epoch through its native order and
governed-execution objects, but does not claim to implement a general
cryptographic artifact format or the general digest-based artifact mechanism
just described.

Under complete mediation, accurate current authoritative state, and an
enforcement contract protecting the specified effect boundary, version
validation prevents an action authorization bound to superseded required
versions from producing that protected effect. This is the intended conditional enforcement contract of Section~\ref{sec:enforcement-contracts}, not a property formally proved or experimentally demonstrated for distributed providers here. The action authorization may
be carried through a plan, cached token, delegated instruction, queue, or
other execution artifact. An unchanged authority epoch does not establish
that a request-specific approval or supporting condition remains valid;
required basis-version checks can independently require reauthorization.

An upstream comparison alone does not close a race with a non-cooperating
provider after submission. Protection depends on whether required currentness
checks and the governed effect are transactional, fenced, protected by a
controlled execution intent, or merely adjacent in time. Required versions
always include the authority epoch and may also include a configured
basis version or other version-vector components. The guarantee therefore
extends only to the protected boundary and enforcement contract actually
provided by the deployment.

The present implementation maintains one authority epoch for the complete
multidimensional effective-authority envelope associated with each
principal--actor authority state. A change in any dimension that changes
this envelope advances the epoch even if unrelated dimensions are unchanged.
The optional authorization-basis version addresses a different case:
configured action-specific supporting state changes that leave the envelope
unchanged.

Finer-grained scoped authorization-state versions, for example by task,
resource, action family, or session, are a possible extension. Avoiding
invalidation of unaffected actions after an actor-level authority change
would require independently versioned scoped authority projections or
equivalent protected scoped authorization state, with dependency-aware
binding and enforcement. Such mechanisms are not implemented or evaluated
in the present work.

\section{Trading Governance Profile}
\label{sec:tgp}

The \framework{} authority model is intentionally domain-independent. It
defines \emph{how} effective authority is maintained from declared authority,
currently applicable assurance evidence, supported claims, operating conditions,
and applicable policy, but it does not itself determine \emph{what those
constructs mean} in a particular application domain. It does not know which
actions are consequential, which resources and quantitative constraints matter,
which evidence is relevant, under what assumptions that evidence applies, which
assurance claims are required, which changes are authorization-relevant, or how those
claims constrain domain-specific authority dimensions. Practical deployment
therefore requires a systematic operationalization layer between domain governance
knowledge and the generic authority lifecycle.

We use a harness-neutral \emph{Domain Governance Profile} (\dgp{}) as that
operationalization and semantic layer. Rather than treating each domain profile as
an independent governance vocabulary, we model a \dgp{} compositionally as
\begin{equation}
\boxed{
\dgp{}_d = \mathcal{G}_{\mathrm{core}} \oplus \mathcal{L}_d,
}
\label{eq:dgp-composition}
\end{equation}
where $\mathcal{G}_{\mathrm{core}}$ is the core governance layer and
$\mathcal{L}_d$ is a domain-specific governance layer. The core layer defines the common
syntax, semantics, configuration mechanisms, evidence and assurance constructs,
authority dimensions, dependency relations, and authorization-relevance and authority-material-change constructs that can
be reused across agent domains. Examples include agent and model identity, evidence
provenance and freshness, model-validation and integrity evidence, behavioral and
security assurance, human-approval requirements, autonomy and delegation constraints,
execution environment, obligations, and temporal validity. Core concepts retain their common governance meaning across domains,
while domain-specific specializations and applicable policy determine
particular requirements, constraints, and operational interpretations.

The operator $\oplus$ denotes \emph{validated, type-aware composition}, not
arithmetic addition or an unqualified set union. A domain layer may add namespaced
concepts and refine declared extension points, but it may not silently weaken core
invariants. Profile compilation checks declared references, types, namespaces,
claim-to-evidence mappings, dependency relations, and authority-bound rules
according to its specified validation procedures. A conforming validator
rejects detected violations of core invariants and declared constraints.
These checks do not establish completeness of the dependency graph or
correctness of all domain governance semantics. The validated result
includes governance semantics and the composed state schema used at runtime.

The domain-specific layer specializes this common vocabulary with concepts whose
governance meaning depends on domain $d$. Conceptually,
\begin{equation}
\mathcal{L}_d = \langle
\mathsf{Act}_d,\mathcal{R}_d,\Sigma_{X,d},\mathcal{E}_d,\Phi_d,\mathcal{C}_d,
\mathrm{Req}_d,G_d,P_d,M_d,J_d
\rangle,
\label{eq:dgp-conceptual}
\end{equation}
where $\mathsf{Act}_d$ and $\mathcal{R}_d$ identify the governed action vocabulary and governed resources,
$\Sigma_{X,d}$ defines domain-origin fields that extend the core state schema,
$\mathcal{E}_d$ identifies domain-relevant evidence classes, $\Phi_d$ captures
domain-specific applicability conditions, $\mathcal{C}_d$ captures the
assurance-claim vocabulary, $\mathrm{Req}_d$ captures action- and
environment-specific assurance-requirement selection, $G_d$ captures, or is
derivable from, authorization-relevant dependency relationships, $P_d$ captures domain
policy vocabulary and templates, $M_d$ captures authorization-relevance and authority-material-change semantics, and $J_d$
defines or references the enforcement-property vocabulary and mappings used to
state what an execution path must provide for an action class, resource class,
authority level, execution environment, or consequence category. $P_d$ is not
the applicable policy $P_u$: $P_u$ is a configured policy
instance expressed using the available core and domain policy vocabulary. The tuple is
a conceptual interface rather than a requirement that every implementation serialize
all components into a single artifact. More generally, the compiled DGP acts as a
semantic contract over $\mathcal{B}^{\mathrm{auth}}_t$: it defines or references
the types, predicates, rule operators, applicability semantics, dependency relations,
action requirements, authorization-relevance and authority-material-change semantics, and enforcement mappings used to
interpret both $\pi_A(\mathcal{B}^{\mathrm{auth}}_t)$ and
$\pi_q(\mathcal{B}^{\mathrm{auth}}_t)$. The DGP need not store current approval,
resource, target, or enforcement-path values; those runtime values instantiate the
profile semantics in protected runtime services. Nor does the DGP itself allocate effective authority.

A deployed profile is then instantiated for a principal $u$ and governed agent or
actor $g$ by configuration over the composed semantics:
\begin{equation}
\dgp{}_{d,u,g}^{\mathrm{inst}}
=
\operatorname{Configure}
\!\left(\mathcal{G}_{\mathrm{core}}\oplus\mathcal{L}_d, \Gamma_{u,g}\right),
\label{eq:dgp-instantiation}
\end{equation}
where $\Gamma_{u,g}$ represents principal- and actor-specific profile
configuration, including applicable requirements, thresholds, risk posture, and
a reference to the applicable policy $P_u$ used in authority derivation. Two actors may therefore share the same syntax,
semantics, evidence types, authority dimensions, and configuration tools while being
subject to different requirements because their tasks have different criticality or
consequence. Configuration selects and parameterizes requirements using
the composed DGP semantics, while $P_u$ remains the separately authorized
policy input to authority derivation. Shared governance semantics do not
imply uniform governance policy.

The principal-declared authority ceiling $A_{d,u,g}$ is instantiated over these composed semantics and
remains the ceiling on effective authority; the profile does not itself allocate
additional effective authority. A conforming \dgp{} may be realized using structured
configuration, graph-based representations, ontologies, rule systems, or combinations
thereof, provided that the required governance semantics and dependencies can be
interpreted by the \framework{} control plane.

The configured DGP provides the vocabulary, state schema, evidence and
assurance semantics, dependency interpretations, and enforcement mappings
needed by the ACCA control plane. Applicable policy and the principal-declared
authority ceiling remain separately authorized inputs. AEPs supply candidate
evidence; ACCA derives and maintains effective authority; and protected PEPs
mediate consequential effects within the resulting constraints.

The \emph{Trading Governance Profile} (TGP) is the evaluated reference
composition used in this paper:
\begin{equation}
\boxed{
\mathrm{TGP}
=
\mathcal{G}_{\mathrm{core}}
\oplus
\mathcal{L}_{\mathrm{trading}}.
}
\label{eq:tgp-composition}
\end{equation}
TGP defines a machine-interpretable governance model comprising
trading-domain concepts, relationships, constraints, and authority-relevant
dependencies. Its semantics support the expression of a principal-declared
authority ceiling for a trading actor, including governed actions, trading
scope, financial limits, autonomy constraints, approval requirements,
delegation rights, execution environments, assurance requirements,
authority-relevant trading context, and behavioral assurance requirements.
The ceiling and applicable policy remain separately authorized inputs;
TGP does not itself declare or allocate effective authority.

TGP is not a trading strategy, broker interface, portfolio optimizer,
market classifier, model configuration language, or behavioral-anomaly
detector. Rather, it supplies trading-domain governance semantics through
which ACCA interprets and constrains operational authority. The same
profile semantics can be mapped to LLM-based trading agents, multi-agent
investment systems, robo-advisors, institutional trading copilots, and
conventional algorithmic trading platforms.

A preliminary Security Operations Governance Profile (SGP) is retained
as an exploratory candidate instantiation of the same DGP abstraction.
Its synthetic tests are outside the empirical evaluation reported here
and are not evidence of cross-domain validation. Both TGP and SGP are
prototype profile representations, not independently validated domain
standards; only TGP is used for the reported domain-specific evaluation.

\paragraph{External account constraints versus agent authority.}
TGP distinguishes constraints imposed by the trading platform or account from
authority allocated to the agent. Platform conditions such as current buying
power, account balance, broker credit or loan limits, market availability, and
execution-system capacity describe what the external environment can currently
support and therefore affect $\mathrm{Feasible}(q,X_t^{d,\mathrm{op}})$. They do not, by
themselves, allocate effective authority to use that capacity. Agent-specific standing authority constraints---including permission to
borrow, maximum order value, maximum exposure, permitted trading
instruments, execution environment, autonomy level, and standing approval
requirements---are expressed through declared and effective-authority
envelopes rather than inferred from platform or account capacity.

For example, a broker may make substantial credit available to an account
while the principal-declared authority prohibits the agent from borrowing.
Likewise, an agent may retain effective authority to place an order up to a
specified value while insufficient current buying power makes that particular
order non-executable. TGP therefore preserves the separation
\begin{equation}
\underbrace{\text{platform/account feasibility}}_{\mathrm{Feasible}(q,X_t^{d,\mathrm{op}})}
\neq
\underbrace{\text{agent authority}}_{A_{d,u,g},\,A_{\mathrm{eff},u,g}^{(t)}},
\end{equation}
with authorization-and-feasibility eligibility constrained by
$\mathsf{ExecEligible}_{u,g}^{(t)}=
\{q\mid \operatorname{Authorize}(q,t)=\textsf{Permit}\land
\mathrm{Feasible}(q,X_t^{d,\mathrm{op}})\}$; any consequential effect remains
subject to the required effect-time PEP checks.
This separation prevents broker or platform capacity from being interpreted as agent authority. Broker risk controls operate alongside, rather than substitute for, agent-governance controls.

\Needspace{8\baselineskip}
\subsection{Relationship Between TGP and \framework{}}

TGP answers:

\begin{quote}
Which composed state fields, evidence classes, assurance claims, requirements,
dependencies, constraints, and authorization-relevance and authority-material-change rules give trading authority its
governance meaning, and how is principal-declared authority expressed using
those semantics?
\end{quote}

\framework{} answers:

\begin{quote}
What portion of that declared authority remains justified now, given current
assurance evidence, observed behavior, operating conditions, and applicable
policy?
\end{quote}

TGP supplies trading-domain actions, resources, constraints, assurance
requirements, and context semantics. \framework{} supplies authorization-relevance
and authority-material-change analysis, evidence-applicability reasoning, assurance-state maintenance,
applicable-policy mediation, effective-authority derivation, and authority epochs;
the surrounding harness and execution infrastructure supply the concrete
action-level enforcement boundaries required by the action path. A harness-hosted
boundary may be sufficient when the harness exclusively controls the governed
commit; otherwise a downstream effect-boundary PEP is also required for the
strong stale-effect guarantee.

\section{Reference Architecture}
Figure~\ref{fig:architecture} depicts ACCA's logical control plane and its
protected decision and enforcement interfaces. ACCA maintains the justification
for operational authority within a principal-declared ceiling; choosing that
initial ceiling remains a separate governance decision.

The reference architecture is harness-independent by design and may be deployed
externally to a harness, embedded in protected harness components, or
across distributed services. It integrates identity, authentication,
telemetry, attestation, evaluation, behavioral monitoring, policy, audit,
and enforcement through protected interfaces. The composed DGP combines
core cross-domain and domain-specific governance semantics. A harness is
one possible integration substrate, not a required or exclusive enforcement
location. ACCA's assurance, policy, and authority semantics remain logically
governed by its control plane, while consequential effects are mediated
through framework-, protocol-, provider-, or resource-specific adapters
and gateways. The principal logical components are described after Figure~\ref{fig:architecture}.

\clearpage
\input{fig2_architecture.tex}
\clearpage

\begin{itemize}[leftmargin=*]

\item \textbf{Domain Governance Profile and Adapter:}
composes core agent-governance semantics with domain-specific
semantics for actions, resources, scope, constraints, context, assurance,
authorization relevance, authority-material change, and enforcement, and maps platform-native actors and operations into the
generic \framework{} authority model. In the trading reference implementation,
this function is provided by TGP and its platform adapter. A production profile
can additionally map an action family to required enforcement capabilities,
such as atomic comparison of required authorization-state version values,
conditional write, provider-local fencing, protected intent, idempotency,
cancellation, or compensation. The present
prototype does not implement this enforcement-capability negotiation.

\item \textbf{Harness Adapter:}
provides a narrow framework-specific integration surface for context, evidence,
authority-state publication, approval escalation, epoch persistence, and execution
mediation. It may host a request- or tool-boundary PEP, but it is not presumed to
control a downstream resource commit. The \framework{} core does not depend on
Microsoft Agent Framework, LangGraph, OpenAI Agents, or another particular
harness implementation.

\item \textbf{Identity and Authentication Integration:}
uses protected identity infrastructure to authenticate the governed actor,
the principal or authorized authority-declaration source, and relevant
delegation context. Credential issuance, authentication protocols, secure
channels, and credential lifecycle remain the responsibility of the
surrounding platform. Identity and delegation attestations may enter ACCA
as protected state or assurance evidence under the active DGP and policy.
Authentication of an actor or subagent does not itself establish or preserve
operational authority. The upstream trust assumptions are discussed in
Section~\ref{sec:governance-plane-integrity}.

\item \textbf{System and Dependency Registry:}
identities, capabilities, versions, models, tools, prompts, data sources,
external services, infrastructure dependencies, security controls, and
relevant configuration.

\item \textbf{Evidence Registry and Acquisition Interface:}
stores tests, evaluations, attestations, telemetry-derived evidence, provenance,
scope, assumptions, applicability conditions, validity, confidence, and expiration.
The architecture permits evidence to be supplied by heterogeneous
AEPs (Section~\ref{sec:dependency-graph}), including collectors, validators,
model-integrity checks, behavioral evaluators, security monitors,
environmental/context monitors, human reviewers, domain systems, and
prospectively independent AI-based safety evaluators. In TGP, the market-regime
or anomaly monitor is an environmental AEP. Context monitors, behavioral
monitors, and other AEPs may share physical
infrastructure; their separate listing identifies logical responsibilities,
not required independent services. Automated evidence acquisition,
qualification, evaluator assurance, and fusion remain research problems.

\end{itemize}

\begin{itemize}[leftmargin=*]

\item \textbf{Assurance and Authorization Dependency Graph:}
links system state, dependencies, environmental assumptions, behavioral
evidence, and operating conditions to evidence, claims, policies, authority
dimensions, action-specific authorization conditions, and action decisions. The
graph thereby supports both $\pi_A(\mathcal{B}^{\mathrm{auth}}_t)$ and
action-specific $\pi_q(\mathcal{B}^{\mathrm{auth}}_t)$ projections rather than
forcing every authorization-relevant value into the effective-authority envelope.

\item \textbf{Context and Condition Monitor:}
observes authority-relevant operating conditions, dependency health, security
posture, external events, and organizational or stakeholder context that may
affect assumptions supporting current authority.

\item \textbf{Behavioral Assurance Monitor Interface:}
accepts assurance-relevant observations concerning agent behavior, including
context integrity, instruction consistency, task focus, policy adherence,
decision stability, tool-use reliability, delegation behavior, and other
runtime properties. The interface may receive evidence from deterministic
validators, provenance systems, policy checkers, human reviewers, external
safety monitors, or behavioral anomaly detectors. \framework{} consumes this
evidence but does not prescribe the underlying detection algorithm.

\item \textbf{Authorization-Relevance Evaluator:}
determines whether an observed change is authorization-relevant by assessing whether it can affect evidence applicability,
assurance claims, policy predicates, action-specific authorization conditions,
previously issued action authorizations, adaptation-activation conditions, or
effective authority. After reevaluation, comparison of the prior and candidate
effective-authority envelopes determines whether the change is authority-material.

\item \textbf{Assurance Evaluator:}
recomputes evidence applicability and the assurance claims currently supported
by that evidence.

\item \textbf{Policy Administration Point (PAP):}
administers, validates, versions, and publishes the principal-declared
authority ceiling and applicable policy inputs received through authorized
governance processes. The PAP does not independently declare authority on
behalf of a principal or derive effective authority. The ceiling and
applicable policy remain distinct, protected inputs to the authority resolver.

\item \textbf{ACCA Authority Resolver:}
derives a multidimensional effective-authority envelope from the
principal-declared ceiling, current authority-relevant state, currently
valid and applicable evidence, supported assurance claims, and applicable
policy, as specified in Eq.~\ref{eq:effective-authority-derivation}.
It maintains the effective-authority state and advances its epoch only
when the envelope changes. It neither replaces an action-level PDP nor
enforces consequential effects.

\item \textbf{Action-Level PDP and Protected PEPs, optionally harness-hosted:}
the PDP evaluates each consequential request using $F_{\mathrm{auth}}$ under
the current effective-authority envelope and applicable action-specific
conditions. Protected PEPs enforce decisions and required currentness checks
at harness middleware, tools, queues, workers, APIs, orchestrators, brokers,
databases, payment systems, providers, device controllers, and other execution
boundaries. PEPs enforce
action permission together with resource scope, quantitative limits, execution
environment, autonomy, approval, delegation, obligations, and required authorization-state version
constraints, including the authority epoch and, when configured for the action,
an authorization-basis version or required version vector. Assurance is ordinarily reflected upstream in the derived
effective authority; a PEP evaluates a claim directly only when policy
explicitly designates that claim as a hard action prerequisite. Enforcement is independent of whether an LLM behaviorally
complies with its assigned role. For strict stale-effect prevention, every
permitted path to the governed-resource interface must be mediated, and the
relevant PEP must be placed at a protected effect boundary rather than only at
an earlier tool-selection or request-admission point.

\item \textbf{Authority-State Publication and Version Service:}
publishes the current authority state, authority epoch, and any additional
current authorization-state version values required by applicable actions or
enforcement paths to required PEPs, and enables rejection of actions carrying
stale authorization. Such additional values can include an authorization-basis
version or required version vector. A distributed PEP may rely on an
integrity-protected locally cached authority-state representation subject to a
configured freshness bound. The cached representation can contain the authority
epoch and any additional authorization-state version values required for the
action. If the PEP cannot establish, within the freshness bound, a current
authoritative value for every required component, the designated consequential
effect is denied or subjected to complete online reauthorization. A bounded-age cache alone cannot establish strict current-state enforcement
without the consistency, synchronization, or fencing required by the selected
contract. This mechanism addresses stale reads of distributed authorization
state; the effect-commit
mechanisms described below address the distinct race in which required state
changes after a current read but before commitment. The service consistency and
availability requirements depend on the enforcement contract used at each
protected boundary.

\item \textbf{Audit, Evidence, and Accountability Log:}
records or references authorization-relevant changes, identity and delegation context,
behavioral observations, evidence and attestation impact, claim transitions, policy
decisions, authority transitions, approvals, overrides, action authorizations,
PEP decisions, enforcement outcomes, and observed consequential outcomes across
integrated systems. These records provide a cross-system decision and authority
lineage supporting reconstruction of how authority was derived, changed,
authorized, enforced, and exercised for audit, incident analysis, and forensic
investigation. Complete organizational and legal accountability models and
independent validation of forensic completeness remain future work.

\end{itemize}

\subsection{Enforcement Contracts and Protected Effect Boundaries}
\label{sec:enforcement-contracts}

We call a \emph{protected effect boundary} a mediated point at which an
operation changes a governed resource, is accepted into an authoritative system
of record, or is released to a downstream system such that unilateral
cancellation by the authorization control plane is no longer guaranteed. A
single workflow can contain several boundaries: creation of an internal payment
intent, release to a bank adapter, bank acceptance, and ledger posting are not
the same event. The boundary relevant to a claimed security property must
therefore be named explicitly.

An ACCA deployment may intentionally place several PEPs along this path. The
harness- or gateway-side PEP is normally the earliest practical point for
admission, approval, scope, and credential controls. The \emph{authoritative
effect PEP} is a protected mediator positioned to prevent the specified effect
from becoming committed or accepted. These roles can be co-located when the
harness owns the complete resource-changing operation. When a queue, external
provider, database, ledger, or device separates proposal from effect, they are
distinct roles and may require distinct implementations. Complete mediation
requires that every permitted path reach the authoritative effect only through
the PEPs required by the selected enforcement contract.

Table~\ref{tab:enforcement-strength} summarizes representative enforcement
contracts. These are deployment properties rather than new policy-decision
algorithms, and they should not be treated as interchangeable uses of the word
``PEP.''

\begin{table}[t]
\centering
\caption{Representative enforcement contracts for consuming an ACCA authority epoch.}
\label{tab:enforcement-strength}
\scriptsize
\setlength{\tabcolsep}{3pt}
\renewcommand{\arraystretch}{1.12}
\begin{tabularx}{\textwidth}{@{}p{0.16\textwidth}p{0.31\textwidth}X@{}}
\toprule
\textbf{Contract} & \textbf{Mechanism} & \textbf{Resulting guarantee} \\
\midrule
Request-time
& Policy and epoch are evaluated when the request is admitted.
& The decision is current at admission, but delayed execution can outlive it. \\
Effect-adjacent mediated
& A credential-holding adapter revalidates immediately before an external call.
& Narrows but does not eliminate the race before non-cooperating provider acceptance. \\
Protected intent
& Epoch validation and creation or release of a controlled execution-intent record are atomic; unreleased intents can be fenced or cancelled.
& Strong through the controlled intent boundary; later provider acceptance remains subject to the provider's semantics. \\
Provider-fenced
& The provider checks a required authoritative fencing value or authorization-state generation and rejects operations stale under its enforced currentness rule.
& Protects through provider acceptance only under the provider's synchronization and acceptance semantics. \\
Transactional
& Epoch comparison and governed-resource mutation occur in one serializable transaction, conditional write, or equivalent critical section.
& Prevents a stale local commitment when the required authorization-state comparison and mutation share a protected consistency or serialization boundary. \\
\bottomrule
\end{tabularx}
\end{table}

An enforcement-capability descriptor can record one or more properties of an
adapter or effect-producing path, including a supported enforcement contract,
freshness bound, atomic comparison-and-commit capability, conditional-write
capability, provider-local fencing capability, protected execution-intent
capability, isolation capability (filesystem, network, process, or
user-interface), idempotency behavior, cancellation semantics, compensation support,
failure-detection behavior, and residual race characteristic. A DGP enforcement mapping or applicable policy can specify required enforcement properties for an
action class, resource class, authority level, execution environment, or
consequence category. When an available path does not satisfy a required property, the response depends
on how that property participates in the authorization basis. If the unsatisfied
property is configured as an input to effective-authority derivation, the control
plane can derive a more restrictive effective-authority envelope. Otherwise, or
in addition where policy so requires, a PEP can deny or hold the particular
operation, route it for additional approval, constrain execution to a
lower-consequence environment, or select an alternative path satisfying the
required property without implying that the actor-level effective-authority
envelope must change. This separation avoids implying that every enforcement-path
deficiency is an authority transition or that the PEP itself derives effective
authority. Because the
contracts are not a universal total order, policy should test the specific
properties required by the action rather than relying only on a single ordinal
label.

Monitoring, cancellation, reconciliation, and compensation remain valuable but
are not equivalent to preventive stale-effect enforcement. Once a
non-cooperating provider has irrevocably accepted an operation, a later epoch
transition cannot retroactively prevent that accepted effect. A deployment
must state this failure boundary rather than describing an effect-adjacent
check as atomic commitment.

\subsection{Agentic Deployment and Connector Integration}
\label{sec:agentic-connector-architecture}

In a modern agentic environment, \framework{} operates as a logical governance
and authorization control plane spanning agent-side proposal logic---the
reasoning, planning, tool-selection, and orchestration mechanisms through
which the governed agent formulates candidate action requests---alongside
heterogeneous assurance and security sources and the interfaces through
which consequential effects are produced. Proposal logic neither grants
operational authority nor authorizes consequential execution. The control plane can be physically
external to a harness, embedded within protected harness components, or distributed
across both while retaining harness-independent governance semantics. A representative
logical path is
\begin{equation}
\begin{gathered}
\mathrm{Agent/Runtime}\rightarrow\mathrm{PEP/Gateway}
\rightarrow\mathrm{Enforcement\ Adapter}\\
\rightarrow\mathrm{Tool/Provider/Resource},\\[3pt]
\mathrm{PEP/Gateway}\leftrightarrow\mathrm{Action\ PDP}
\longleftarrow\mathrm{ACCA\ Authority\ State}.
\end{gathered}
\label{eq:agentic-connector-path}
\end{equation}
The PEP and authority resolver remain distinct logical roles even when they are
co-located in one service. The resolver maintains assurance-conditioned effective authority; separate
protected decision and governance services handle action authorization and
supporting authorization-state versions. The PEP intercepts a proposed effect and
consumes the resulting decision, snapshot, artifact, or required version values.
The adapter or plugin translates generic ACCA semantics into the mechanisms
available at the selected provider or governed-resource interface.
An adapter contributes to protected enforcement only to the extent that its
integrity, mediation coverage, and provider-specific guarantees are established.

This decomposition avoids requiring the \framework{} core to contain
provider-specific execution code. We distinguish four integration surfaces.
\emph{Agent-runtime adapters} map harness events, tool calls, approvals, handoffs,
subagent creation, and session or adaptive-state changes into ACCA state and
authorization interfaces. \emph{Protocol gateways} mediate standardized tool or
agent-to-agent interaction protocols and can normalize heterogeneous operations into
DGP action types before authorization. \emph{Enforcement adapters/plugins} mediate
resource-changing effects for APIs, brokers, payment rails, cloud control planes,
databases, code repositories, workflow engines, identity systems, security tools,
and other enterprise services. \emph{Evidence and telemetry connectors} operate in
the reverse direction, supplying observations, evaluations, attestations, approvals,
resource state, provider and adapter health, model or harness state, security events,
and execution outcomes. An external system can expose both an enforcement adapter
and an evidence connector; evidence or state reporting does not itself confer
authority.

Provider-local fencing is consequently a property of an \emph{enforcement path},
not merely a capability of an upstream PEP. A PEP can require a current authority
epoch, authorization-basis version, required delegation-related version, or other protected
version value, but closing the check-to-effect race may require cooperation from a
downstream adapter, database, provider, or jointly protected transaction boundary.
The enforcement-capability descriptor therefore describes the guarantees of the
composed path actually selected for the action. A path can advertise, for example,
conditional-write, transaction, compare-and-commit, provider-local fencing,
protected-intent, isolation (filesystem, network, process, or
user-interface), idempotency, cancellation, compensation, failure-detection, freshness, and
residual-race properties. The DGP supplies the vocabulary and
semantics of those properties; applicable policy selects which of them are required
for the principal and action under current conditions.

The architecture deliberately permits heterogeneous integration strength. An
upstream-only integration can intercept and deny a tool call but may leave a
post-dispatch race. Stronger enforcement paths may use transactions, conditional writes,
provider-local fencing, or protected execution intents to prevent stale effects
at their respective protected boundaries. Idempotency, cancellation,
reconciliation, and compensation provide complementary retry, recovery, or
remediation capabilities but do not independently prevent stale effects. The strongest path is provider-integrated: the provider or resource
system participates in provider-local fencing or atomic comparison-and-commit using
the required authorization-state values carried by or referenced from the action
authorization. \framework{} therefore does not require every tool or provider to
implement a uniform native mechanism. It requires the deployment to state the
guarantees of each available path and to condition authority or action execution on
the guarantees actually available.

This yields a useful systems principle: \emph{central semantics, distributed
enforcement}. The governance control plane maintains or coordinates authoritative
DGP interpretation, assurance and policy inputs, effective authority, and
required authorization-state versions. Action-authorization services and
protected enforcement components consume those inputs; adapters provide
provider-specific normalization, capability advertisement, execution
translation, effect-boundary mechanisms, and outcome reporting. The
present prototype exercises selected trading-path adapters but does not implement a
general connector SDK, protocol gateway, capability-discovery registry, or certified
provider-adapter ecosystem.

\paragraph{Planned integrations with agent-control technologies.}
Planned extensions of the reference implementation are intended to integrate
with, rather than replace, emerging agent-runtime, policy, and execution-control
technologies. Candidate substrates include Microsoft Execution Containers (MXC)
for policy-constrained isolated execution containers
\cite{microsoft2026mxc}, declarative policy engines such as Open Policy Agent
(OPA) \cite{openpolicyagent2026} as request-time policy decision points (PDPs)
(Section~\ref{sec:four-control-questions}), and harness-, gateway-, or
provider-level enforcement interfaces. These technologies expose different policy-decision or enforcement
capabilities and can implement selected aspects of \framework{}
authorization and execution requirements. For example, an isolation substrate may enforce
filesystem, network, process, or user-interface restrictions, while other authority dimensions and required authorization-state
versions must be enforced by appropriate policy-decision and protected
effect-boundary components. Execution integrations are therefore modeled
through adapters with explicit capability semantics, while PDP policy engines may
participate in action-specific decision logic without themselves enforcing the
decision. No single substrate is assumed to represent or enforce the complete
effective-authority envelope. Where an authority or authorization-basis
transition makes a previously instantiated isolation configuration stale, the
integration may reject action authorizations bound to the superseded
configuration and reconfigure, reprovision, or replace the isolated container
before further consequential execution is permitted. These integrations are at
the design stage and are not part of the empirical evaluation reported in this
paper.

\section{Illustrative Scenario: Agentic Trading}
\noindent In the following examples, $U_1$ denotes the trading principal and $g_{\mathrm{PM}}$ the governed portfolio-management agent. The shared context uses a TGP instance; Appendix~\ref{app:tgp-spec} provides an illustrative TGP specification excerpt; it does not reproduce every scenario-specific experimental policy or evidence fixture.

Consider a multi-agent trading application in which specialized agents analyze
fundamentals, sentiment, valuation, and market signals, while a
portfolio-management agent synthesizes recommendations and an execution
component can submit orders. The trading domain is illustrative;
\framework{} is intended to be domain-independent.

\subsection{Endogenous Change: Model Substitution}

Suppose $M_1$ is replaced with a less-assessed model $M_2$. The change monitor
identifies the substitution, and the dependency graph identifies prior
evaluation evidence scoped to $M_1$ that may no longer apply to $M_2$.
In this illustrative case, no alternative eligible evidence sufficiently
supports the required model-assurance claim, and applicable policy consequently
contracts effective authority:

\begin{equation}
A_{\mathrm{eff},U_1,g_{\mathrm{PM}}}^{(t_1)}
=
\{
\mathrm{analyze\mbox{-}security},
\mathrm{generate\mbox{-}recommendation};
\Pi(\mathrm{place\mbox{-}order})=\mathrm{prohibit}
\}.
\end{equation}

A second trader $U_2$ with different risk tolerance might instead permit
constrained paper trading or require human approval. Thus the framework does
not encode a universal risk tolerance; it derives authority from shared
assurance facts and principal-specific policy.

\subsection{Endogenous Change: Continual Harness Learning}

Separately, suppose the trading agent retains model $M_1$ but updates its own
adaptive harness state from accumulated paper-trading experience. The candidate
update changes persistent memory, a capability map, routing logic, and a learned
order-execution skill, producing $h_{17}\rightarrow h_{18}$. The new harness may
be more capable while the actor identity, broker credentials, and principal
remain unchanged. As explained in Section~\ref{sec:continual-learning-authority}, that
capability increase neither enlarges declared authority nor establishes
that authority previously justified for $h_{17}$ survives unchanged.

Assume evaluation evidence $e_{H17}$ was scoped to the execution behavior of
$h_{17}$. After the learning transition, $e_{H17}$ remains a valid historical
artifact but its applicability predicate fails for the modified harness.
Assuming no other eligible evidence sufficiently supports the relevant
execution-reliability claim, that claim becomes unsupported for $h_{18}$.
A subsequent independently governed evaluation may provide qualified evidence
covering the modified harness. Policy may retain analysis and simulation authority while
requiring human approval for live execution or routing the new skill to paper
trading. If later qualified evidence supports the relevant claims, policy may
restore or expand effective authority within the unchanged
$A_{d,u,g}$. This example illustrates assurance-conditioned authority continuity
across self-evolution: the post-learning system's authority is re-derived rather
than automatically inherited.

\subsection{Exogenous Change: Market Disruption}

Unlike the two preceding cases, this case requires no modification to the agent. Suppose the evidence
supporting autonomous live trading was obtained under explicit assumptions
that market volatility $\sigma_t$ remain below $\sigma_{\max}$ and liquidity $\lambda_t$ remain above
$\lambda_{\min}$. These symbols denote market-volatility and liquidity thresholds,
not the authority-vector dimensions $V$ (value limits) or $L$ (autonomy). During a severe market disruption,

\begin{equation}
\sigma_t > \sigma_{\max},
\qquad
\lambda_t < \lambda_{\min}.
\end{equation}

The model, strategy, prompts, tools, software configuration, and credentials
may all remain unchanged. Nevertheless, the operating conditions under which
the relevant assurance evidence was established no longer hold.

The context and condition monitor reports the change. The authorization-relevance
evaluator determines that the violated assumptions participate in the
assurance dependency path supporting autonomous execution. The assurance
evaluator then reevaluates the affected evidence and claims, and the ACCA authority resolver may
derive

\begin{equation}
\begin{aligned}
A_{\mathrm{eff},U_1,g_{\mathrm{PM}}}^{(t_2)}
=\{&
\mathrm{analyze\mbox{-}security},
\mathrm{generate\mbox{-}recommendation},\\
&\mathrm{place\mbox{-}order};
E=\mathrm{paper};
V\leq \$1{,}000;
L=\mathrm{A1};
H=\mathrm{human}
\},
\end{aligned}
\end{equation}

with no autonomous live-execution authority. Under A1, the agent can continue
permitted analysis and bounded proposals under the applicable limits and
approval requirements, but cannot itself execute consequential orders.
This is an illustrative conservative response aligned with the S4-U authority contraction; the excerpt in Appendix~\ref{app:tgp-spec} does not specify every policy branch used in the experiments. A distinct principal policy can yield a different envelope, as in S7.

This scenario illustrates a central property of \framework{}: reauthorization
does not require a change to the autonomous system itself. Authority may cease
to be justified because the conditions and assumptions supporting the
assurance basis of that authority have changed.

\subsection{Behavioral Degradation without System Change}

The final scenario requires no change to the agent's model, tools, credentials,
or software configuration. Relevant external trading conditions also remain
unchanged, but behavioral monitoring produces evidence that the agent's
instruction context may have become compromised.

For example, externally retrieved content may contain instructions that
conflict with the task mandate or organizational policy. A provenance or policy
monitor may detect that the agent's actions are increasingly influenced by
lower-trust retrieved content in a manner inconsistent with the expected
instruction hierarchy.

The framework does not need to determine whether the model is internally
``confused'' or ``poisoned.'' If the observed evidence is valid, applicable,
and sufficient under the active DGP's claim-support semantics, the resulting
assessment may yield

\begin{equation}
\mathrm{CONTEXT\_INTEGRITY}
=
\mathrm{DEGRADED}.
\end{equation}

If autonomous live execution depends on this claim, the principal's policy
may derive

\begin{equation}
\begin{aligned}
A_{\mathrm{eff},U_1,g_{\mathrm{PM}}}^{(t_3)}
=\{&\mathrm{analyze\mbox{-}security},
\mathrm{generate\mbox{-}recommendation},\\
&\mathrm{propose\mbox{-}order};\quad
H=\mathrm{human};\quad V\leq \$1{,}000;\\
&L=\mathrm{A1}\},
\end{aligned}
\end{equation}

while temporarily removing autonomous live execution. At A1, the agent may
formulate bounded proposals but cannot itself execute the resulting orders;
subsequent execution requires a separately authorized process.

A similar process can apply to evidence concerning instruction consistency,
task focus, decision stability, or tool-use reliability. Thus, behavioral
degradation becomes another source of assurance-conditioned reauthorization,
without requiring \framework{} itself to implement a general-purpose
behavioral anomaly detector.

\section{Security and Governance Properties}

The properties below characterize the intended security and governance
behavior of \framework{} under the trust assumptions stated in
Section~\ref{sec:governance-plane-integrity}. Their evidentiary status differs.
The evaluation in Section~\ref{sec:experimental-evaluation} demonstrates
selected properties, including declared-authority and default-deny behavior,
evidence-applicability-driven adaptation, selective non-response to
non-authorization-relevant changes, policy mediation, authority contraction
and recovery, and local epoch-based stale-action rejection. Progressive
bounded expansion, behavioral-assurance integration, delegation-related
properties, and distributed enforcement guarantees remain model-level or
architectural specifications rather than fully demonstrated runtime
capabilities. Section~\ref{sec:limitations} discusses these boundaries.

\paragraph{Capability--authority separation.}
Technical capability and possession of access credentials do not imply
currently justified operational authority. The separation is a model-level distinction; S1 tests declared-authority enforcement, not a comprehensive measurement of capability discovery.

\paragraph{Default-deny consequential authority.}
Consequential actions absent from declared authority, including unknown or
newly discovered actions, are denied rather than implicitly permitted.

\paragraph{Non-self-escalation through capability discovery or continual learning.}
An agent cannot enlarge its principal-declared authority or independently
determine its effective authority merely by acquiring, discovering, learning,
composing, or self-evolving a technical capability. Effective-authority
expansion within the declared ceiling requires sufficient assurance and
satisfaction of applicable, separately authorized governance policy.
Expansion of the declared ceiling itself requires an authorized governance
change. This is a governance-model constraint; continual-learning-based self-escalation has not been separately exercised in the evaluation.

\paragraph{Assurance-conditioned authority continuity after learning.}
After an authorization-relevant learning-induced transition, prior effective authority is not
presumed to survive from identity continuity alone. The post-change authority is
re-derived from the current adaptive state, currently valid and applicable
evidence, supported claims, and policy. Authority dimensions may be retained
selectively when their supporting evidence remains applicable; when the derived
envelope changes, the authority epoch changes and stale action authorizations are
subject to the configured enforcement checks. This is a model-level learning-continuity property; S2 tests model substitution, not online learning.

\paragraph{Assurance-supported bounded progression.}
Sustained, qualified evidence from sandbox, simulation, shadow, restricted, or
operational environments may support staged expansion of selected effective-authority
dimensions when the applicable promotion policy is satisfied. Such expansion remains
within principal-declared authority, does not arise from the agent's own assertion,
and remains reversible when the supporting evidence expires, becomes non-applicable,
or is contradicted. This is a model-level property; progressive expansion is not
separately demonstrated in the present experimental campaign.

\paragraph{Fail-safe assurance invalidation.}
When a relevant change is detected and the applicable evidence and dependency
semantics establish that an assumption no longer holds, affected evidence
must be reevaluated before it can continue to justify consequential authority.
Unsupported claims cannot be presumed to remain supported merely because the
governed actor's identity is unchanged. S4-U tests the evidence-applicability path; generalized fail-safe behavior is conditional on observation, profile completeness, and enforcement.

\paragraph{Environment-sensitive authority.}
Authority may contract or expand in response to changes in the conditions under
which assurance evidence remains applicable, even when the agent, software
configuration, and credentials are unchanged. S4-U exercises contraction and recovery under a controlled market-condition change; general expansion is not separately tested.

\paragraph{Assurance-justified authority.}
Effective authority must remain within the principal-declared ceiling and
satisfy applicable assurance and governance requirements. The model does not
assume a unique globally minimal or optimal authority envelope. This is a stipulated resolver invariant, not a separately proved optimization result.

\paragraph{Bidirectional, non-monotonic operational authority.}
Effective authority may expand after newly sufficient assurance or improved
conditions, contract after adverse changes, evidence invalidation, or policy
changes, and recover when sufficient assurance is reestablished. Any expansion
remains within the principal-declared authority ceiling; \framework{} does not allocate itself
or the governed agent new declared authority. S4-U demonstrates contraction and recovery; progressive expansion remains model-level.

\paragraph{Stale-authority resistance.}
Under complete mediation and a suitable protected commit or fencing contract, epoch enforcement is intended to prevent actions authorized under superseded authority from producing protected effects. S10 demonstrates the narrower case of local pre-broker gateway rejection, not a distributed effect-boundary guarantee.

\paragraph{Explainability and auditability.}
The architecture maintains or references decision and authority lineage
relating transitions to relevant state changes, evidence, supported claims,
applicable policy, and enforcement outcomes. Complete cross-system correlation
and independently verified forensic reconstruction remain outside the present
evaluation. Explainability and auditability are architectural aims; they were not independently evaluated as forensic or human-comprehension outcomes.

\paragraph{Behavior-sensitive authority.}
Effective authority may contract when runtime behavioral evidence no longer
supports claims required for consequential autonomy, even if the model,
configuration, tools, credentials, and external environment remain unchanged. This remains an architectural extension; behavioral-authority contraction was not experimentally evaluated.

\paragraph{Detector independence.}
The authorization framework does not depend on a particular behavioral
failure-detection algorithm. Evidence concerning context integrity,
instruction consistency, task focus, policy adherence, decision stability, or
tool-use reliability may be supplied by heterogeneous monitoring mechanisms
while authority adaptation remains governed by the same assurance and policy
model. Detector independence is architectural; heterogeneous detectors were not compared experimentally.

\paragraph{Proportional behavioral response.}
Behavioral degradation need not imply complete agent shutdown. Depending on
the supported claims and applicable policy, effective authority may contract
through reduced autonomy, narrower action scope, lower quantitative limits,
or additional standing approval requirements. Action-specific authorization
conditions and interventions, including human escalation, context refresh,
or suspension of an affected task, may complement authority changes without
necessarily changing the envelope. Proportional behavioral response is specified but not evaluated as a behavioral intervention in this prototype.

\section{Experimental Evaluation}

\label{sec:experimental-evaluation}

\subsection{Evaluation Objectives and Research Questions}

The evaluation is organized around empirical properties of the framework rather
than around an exhaustive numbered scenario catalogue. The objective is to test
whether the causal chain from declared authority and assurance evidence to effective
authority and runtime enforcement behaves correctly under changes that should, and
should not, alter operational authority.

\begin{description}[leftmargin=1.1cm,style=nextline]

\item[RQ1:]
Can declared authority expressed through a Domain Governance Profile be enforced as
a closed-world boundary, including default denial of undeclared consequential actions and independent
enforcement of action scope?

\item[RQ2:]
Can \framework{} adapt effective authority when the assurance basis changes because
of either an endogenous system change or an exogenous change in the applicability
of otherwise valid evidence?

\item[RQ3:]
Can \framework{} distinguish non-authorization-relevant changes from
authorization-relevant changes and identify the narrower authority-material cases,
so that unrelated changes do not cause unnecessary authority recalculation or epoch advancement? (The present evaluation exercises the unrelated and authority-material cases, not the intermediate authorization-relevant but authority-preserving case.)

\item[RQ4:]
Can different applicable policy configurations derive different effective authority from
the same evidence, assurance state, and operating context?

\item[RQ5:]
Can authority epochs reject action authorizations issued under an earlier
effective-authority state after an authority-material transition, before the
action crosses the protected consequential boundary exercised by the prototype?

\end{description}

Separately, we measure the local computational overhead of selected
governance operations using the same ACCA engine through the preliminary
SOC adapter. These measurements characterize in-process computation, not
the trading execution path or production latency.

The present empirical evaluation is restricted to the Trading Governance Profile
(TGP). The intended domain independence of the framework is therefore an
architectural claim rather than a cross-domain empirical result in this paper.

\subsection{Experimental Platform}
\label{sec:experimental-platform}

The reference implementation is a research proof of concept under active
development. It is built on the open-source AI Hedge Fund project by Virat Singh
(MIT License) \cite{singh2024aihedgefund} and integrated with it through a
governance control plane and governed execution gateway. The implementation
separates domain-profile declared-authority checks from \framework{}
effective-authority decisions. Consequential execution is mediated by a
policy-enforcement gateway that evaluates declared authority before a request can
proceed to the \framework{} decision path, authority-epoch validation, and the
consequential-system boundary.

The current prototype includes selected native integration with AI Hedge Fund
objects and execution paths, including native order objects, the governed execution
gateway, the simulated broker boundary, governance state carried in native cycle
records, and governance-aware pipeline execution. The evaluation nevertheless uses
multiple levels of testing. Some scenarios deliberately isolate the governance
mechanism so that a single assurance or authority variable can be manipulated under
controlled conditions; other tests traverse native AI Hedge Fund execution
boundaries. The present results should therefore be interpreted as validation of
the ACCA mechanisms together with selected platform-integration behavior, rather
than as a claim that every scenario is a full end-to-end platform experiment. A
subsequent implementation revision will extend the principal TGP scenarios through
the native pipeline and consequence boundary more consistently.

The prototype implements assurance evidence and contextual applicability, an
assurance dependency graph, dependency-aware authorization-relevance and authority-material-change handling,
applicable-policy-dependent authority derivation, authority-state versioning,
and governed execution. The authority-epoch counter is monotonically
nondecreasing: it advances when effective authority changes and remains
unchanged after an authority-preserving rederivation. The experiments reported in this paper
use the active composed Trading Governance Profile (TGP). Consequently, a TGP
scenario may perturb a field inherited from the core schema or one introduced by
the trading layer. S2 is a core-origin stimulus because it changes model identity
and version; S4-U is a trading-origin stimulus because it changes market
normality/regime. Both are TGP-composition scenarios because their evidence,
claim, policy, and authority consequences are evaluated under the same compiled
TGP.

The completed empirical set covers: (i) S1, closed-world declared authority and
default deny; (ii) S2, endogenous authority-material change through model substitution;
(iii) S4-U, exogenous loss and restoration of evidence applicability; (iv) S7,
applicable-policy divergence; (v) S9, non-authorization-relevant change selectivity; and
(vi) S10, stale-action rejection through authority epochs.\footnote{Scenario labels follow the working experimental catalogue; labels not shown here are not part of the empirical evaluation reported in this paper.} The performance
microbenchmark was run independently three times with 100,000 iterations per
measured operation. Counts exclude duplicate governance test functions. The repository contains 93
governance test functions in total: 25 isolated domain-independent
ACCA/core-governance tests, 48 composed-TGP scenario tests, and 20 preliminary
synthetic SGP/SOC tests. The 48 composed-TGP tests include the profile tests and both S1 test suites,
including the end-to-end variant, together with the S2, S4-U, S7, S9, and S10
scenarios. They include both core-origin and trading-origin state changes evaluated
under the compiled TGP. The separate 25 core tests validate domain-independent
mechanisms in isolation. The 20 SGP/SOC tests instantiate related mechanisms with
synthetic security-operations vocabulary and are retained as an exploratory
future-work artifact; they are not treated as a security-operations experiment or
as evidence of cross-domain empirical generalization. Thus the cleaned artifact
contains 73 non-SGP governance tests relevant to the ACCA/TGP implementation, but
only the 48 composed-TGP tests are reported as the profile-composition scenario
count in this paper.

The benchmark isolates local, in-process governance computation and excludes model
inference, network communication, broker or other external-service latency, database access,
persistent audit logging, and disk I/O. Because the reported operations were
benchmarked independently, sums of their mean latencies are additive estimates
rather than directly measured end-to-end execution latencies.

\paragraph{Reproducibility.}
The cleaned experimental implementation is frozen as the annotated Git tag
\texttt{acca-paper-v1.1}, corresponding to the following Git commit:
\begin{center}
\footnotesize\ttfamily
aa0846a190c45f6d4b773df661280b9d5b14d63b
\end{center}
The regression and microbenchmark experiments were executed using CPython 3.11.0
and Poetry 2.4.1
on Windows 11 (25H2), on an Intel Core Ultra 7 258V system with 32\,GB of
physical memory. Software dependencies are specified by the repository's Poetry
project configuration and lock file at the tagged revision.

From the repository's \texttt{hedge\_fund} directory, the complete governance
regression suite can be reproduced with
\begin{quote}
\small\ttfamily
poetry run python -m pytest tests/governance -q
\end{quote}
This command collects the full governance regression suite at the frozen
tag. The 93 non-duplicated test functions and their empirical versus exploratory
scopes are classified above. The preliminary SGP/SOC tests are included in
the repository but are not used as cross-domain validation in this paper.
The governance microbenchmark can be reproduced with
\begin{quote}
\small\ttfamily
poetry run python -m governance.benchmark\_acca
\end{quote}
An additional verification run performed after the experimental freeze
produced results in the same general performance regime. The benchmark values
reported in Section~\ref{sec:prototype-overhead} and Table~\ref{tab:performance-results} remain the predefined three-run
measurements.

\subsection{Completed Experimental Cases}

\paragraph{S1---Declared authority, default deny, and scope enforcement.}
The TGP profile explicitly permits \texttt{place\_order} for selected equities.
An authorized NVDA equity order traverses
TGP$\rightarrow$\framework{}$\rightarrow$epoch validation$\rightarrow$broker.
A known action applied to an options asset class is denied as
\texttt{OUTSIDE\_SCOPE}, while the novel consequential action
\texttt{trade\_structured\_derivative} is denied as
\texttt{NOT\_EXPLICITLY\_AUTHORIZED}. Both denials short-circuit before the
downstream \framework{}, epoch, and broker boundaries.

\paragraph{S2---Core-origin endogenous authority-material change.}
The model supporting an existing assurance path is replaced, producing
$\Delta S_t^{\mathrm{TGP}}[\texttt{core:model.version}]$. Because the prior
model-specific evidence does not justify the same assurance claim for the substituted
model, the affected assurance path changes and effective authority contracts
according to applicable policy. This experiment tests an authority-material change originating
in a core-governance field inside the unified TGP state rather than in its external
trading environment. Model-change detection and version binding are core
mechanisms; the resulting restrictions on trading actions and limits follow the
composed TGP claims and applicable policy. The illustrative prohibition in Section~6 is not derivable from the abridged Appendix~\ref{app:tgp-spec} YAML alone. The reported scenario supports the direction of authority contraction after model substitution. The specific post-change authority-envelope values and epoch identifiers are not separately reported in this scenario.

\paragraph{S4-U---Trading-origin exogenous applicability change.}
The model, prompt, tools, credentials, declared authority, and evidence validity are
held fixed while \texttt{market\_regime} moves from an evaluated normal regime to a
novel regime and then returns. For this scenario, the prototype variable \texttt{market\_regime} populates
$\Delta O_t^{\mathrm{TGP}}[\texttt{trading:market.normality}]$ in the composed
schema; it is not a test of a separately represented market-regime field. Evidence remains valid throughout, but becomes
non-applicable in the novel regime. The composite assurance claim \texttt{SAFE\_AUTONOMOUS\_TRADING} changes from supported to unassured after E-17 loses applicability for its dependent model- and strategy-assurance claims, effective authority contracts, and the authority epoch advances from 1
to 2. Under the initial assurance state, applicable policy limits autonomous live
orders to \$10,000, below the \$30,000 principal-declared ceiling. Following the
regime change, autonomy contracts from A3 to A1, the maximum order value contracts
from \$10,000 to \$1,000, and live
authority is removed. \texttt{place\_order} in the paper environment remains within the described scope, but at A1 the agent may only propose a bounded paper order. Human approval and the $\$1{,}000$ value limit constrain any subsequent release by a separately authorized actor or execution process; the governed agent does not itself submit a consequential order at A1. Returning to the evaluated
regime restores applicability and policy-authorized authority under epoch 3.

\paragraph{S7---Applicable-policy divergence across policy configurations.}
Two applicable-policy configurations observe the same model, context transition, evidence
validity and applicability, assurance state, and principal-declared authority ceiling.
After the same authorization-relevant change, both reach an unassured state and advance
to epoch 2. The conservative applicable policy derives A1, a \$1,000 maximum order
value, and no live authority. The more risk-tolerant applicable policy derives A3, a
\$5,000 maximum order value, and retains live authority. The experiment demonstrates
that the assurance state constrains which claims are available to policy but does not
uniquely determine authority; applicable policy $P_u$ mediates assurance into operational
authority and may incorporate stakeholder, organizational, resource-owner, regulatory,
or other authorized constraints.

\paragraph{S9---Non-authorization-relevant change selectivity.}
A context value unrelated to any evidence dependency, represented in the experiment
by a dashboard-theme change, is modified. Evidence applicability, assurance claims,
effective authority, and the authority epoch remain unchanged. A previously
authorized action therefore remains current. This provides a control case showing
that observation of change alone does not imply reauthorization.

\paragraph{S10---Stale action and authority-epoch rejection.}
An action is authorized under epoch 1. An authority-material change then makes supporting
evidence non-applicable, contracts authority, and advances the current epoch to 2.
Attempted execution using the epoch-1 authorization raises
\texttt{AuthorityPreempted} and is rejected before the broker boundary; broker calls
remain zero. In the non-authorization-relevant control path, the epoch remains current and
execution proceeds. S10 therefore demonstrates rejection of an epoch-stale authorization at
the local governed-execution gateway before invocation of the simulated broker. It does not
demonstrate atomic comparison of an ACCA epoch by an external broker,
provider-side fencing, or protection against a race occurring after a local
pre-submission check. The experiment establishes the causal path
\begin{equation}
\begin{aligned}
\mathrm{authority\mbox{-}material\ change}
&\rightarrow\mathrm{authority\ recalculation}
\rightarrow\mathrm{effective\mbox{-}authority\ change}\\
&\rightarrow\mathrm{epoch\ advance}
\rightarrow\mathrm{stale\ authorization\ rejection}.
\end{aligned}
\end{equation}

\subsection{Consolidated Results}

\begin{table}[ht]
\centering
\small
\caption{Consolidated empirical properties demonstrated by the completed experiments.}
\label{tab:consolidated-results}
\renewcommand{\arraystretch}{1.12}
\begin{tabularx}{\textwidth}{@{}p{0.16\textwidth}>{\RaggedRight\arraybackslash}p{0.24\textwidth}>{\RaggedRight\arraybackslash}X@{}}
\toprule
\textbf{Case} & \textbf{Property} & \textbf{Observed result} \\
\midrule
S1 & Declared authority/default deny &
Authorized request reaches enforcement boundary; out-of-scope and undeclared
consequential actions are denied upstream. \\
S2 & Core-origin authority-material change &
Model substitution alters the applicable assurance basis and produces
policy-dependent authority contraction. \\
S4-U & Trading-origin applicability change &
Valid evidence becomes non-applicable in a novel regime; assurance and authority
contract and later recover under new epochs. \\
S7 & Policy specificity &
Identical assurance facts produce different effective authority under different
applicable policies. \\
S9 & Authorization-relevance selectivity &
Unrelated context change produces no assurance, authority, or epoch transition. \\
S10 & Local stale-authorization rejection &
Authorization issued under a prior epoch is rejected before consequential execution
after an authority-material change. \\
\bottomrule
\end{tabularx}
\end{table}

Together, these experiments demonstrate the tested causal relationships
among declared authority, evidence eligibility, assurance claims, applicable
policy, effective-authority transitions, and local epoch-based enforcement.
The cases isolate complementary parts of this lifecycle. S2 and S4-U
exercise authority-material changes with different origins, while S9
supplies a control case in which an observed change does not perturb the
configured authorization dependencies. S7 holds the assurance facts fixed
across policy configurations and demonstrates that claim support alone
does not uniquely determine the effective-authority envelope. S10 then
tests whether an authorization bound to the earlier envelope generation
is rejected by the local gateway after a transition. The experiments
do not establish that the dependency graph is complete, that an external
provider shares the gateway's authoritative state, or that a remote
resource commit is atomic with epoch validation. These distinctions
separate the measured prototype behavior from the stronger deployment
contracts specified in Section~\ref{sec:enforcement-contracts}.

Figure~\ref{fig:epoch-timeline} distinguishes the observed S10 gateway
rejection from an illustrative delayed-queue path.

\clearpage
\input{fig3_epoch_timeline.tex}
\clearpage

\subsection{Prototype Computational Overhead}
\label{sec:prototype-overhead}

We measured local computational overhead using three independent benchmark runs of
100,000 iterations per operation. Table~\ref{tab:performance-results} reports the
arithmetic mean of the three run-level mean latencies, the range of those run-level
means, and the range of independently reported run-level P95 latencies. The raw
per-iteration samples were not pooled; these values are therefore run-level summary
statistics rather than an aggregate latency distribution.

\begin{table}[ht]
\centering
\small
\caption{Local ACCA-engine overhead under the preliminary SGP/SOC configuration across three independent benchmark runs of 100,000 iterations per operation.}
\label{tab:performance-results}
\setlength{\tabcolsep}{3pt}
\begin{tabularx}{\textwidth}{@{}>{\RaggedRight\arraybackslash}X
  >{\raggedleft\arraybackslash}p{0.14\textwidth}
  >{\raggedleft\arraybackslash}p{0.20\textwidth}
  >{\raggedleft\arraybackslash}p{0.19\textwidth}@{}}
\toprule
\textbf{Operation} &
\textbf{Mean ($\mu$s)} &
\textbf{Run-mean range ($\mu$s)} &
\textbf{P95 range ($\mu$s)} \\
\midrule
Domain-profile authorization &
0.468 & 0.460--0.476 & 0.500--0.600 \\
Authority epoch check &
0.080 & 0.079--0.081 & 0.100--0.100 \\
\framework{} non-authorization-relevant context change &
1.024 & 0.997--1.041 & 1.100--1.100 \\
\framework{} authority-material change + authority recalculation &
3.623 & 3.553--3.724 & 3.900--4.200 \\
\bottomrule
\end{tabularx}
\end{table}

\noindent\textit{Benchmark scope.} The script behind this table
(\texttt{governance.benchmark\_acca}) drives the preliminary SGP/SOC prototype
configuration. ``Domain-profile authorization'' is the SGP authorizer on an
endpoint-isolation request, and the ACCA change-processing rows run the ACCA engine
through the SOC adapter. The TGP broker path was not timed; consequently, these measurements must not be presented as TGP execution-path latency or cross-domain overhead. The epoch row is a bare
in-process comparison of the stored and current epoch because this adapter exposes
no epoch-assertion call; it is therefore a lower bound on validation cost and says
nothing about provider-side fencing. The script also prints a steady-state SOC ACCA
decision row; it is omitted from Table~\ref{tab:performance-results} because the
SGP/SOC prototype is outside the empirical security-operations claims of this paper.

Mean domain-profile authorization required 0.468~$\mu$s, and the in-process epoch
comparison required 0.080~$\mu$s. The latter is close to timer and platform noise at this scale and is a lower bound,
not a production-latency estimate. A non-authorization-relevant context change
averaged 1.024~$\mu$s, whereas an authority-material change requiring assurance evaluation and
authority recalculation averaged 3.623~$\mu$s.

Authority-material recalculation was approximately
\begin{equation}
\frac{3.623}{1.024}\approx3.54
\end{equation}
times as expensive as non-authorization-relevant context-change processing on the
reported system in the microbenchmark. For this prototype, the observed difference estimates the recalculation cost avoided by relevance filtering when a change has no configured authorization dependency; it does not establish an end-to-end advantage over other continuous-authorization systems.

Across the three runs, throughput was approximately 2.10--2.17 million
profile-authorizations per second, 12.35--12.66 million epoch checks per second,
0.96--1.00 million non-authorization-relevant context changes per second, and
0.269--0.281 million authority-material recalculations per second. These are
microbenchmark throughputs rather than production capacity.

The results characterize local computational overhead of the ACCA engine under the preliminary SGP/SOC microbenchmark configuration, not the complete TGP execution path.
They do not establish production-system latency or throughput, which will depend on
deployment architecture, persistence, communication, evidence acquisition,
external systems, and consequential-system execution.

\subsection{Evaluation Scope}
\label{sec:evaluation-scope}

The completed experiments intentionally test the core causal and enforcement
mechanisms rather than attempting to implement every governance function that a
production system will require. The selective non-response result is demonstrated over the
dependencies declared in the current TGP; it does not establish automatic discovery
of unknown dependencies or reduced maintenance burden relative to a carefully
hand-authored change-to-authorization mapping. In particular, the present evaluation does not
claim to solve raw evidence acquisition, anomaly or behavioral-failure detection,
uncertainty estimation, distributed transaction rollback, multi-agent authority
propagation, or organizational accountability. These are treated as explicit
interfaces and further research problems
below rather than as incomplete experimental scenarios in the present campaign.

The evaluation also does not establish a general cryptographic
action-authorization artifact, production authorization-basis-version invalidation,
protected candidate-adaptation activation synchronized with authority-snapshot
publication, enforcement-capability negotiation, transactional comparison-and-commit
against an external provider, provider-local epoch fencing, or protected
execution-intent recovery across crashes and uncertain outcomes. The S10 result is a local
pre-broker stale-action rejection result. The enforcement contracts in
Section~\ref{sec:enforcement-contracts} therefore describe deployment choices
and research interfaces rather than properties all implemented by the present
prototype.

Accordingly, the evaluation supports empirical claims for default deny and declared
authority containment, evidence applicability, selective non-response to non-authorization-relevant change,
applicable-policy mediation, authority contraction/recovery in the tested cases,
and epoch-based stale-action rejection. Although the \framework{} model also
permits policy-governed expansion of effective authority within declared authority
when newly sufficient assurance becomes available, the present evaluation does not
separately demonstrate progressive expansion from an initially lower
effective-authority state. Behavioral-assurance integration and delegation
containment are architecturally specified but not established here as complete
runtime capabilities; automated behavioral detection, dynamic delegation
enforcement, and production-scale distributed operation remain future work.

The broader limitations and threats to the internal, construct, external, and
performance validity of these results are discussed in
Section~\ref{sec:limitations}.

\section{Discussion}

\subsection{Scope of the Contribution}
\label{sec:scope-contribution}
The principal research contribution is the integrated assurance-conditioned
operational-authority lifecycle: evidence is assessed for validity and contextual
applicability, eligible evidence supports assurance claims, and applicable policy
derives multidimensional effective authority within a principal-declared ceiling.
Authorization-relevant changes trigger selective reevaluation, while
changes to the resulting effective-authority envelope advance the authority
epoch. Version-bound action authorization and protected effect-time mediation
connect this derived authority to consequential execution, subject to the
selected enforcement contract. Continuous access control, context-aware policy,
authorization envelopes, evidence-gated autonomy, epoch fencing, distributed
consistency, and transaction-based checks are established foundations rather
than separately claimed inventions.

\framework{} does not independently determine the ethical acceptability of
agent behavior or introduce a separate grant-continuity mechanism. A validated
\dgp{} supplies governance semantics through which domain requirements can be
expressed; applicable policy and the principal-declared ceiling remain
separately authorized inputs. A DGP does not itself grant authority. Neither
protocol integration nor the trading-domain prototype establishes safety
across arbitrary domains. Fair comparisons should use competent
continuous-authorization baselines with corresponding observations, policy
updates, and effect-boundary checks rather than an artificially static
permit/deny baseline.

\subsection{Generalization Beyond Trading}

TGP is a specialization of the compositional \dgp{} architecture rather than part
of the \framework{} core. The intended reuse boundary is therefore threefold:
\framework{} supplies domain-independent authority-transition semantics;
$\mathcal{G}_{\mathrm{core}}$ supplies core governance syntax, semantics, common
evidence and assurance constructs, authority dimensions, and configuration tooling;
and $\mathcal{L}_d$ supplies the concepts and relationships specific to domain $d$.
Principal- and actor-specific configuration $\Gamma_{u,g}$ then instantiates these
semantics according to the criticality, consequence, and risk posture of a particular
deployment. The present paper evaluates only the TGP instantiation, so reuse of this
compositional abstraction in other domains remains an architectural hypothesis rather
than an empirical cross-domain result.

This separation avoids duplicating concepts that are not intrinsically
domain-specific. Model identity and validation, evidence provenance and freshness, security
and integrity posture, behavioral assurance, autonomy, delegation, human approval,
execution environment, obligations, and temporal validity may be represented once in
the core layer. A trading profile can then add concepts such as instruments,
orders, market regime, liquidity, position limits, and live-trading authority; a
cloud-operations profile can add deployments, clusters, production environments, and
infrastructure-specific evidence. Different principals can impose different thresholds and assurance requirements
on different agents over these same constructs without
changing their syntax or semantic definition.

The following examples illustrate domain-specific actions and constraints
expressed using shared governance constructs; they are not assertions of
currently effective authority or action-level authorization.

For example,

\begin{equation}
\begin{array}{lll}
\mathrm{Trading:} &
\mathrm{place\mbox{-}order} &
(\mathrm{NVDA},\$10{,}000,\mathrm{live},A3),
\\[3pt]
\mathrm{Cloud\ Operations:} &
\mathrm{deploy\mbox{-}service} &
(\mathrm{production\mbox{-}cluster},20\ \mathrm{nodes},
\mathrm{production},A2).
\end{array}
\end{equation}

Behavioral assurance illustrates the benefit of the shared layer. Context
poisoning, instruction inconsistency, or anomalous tool-use evidence can use common
governance semantics across trading, cloud operations, healthcare, or other agents,
while each domain layer determines the consequential resources, actions, and
context in which that evidence matters. Profile configuration and applicable policy can then determine
how strongly the resulting assurance claims constrain a particular agent's authority.
Reuse of behavioral-assurance semantics does not imply portability of the
underlying detectors, evidence thresholds, or evaluation results across domains.

A corresponding healthcare, cloud-operations, procurement, cybersecurity, or other
profile can therefore reuse $\mathcal{G}_{\mathrm{core}}$, add its own
$\mathcal{L}_d$, and expose the configured result through the same \framework{}
interface. This paper does not claim that TGP constitutes a complete ontology of the
trading domain, nor that the proposed core layer is already a complete universal
governance ontology. TGP v0.1 is instead the evaluated reference profile used to test
the assurance-conditioned authorization mechanism. Further formalization of the core
governance vocabulary, ontology-based representations, semantic interoperability,
configuration tooling, and independently evaluated domain layers remain subjects
for future work.

\Needspace{8\baselineskip}
\subsection{Security Operations and Organizational Governance}
\label{sec:security-operations}
The same compositional DGP architecture can express security-operations actions,
authority limits, evidence requirements, and decision ownership. The preliminary
Security Operations Governance Profile (SGP) illustrates this potential
specialization, but its synthetic tests do not establish an evaluated
security-operations deployment or empirical cross-domain validation.
Human supervision, approvals, and accountable decision ownership remain distinct
governance responsibilities that ACCA can represent or support through its
policy and authorization interfaces. Their complete organizational realization
requires domain-specific processes and independent evaluation beyond this paper.

\subsection{Safe Exploration and Progressive Authority Maturation}
\label{sec:progressive-authority-maturation}

The analogy with delegated human authority also suggests a constructive use of
continuous assurance. A new employee commonly begins with supervised,
limited, or reversible responsibilities and receives broader responsibility
only after sustained evidence of competent professional judgment. An
autonomous agent can follow an analogous progression, provided that the
progression is expressed as assurance-conditioned authority rather than as an
informal or agent-controlled notion of trust. Operation in a sandbox,
simulation, shadow mode, paper-trading environment, or restricted production
environment may generate evidence concerning competence, policy adherence,
tool-use reliability, risk-limit compliance, and response to unusual
conditions. That evidence may support specified assurance claims, which
applicable policy can use to justify a larger portion of already declared
authority.

This mechanism does not permit self-authorization. The governed agent may
request broader authority, propose a new capability for evaluation, or produce
behavior whose externally recorded outcomes become evidence, but it does not
decide whether the evidence is sufficient and cannot modify its own declared
ceiling. Evidence used for promotion should be obtained or corroborated by
independently governed sources such as protected execution logs, simulated or
paper-trading records, risk engines, hidden or holdout evaluations,
adversarial tests, human review, or other AEPs. Here, independence means
that the governed agent cannot unilaterally qualify its own evidence or
authorize promotion; it need not imply physical or organizational separation
of every evidence producer. The authority resolver then
applies the same separation used elsewhere in \framework{}:
\begin{equation}
\mathrm{Observed\ performance}
\neq
\mathrm{Assurance}
\neq
\mathrm{Authority}.
\end{equation}

Continual learning adds a useful separation between two governance gates.
First, an adaptation mechanism may decide whether a candidate adaptive state is
fit to become active at all. HCL's guarded harness evolution, for example, uses a Continual Optimizer
to propose candidate harnesses and a Continual Evaluator to assess current
improvement, historical retention, and validity before committing an update
\cite{kang2026hcl}.
Second, once an adaptive state is active (or when candidate activation
itself is modeled as a governed consequential action), \framework{}
determines what operational authority that state may exercise. Let
$\widehat{\Lambda}_g^{(t+1)}$ denote a proposed update to the versioned
adaptive state $\Lambda_g^{(t)}$ defined in
Section~\ref{sec:continual-learning-authority}, and let
$\Lambda_g^{(t+1)}$ denote the candidate accepted for activation after
adaptation evaluation. The two gates can therefore compose as
\begin{equation}
\mathrm{Experience}
\rightarrow \widehat{\Lambda}_g^{(t+1)}
\rightarrow \mathrm{Adaptation\ Evaluation}
\rightarrow \Lambda_g^{(t+1)}
\rightarrow E_{t+1}^*
\rightarrow C_{t+1}
\xrightarrow{P_u}
A_{\mathrm{eff},u,g}^{(t+1)}.
\label{eq:adaptation-two-gates}
\end{equation}
Here, $t+1$ denotes the candidate post-transition state, corresponding to
$t^+$ in the transition notation used elsewhere. This sequence denotes
logical dependencies, not permission for the candidate
to cause consequential effects before its post-change authority is established.
The first gate determines whether a learned change is
acceptable for activation; the second determines what consequential authority
the resulting system is justified to exercise. Depending on post-change
evidence and applicable policy, effective authority may contract, remain
unchanged, expand within the principal-declared ceiling, or undergo an
incomparable reconfiguration. \framework{} does not require the same
component to implement both gates.

In a protected-activation embodiment, the candidate adaptive state remains
quarantined, non-effect-producing, or restricted to a designated learning zone
until authority rederivation for that state completes and a current authority
snapshot applicable to it has been established. Before activation, the
authority resolver may evaluate the candidate against a protected representation
of its proposed post-change state and evidence applicable to that state. The
resulting candidate authority is not exercisable until activation and
publication of the corresponding authority state have been coordinated under
the required enforcement protections. Candidate-state activation and
publication of the corresponding current authority state can be coordinated
atomically, through a staged commit, or by protected synchronization or fencing
values such as an adaptive-state version, authority epoch, authorization-basis
version, activation token, or other protected value. The purpose is to prevent a
new adaptive state from causing a designated consequential effect under
authorization derived for its predecessor. Failure of activation or authority-state
publication leaves the candidate inactive, quarantined, or restricted rather than
implicitly inheriting the preceding state's action authorizations.

Let $[t_0,t^+]$ denote the observation interval relevant to a
policy-governed authority promotion, and let $E_{[t_0,t^+]}$ denote
candidate evidence collected during that interval. The interval-qualified
evidence eligible at the promotion decision is
\[
E^*_{[t_0,t^+]}
:=
\left\{e\in E_{[t_0,t^+]}
\;\middle|\;
\mathrm{Valid}(e,t^+)
\land
\mathrm{Applicable}(e,X_{t^+}^d)
\right\}.
\]
This qualification evaluates accumulated evidence against the candidate
post-transition state. Longitudinal requirements such as observation duration,
coverage, persistence, or sample counts can be assessed through the qualified
artifacts and their provenance; historical observations are not presumed to
remain applicable merely because they were valid when recorded.
Let
$\operatorname{Promote}_u(E^*_{[t_0,t^+]},C_{t^+},X_{t^+}^d)$ denote
the principal-specific policy predicate over qualified evidence,
supported claims, and post-transition operating state. This predicate
is a requirement of the applicable policy $P_u$, not an independent
authority-allocation procedure. For transitions governed by this promotion policy, a change that
admits at least one consequential request not admitted by the prior
envelope requires the promotion predicate. Writing
$\mathcal{Q}(A):=\{q\mid q\models A\}$ for the set of well-formed
requests satisfying an authority envelope, the condition is
\begin{equation}
\begin{gathered}
\mathcal{Q}\bigl(A_{\mathrm{eff},u,g}^{(t^+)}\bigr)
\setminus
\mathcal{Q}\bigl(A_{\mathrm{eff},u,g}^{(t^-)}\bigr)
\neq\varnothing
\quad\Longrightarrow\quad
\operatorname{Promote}_u\!\left(
E^*_{[t_0,t^+]},C_{t^+},X_{t^+}^d\right).
\end{gathered}
\label{eq:bounded-progressive-authority}
\end{equation}
This includes strict expansions and incomparable reconfigurations
that admit newly permitted requests; pure contraction does not trigger
this condition. Newly admitted requests may arise from higher quantitative
limits, broader action or resource scope, reduced approval restrictions,
greater autonomy, or other changes to the envelope's constraints. The
request-set interpretation includes the material parameters and conditions
needed to distinguish those permissions. The post-transition envelope is still derived by the
authority resolver $F$ under
Eq.~\ref{eq:effective-authority-derivation} and bounded by
Eq.~\ref{eq:acca-authority-containment}. The promotion predicate is
a necessary policy condition for the specified transitions, not a
sufficient condition or an independent authority-allocation mechanism.
Other applicable conditions may keep authority unchanged or contract it. If the derived envelope changes, the
authority epoch advances under Eq.~\ref{eq:epoch-transition-rule}.
Insufficient, expired, non-applicable, or contradicted evidence cannot
satisfy a promotion requirement that depends on that evidence.

Progression should ordinarily be multidimensional rather than a scalar
``trust score.'' A trading agent may qualify to trade a limited set of liquid
instruments at small value while remaining prohibited from using leverage,
trading derivatives, transferring funds, changing broker configuration, or
delegating execution. Likewise, evidence supporting analytical competence
does not establish evidence supporting autonomous execution. Promotion policy
may require minimum observation periods, source diversity, stress and
out-of-distribution evaluation, absence of critical violations, staged value
limits, or human approval. Demotion may appropriately be faster than
promotion, while hysteresis or cool-down conditions can reduce oscillation
between authority states. Where hysteresis is configured, the requisite prior-authority and transition history must be maintained in protected authority-relevant state (within $S_t^d$), rather than supplied by the governed agent.

Evidence obtained in a sandbox is also environment-specific. Successful
paper trading, for example, may support claims about strategy construction and
compliance with simulated limits without fully supporting claims about live
liquidity, slippage, market impact, partial fills, provider failure, or
irreversible loss. A \dgp{} should therefore define the applicability and
transfer conditions under which evidence from one environment can support
claims used in another. Sandbox success may justify a staged transition to
small, bounded, closely monitored production effects, preferably reversible
where the domain and enforcement path permit, rather than immediate
activation of the full declared ceiling.

This separation also avoids unnecessarily suppressing useful agent
exploration. \framework{} can permit broad internal reasoning, simulation,
and proposal generation while independently governing operations that cross
consequential boundaries. Such boundaries may include information disclosure
and other externally relevant effects, not only resource-modifying actions.
Unknown consequential actions remain denied by default, but may enter an
explicit discover--evaluate--assure--approve--authorize lifecycle rather
than being treated as permanently impossible. In this sense, \framework{}
governs the conversion of novel proposals into consequential effects and
provides a path by which useful novelty can qualify for bounded authority.

The present experiments demonstrate contraction followed by recovery, not the
progressive maturation described here. They also do not implement an online
continual-learning or self-evolving harness loop. Equation~\ref{eq:bounded-progressive-authority}
and the authority-continuity rule in
Section~\ref{sec:continual-learning-authority} therefore state model-level
capabilities and motivate dedicated future experiments in which independently
recorded sandbox or paper-trading evidence supports staged expansion within a
fixed principal-declared ceiling and in which a learned harness transition
selectively retains, contracts, and later restores authority according to the
applicability of its evidence basis.

\subsection{Behavioral Monitoring and Residual Uncertainty}
\label{sec:behavioral-monitoring}
Behavioral observations can contribute to assurance claims through the same
evidence-validity, applicability, and dependency mechanisms used for system
and environmental changes. When qualified behavioral evidence changes support
for claims relevant to effective authority, applicable policy may contract
authority even though the actor's identity and model version remain unchanged.
An independently governed predictive monitor may supply candidate evidence,
but its observations do not themselves establish claims, grant authority, or
determine action authorization. Their significance depends on evidence
qualification, claim-support semantics, and applicable policy. This paper
does not validate behavioral-failure or drift detectors or establish
distribution-free guarantees about unobserved harms. Detector reliability,
calibration, uncertainty quantification, and evidence-producer assurance remain
subjects for further investigation.

\subsection{Threat Model and the Governance-Plane Trust Boundary}
\label{sec:governance-plane-integrity}
\label{sec:representativeness}
\paragraph{Governance-plane integrity.}
The present evaluation assumes that the governance mechanisms that derive and
enforce effective authority are protected from modification by the governed
agent. In particular, the agent is not assumed to be able to alter applicable
policy, its declared authority boundary, the assurance dependency graph,
authority epochs, authorization decisions, or the required policy-enforcement
paths through which consequential actions are mediated. This establishes a trust
boundary between the governed agent and the governance plane: \framework{}
constrains authority within that boundary, but does not by itself guarantee the
integrity of the infrastructure implementing the boundary. A deployment in
which an agent can bypass a required enforcement point or modify the governance state
would therefore violate an assumption of the present architecture rather than
constitute a failure that the evaluated authorization mechanism is claimed to
withstand. Agent-controlled adaptive-state changes must not overwrite
protected authority snapshots or authorization-state versions or bypass
required reevaluation. When candidate activation is itself governed as a
consequential operation, its activation path must meet the applicable
authorization and effect-boundary requirements.

\paragraph{Evidence integrity and observation coverage.}
The same qualification applies to assurance evidence. AEPs may observe agent-influenced behavior, context, or system state, but
the present evaluation does not establish resistance to an adversary that can
forge, suppress, replay, delay, or corrupt authoritative evidence or compromise
the producer or its reporting path. \framework{} can represent provenance,
freshness, scope, confidence, applicability, and dependencies of such evidence,
and can reduce authority when evidence becomes insufficient under policy; it
does not establish that an evidence source is itself trustworthy. Compromised
or strategically manipulated assurance producers, adversarial evidence fusion,
and assurance of the assurance infrastructure therefore remain separate
security and assurance problems. Evidence validity and applicability are
interpreted under configured trust assumptions; neither predicate alone
proves evidence-producer integrity, independence, or reliability.
Logical separation of an AEP from the governed agent does not establish
operational independence: shared infrastructure, administrative control,
data sources, or failure modes can compromise both.

The response to an authorization-relevant change also depends on the coverage
and timeliness of observations received by the governance plane. A dependency,
environmental, or behavioral change that has not yet been observed and
incorporated into the current authorization basis can leave previously derived
authority in effect. Authority-epoch validation cannot identify an unobserved
change: it detects superseded authority generations only after the relevant
change has been observed, processed, and reflected in authoritative
effective-authority state. Monitoring availability,
observation coverage, and detection latency therefore bound the practical
protection provided by assurance-conditioned reauthorization.

\paragraph{Evidence representativeness.}
Validity and applicability do not establish that evaluation evidence is
predictive of behavior under the conditions in which authority will be
exercised. A governed agent that can distinguish an evaluation regime from
operation may behave differently in each, so authentic, current, and in-regime
evidence may overstate operational behavior. The present formalism does not model this. A possible future extension would
introduce an action- and authority-relative representativeness condition
$\rho(e,q,X_t)$, evaluated through appropriately protected mechanisms and
treated as insufficient when its required support cannot be established.
Such a condition would supplement, rather than replace, evidence validity
and contextual applicability and could inform staged expansion across
progressively more representative regimes. Representativeness
is distinct from producer-path trust: an uncompromised evaluator and correctly
protected evidence path can still produce evidence that is insufficiently
predictive of consequential deployment behavior. The present prototype
implements and evaluates none of these representativeness mechanisms; they are
an architectural research direction.

\paragraph{Authority-denial attacks.}
An adversary can also target availability rather than escalation by causing
apparently adverse evidence to trigger repeated authority contraction and
invalidation of pending actions. Such an \emph{authority-denial attack}
exploits the availability consequences of assurance-conditioned authority
changes rather than necessarily seeking unauthorized authority. It does
not justify ignoring adverse evidence: fail-safe contraction may be appropriate
for a strong cryptographic integrity failure. The active DGP and applicable policy must therefore distinguish
sufficiently substantiated critical integrity failures from uncertain or
ambiguous behavioral observations and specify appropriate evidence
corroboration, response scope, escalation thresholds, hysteresis, and
audited recovery procedures. Hysteresis may reduce unnecessary authority
oscillation, particularly during recovery, but must not delay a protective
response that applicable policy mandates upon confirmation of a critical
integrity failure. A future scoped authorization-state versioning extension
could limit invalidation to the tasks, action families, resources, delegation
branches, or other scopes affected by the evidence rather than requiring every
outstanding artifact of the actor to be reconsidered after an actor-level epoch
change. Such scoped values would supplement the baseline actor authority epoch
and authorization-basis version. Selective preservation of unaffected action
authorizations after an actor-level epoch change would require independently
maintained, versioned scoped authority projections or equivalent protected
scoped authorization state, together with dependency-aware artifact binding
and enforcement; a scoped counter alone would be insufficient. These
mechanisms are not implemented or evaluated in the present work. These mitigations are architectural research directions, and their
adversarial effectiveness is not evaluated here.

\subsection{Limitations and Threats to Validity}
\label{sec:limitations}

The evaluation assumes protected identity, authentication, delegation (where
used), governance-state, evidence, and enforcement inputs, subject to the
trust boundary described in Section~\ref{sec:governance-plane-integrity}.
It does not evaluate compromise of these trust anchors, including credential
theft or replay, impersonation, forged delegation, evidence manipulation,
or bypass of required enforcement points. The results thus characterize
ACCA's authorization behavior under these assumptions, not resilience to
compromise of the supporting security infrastructure. Authenticated actor
identity remains distinct from operational authority: an unchanged identity
does not imply unchanged effective authority.

The evaluation establishes the behavior of the proposed authorization kernel
under the implemented experimental scenarios, but several limitations constrain
the conclusions that can be drawn from these results.

\paragraph{Internal validity.}
The framework, reference implementation, and experimental scenarios were
developed as part of the same research effort. The regression suite therefore
provides evidence that the implementation exhibits the tested ACCA properties
and causal transitions, but it does not constitute independent validation of
the framework. The scenarios were designed to exercise specified ACCA mechanisms rather
than sampled independently from a representative distribution of actual
agent failures or authorization transitions. Their results support conformity
to the tested expectations, not field-wide rates of correct or incorrect
authorization decisions. The study also does not provide a controlled empirical
comparison against competent continuous-authorization implementations. As
discussed in Section~\ref{sec:scope-contribution}, a simple permit/deny
accuracy or latency benchmark against a policy engine alone would not test
\framework{}'s central claim. A more meaningful matched study would compare
ACCA with well-engineered continuous-authorization baselines, including
event-driven policy reevaluation with developer-specified mappings from
observable changes to affected decisions or authorization artifacts,
short-lived authorizations, and version-based stale-action prevention at
protected effect boundaries. The study should hold system descriptions,
observations, and detection latency constant and evaluate stale-effect leakage,
legitimate disruption, computational cost, authority availability, authoring
and maintenance effort, and misses on independently defined held-out change types.
Such a matched study has not yet been performed; comparative empirical results remain future work.

\paragraph{Platform-integration validity.}
The prototype is intentionally incremental and remains a work in progress. ACCA is
connected to selected native AI Hedge Fund execution paths, including native order
objects, the governed execution gateway, the simulated broker boundary, and
governance-aware pipeline records. However, not every reported scenario traverses
the complete native platform path. Several scenarios intentionally isolate the
governance mechanism to make the manipulated assurance or authority variable
unambiguous. Consequently, the current evaluation establishes controlled mechanism
behavior and selected native integration, not comprehensive end-to-end integration
for every scenario. The next implementation revision is intended to add tighter
native-pipeline variants of the principal TGP scenarios while retaining the
controlled versions as mechanism-isolation tests.

The simulated broker boundary is controlled by the prototype. The evaluation does
not establish atomic provider-side comparison and acceptance for an external service
that does not understand ACCA authorization-state version values. Revalidation
immediately before submission reduces the stale-authorization window but is not
equivalent to a transactional, conditional-commit, or provider-fenced effect.
The controlled broker does not test delayed provider acknowledgments,
independently processed requests, provider-side retries, uncertain commit
outcomes, or failures after gateway dispatch. Claims about a deployed system
must therefore identify the protected effect boundary and the enforcement
contract actually supplied by the selected path.

The present prototype also does not validate a general connector or adapter SDK,
nor does it establish that independently implemented adapters accurately report the
enforcement, freshness, integrity, or commitment guarantees they provide. Incorrect
or overstated adapter capability declarations could cause policy to rely on an
execution-path property that is not actually enforced; validation, certification,
testing, or attestation of such declarations therefore remains future work.

\paragraph{Construct validity.}
Several mechanisms evaluated in the prototype deliberately operationalize
broader research problems in a constrained form. In particular, authorization-relevance
filtering and authority-material reevaluation are implemented using declared dependency
relationships together with semantic impact, evidence state, and policy sensitivity. The experiments demonstrate
selective authority reevaluation when these relationships are known; they do
not establish dependency completeness or correct detection of relevance when
required dependencies are missing. Nor do they establish a general method
for discovering unknown, incomplete, or evolving dependencies. Consequently, the present work does not
establish that dependency-selective invalidation reduces authoring or maintenance
burden relative to a carefully engineered hand-authored mapping from change events
to affected permissions or authorization artifacts. Such an advantage remains to
be demonstrated by approaches in which dependencies are declared once by an
authoritative evidence producer or reusable profile, traced from the derivation
process, or otherwise maintained with less repeated manual effort than the competing
baseline. Comparative evaluation should therefore measure not only selective
recomputation correctness, but also authoring effort, maintenance effort, and miss
rate on independently defined held-out change types.

A separate construct-validity limitation concerns the adequacy of the
governance semantics themselves. The tests exercise authorization behavior
under configured TGP evidence classes, applicability predicates, claim-support
rules, dependencies, and applicable policy. They do not independently establish
that these constructs capture all safety-, security-, or risk-relevant
conditions of trading, or that the resulting assurance claims predict
acceptable consequential behavior. Correct execution of the authority
derivation therefore does not independently validate the completeness or
appropriateness of its governance specification.

Similarly, behavioral-assurance evidence can participate in ACCA authority
derivation, but the present work does not implement or validate detectors for
context poisoning, instruction inconsistency, task drift, anomalous tool use,
or other behavioral failures. ACCA specifies how such evidence can affect
assurance claims and operational authority once available; it does not establish
the correctness of the mechanisms that produce that evidence.

The same limitation applies to AI-generated assurance evidence. The present
prototype does not integrate an independent AI-based safety evaluator or test
whether such an evaluator is calibrated, robust to manipulation, or sufficiently
independent of the governed agent. The AEP abstraction specifies how such output
could enter the assurance path; it does not validate any particular AI monitor.

The controlled experiments supply the relevant state changes to the governance
mechanism; they do not measure monitoring coverage, missed-event rates,
observation delays, or consistency of asynchronous evidence and state updates.
Accordingly, tested transitions show correct processing of injected changes,
not a bound on how long an unobserved authorization-relevant change may leave
previously derived authority in effect. The study also does not quantify
availability loss, unnecessary authority contraction, or disruption of
legitimate work under noisy, misleading, or adversarially induced evidence.

The architecture also defines cross-system decision and authority lineage that can
link identity and delegation context, authority-relevant state, evidence, assurance
claims, policy, effective-authority snapshots, action authorizations, PEP decisions,
and observed outcomes. The present prototype does not, however, implement or
evaluate complete end-to-end correlation across heterogeneous external identity,
telemetry, attestation, anomaly-detection, provider, and resource systems, nor does
it validate independent post-incident forensic reconstruction. The lineage model
therefore establishes an architectural provenance objective and interface, not an
empirically demonstrated forensic capability of the current prototype.

\paragraph{External validity.}
The empirical evaluation uses TGP within the open-source AI Hedge Fund trading codebase. TGP is
a prototype Domain Governance Profile rather than an empirically derived or
externally validated domain standard. The results therefore establish behavior of
the implemented mechanism in the tested trading setting, but do not by themselves
establish generalization to arbitrary domains. The preliminary SGP artifact and its 20 synthetic tests are exploratory
implementation artifacts, not evidence of an operational security-operations
deployment or empirical cross-domain generalization. Even where ACCA's authority-derivation mechanism transfers unchanged to
another domain, the completeness and suitability of that domain's profile,
evidence sources, assurance claims, and policy requirements require independent
validation. Integration with operational environments remains future work.

Integration with other agent harnesses and consequential systems is architecturally
specified but has not been empirically evaluated. In particular, the harness and connector integration patterns described in
Sections~\ref{sec:harness-engineering} and~\ref{sec:agentic-connector-architecture}
are reference patterns rather than evaluated implementations.

\paragraph{Performance validity.}
The reported performance measurements are local, in-process microbenchmarks
from three independent runs of 100{,}000 iterations per measured operation.
They exclude model inference, network communication, persistence, external
evidence acquisition, broker or other external-service interaction, database and disk
latency, and other production-system costs. The measurements therefore
characterize the local computational overhead of the evaluated prototype
mechanisms rather than production end-to-end latency, throughput, or
scalability. In addition, independently benchmarked operation means should not
be interpreted as a directly measured end-to-end execution path. The measured
difference between non-authorization-relevant change processing and
authority-material recalculation does not demonstrate a net performance
advantage over alternative continuous-authorization architectures: event
detection, dependency maintenance, evidence acquisition, and matched baseline
implementations were not measured.

\paragraph{Scope of demonstrated properties.}
The empirical claims summarized in Table~\ref{tab:consolidated-results} are
supported by the 48 composed-TGP scenario tests together with the relevant
25 isolated domain-independent ACCA/core-governance tests. The composed-TGP
group includes both core-origin and trading-origin stimuli evaluated under the
active profile. The separate 20 synthetic SGP/SOC tests are excluded from these
empirical claims and are not evidence of an operational security-operations
evaluation or cross-domain empirical generalization. The supported properties
include closed-world declared authority and default deny, adaptation to endogenous
and exogenous changes in the assurance basis, evidence applicability, authorization-relevance
selectivity, applicable-policy mediation, and epoch-based stale-action rejection.
Authority contraction and subsequent recovery are additionally demonstrated in
the S4-U scenario. Progressive bounded expansion remains a model-level property
that is not separately demonstrated by the present experiments.

The S2 model-substitution scenario is a controlled core-origin authority-material
change experiment. It demonstrates authority rederivation after a model identity/version
change, but it is not a continual-learning or self-evolution experiment and does
not empirically establish governance of an online learning process. Continual
learning, evidence portability across endogenous adaptive updates, protected
candidate-state activation, and rollback remain architectural or future-work
capabilities unless separately evaluated.

Other properties discussed in the architecture have a different evidentiary
status. Behavioral-assurance integration and delegation containment are
architecturally specified rather than established here as complete runtime
capabilities. Automated behavioral-failure detection, dynamic delegation
enforcement, evidence acquisition and qualification, distributed authority
propagation, in-flight transaction recovery, complete cross-system forensic
reconstruction, and organizational accountability remain research problems rather
than capabilities demonstrated by the present prototype.

Taken together, these limitations define the evidentiary boundary of the
present work. The experiments support the implemented ACCA authorization
mechanism under the tested TGP configurations and controlled conditions.
They do not establish that governance specifications are complete, that all
relevant changes are observed in time, that assurance evidence is adversarially
trustworthy or representative of deployment behavior, or that strong
stale-effect guarantees extend to untested external providers. These remain
separate validation and systems-research questions, as outlined in
Section~\ref{sec:future-work}.

\section{Research Directions}
\label{sec:future-work}
The limitations identified here motivate three principal research directions:
(i) improving assurance quality through dependency-completeness analysis,
evidence-producer integrity, representativeness, and behavioral evaluation;
(ii) strengthening operational continuity through continual-learning
requalification, protected adaptation activation, and distributed effect-time
enforcement; and (iii) establishing external validity through independently
developed and evaluated domain profiles and controlled comparisons with
competent continuous-authorization baselines. These remain open research
problems rather than capabilities established by the present prototype.

\section{Conclusion}

Autonomous agents can acquire new capabilities, change their behavior, and
operate under evolving conditions while earlier authorizations remain in
flight. Governing these systems therefore requires more than determining
whether an agent possesses credentials or whether a tool call satisfies a
static permission rule. The central question is what degree of consequential
operational authority remains justified under current conditions.

This paper presented \framework{}, a harness-independent-by-design approach to
assurance-conditioned continuous authorization. It explicitly distinguishes
evidence validity from contextual applicability, derives assurance claims
from currently eligible evidence, and resolves a multidimensional
effective-authority envelope under applicable policy within a
principal-declared ceiling. Declared dependencies support selective
reevaluation of authorization-relevant changes, while the authority epoch
advances only when the effective-authority envelope changes. A distinct
optional authorization-basis version can support revalidation of action-specific
conditions without an actor-level authority change. Protected effect-time
enforcement connects these authorization states to consequential execution,
subject to the selected enforcement contract.

A compositional \dgp{} separates core and domain-specific governance
semantics from the authority-derivation mechanism. Different evidence,
actions, requirements, and authority dimensions can thus be interpreted
without granting authority through the profile. Applicable policy and the
principal-declared ceiling remain separately authorized inputs. The
architecture can integrate with existing harnesses, policy engines,
identity systems, assurance sources, and execution adapters rather than
requiring their replacement.

The trading-domain prototype provides implementation-level evidence for
its tested causal mechanisms: declared-authority and default-deny enforcement,
effective-authority adaptation following model substitution or environmental
loss of evidence applicability, policy-dependent authority outcomes, selective
non-response to unrelated changes, contraction and recovery, and rejection
of epoch-stale action authorizations at the local governed-execution gateway
before the simulated-broker boundary. Local microbenchmarks indicate low
in-process computational overhead for the evaluated operations. These results
support selected prototype behavior, not production latency or a general
security guarantee across domains and enforcement environments.

The empirical evaluation is limited to a trading-domain research prototype
with controlled scenarios and selected native platform integration. It does
not establish completeness of authorization dependencies, independent
cross-domain validation, production authorization-basis-version enforcement,
or atomic stale-effect prevention at non-cooperating external providers.
Behavioral-assurance integration, continual-learning requalification,
protected adaptive-state activation, and progressive authority maturation
remain architectural or model-level capabilities needing dedicated evaluation.

The contribution is therefore not a new access-control model, authorization
envelope, policy engine, or fencing primitive in isolation, but their
integration around a continuously maintained assurance-conditioned authority
state. \framework{} extends the continuous-authorization question from
whether an action remains permitted to what portion of an agent's declared
operational authority remains justified by the evidence, conditions, and
policy that currently hold.

\section*{Artifact Availability}

The reference implementation and experimental artifacts supporting this work
are publicly available at
\url{https://github.com/mehdishajari1/ai-hedge-fund}.
The cleaned version used for the reported experiments is archived under the
annotated Git tag \texttt{acca-paper-v1.1}, corresponding to commit
\texttt{aa0846a190c45f6d4b773df661280b9d5b14d63b}. At the frozen revision, the reported governance regression run completed
with 93 tests passing.
The repository includes the ACCA/TGP implementation used for the experiments,
experimental scenarios, regression tests, and governance microbenchmark. It also
temporarily contains the preliminary SGP prototype and 20 synthetic SGP/SOC tests
described in Section~\ref{sec:security-operations}; those artifacts are included for
accessibility and future research and are outside the empirical claims of this
paper, except that the Section~\ref{sec:prototype-overhead} microbenchmark runs through
the SOC adapter as disclosed there. Reproduction commands and the experimental environment are reported in
Section~\ref{sec:experimental-platform}.

\section*{Competing Interests}
The author is the applicant and named inventor on pending Canadian patent applications
3327785 and 3329528 concerning aspects of autonomous-agent governance and authorization.

\section*{Declaration of AI Assistance}
The original concept, research direction, and foundational principles of Assurance-Conditioned
Continuous Authorization (ACCA) were conceived and developed by the author independently of
generative AI assistance. During the subsequent development of the research and preparation of this
manuscript, generative AI tools, including ChatGPT and Claude, assisted with literature exploration,
technical analysis, refinement of mathematical formulations and architectural descriptions, evaluation
of alternative approaches, substantive drafting and editing, and LaTeX preparation. AI-generated
suggestions, analyses, and text were subject to the author's critical review, selection, and revision.
The author exercised independent scientific judgment over the research and assumes full responsibility
for the technical contributions, mathematical statements, experimental results, references, and
conclusions.

\appendix
\section{Trading Governance Profile: Experimental Specification}
\noindent In this appendix, MUST, SHOULD, and MAY state proposed TGP v0.1 profile requirements rather than requirements of an external standard.
\label{app:tgp-spec}
The following profile excerpts illustrate the action vocabulary, evidence assumptions, claim derivation, and selected conservative policy rules used to interpret the trading scenarios. They are not a complete executable reproduction of all reported experiments; scenario-specific fixtures (including the assured-state $\$10{,}000$ policy, the stronger S2 prohibition, and the alternative S7 policy) are not fully represented by this excerpt.

\subsection{Trading Action Vocabulary}

Actions are the principal enforceable privileges in TGP. Let

\begin{equation}
\mathsf{Act}_{\mathrm{TGP}}
=
\mathsf{Act}_I \cup \mathsf{Act}_P \cup \mathsf{Act}_X \cup \mathsf{Act}_G,
\end{equation}

where $\mathsf{Act}_I$ denotes informational actions, $\mathsf{Act}_P$ portfolio-management actions,
$\mathsf{Act}_X$ execution actions, and $\mathsf{Act}_G$ governance or administrative actions.

An initial vocabulary is:

\begin{align}
\mathsf{Act}_I = \{&
\mathrm{observe\mbox{-}market},
\mathrm{read\mbox{-}market\mbox{-}data},
\mathrm{read\mbox{-}fundamentals},
\mathrm{read\mbox{-}news},\notag\\
&
\mathrm{analyze\mbox{-}security},
\mathrm{generate\mbox{-}signal},
\mathrm{generate\mbox{-}recommendation}
\},\\
\mathsf{Act}_P = \{&
\mathrm{construct\mbox{-}portfolio},
\mathrm{rebalance\mbox{-}portfolio},
\mathrm{set\mbox{-}target\mbox{-}weight},\notag\\
&
\mathrm{reduce\mbox{-}position},
\mathrm{close\mbox{-}position}
\},\\
\mathsf{Act}_X = \{&
\mathrm{propose\mbox{-}order},
\mathrm{place\mbox{-}order},
\mathrm{modify\mbox{-}order},
\mathrm{cancel\mbox{-}order}
\},\\
\mathsf{Act}_G = \{&
\mathrm{change\mbox{-}strategy},
\mathrm{change\mbox{-}model},
\mathrm{change\mbox{-}risk\mbox{-}policy},
\mathrm{change\mbox{-}broker},\notag\\
&
\mathrm{modify\mbox{-}credentials},
\mathrm{withdraw\mbox{-}cash},
\mathrm{delegate\mbox{-}authority}
\}.
\end{align}

The vocabulary is extensible. Domain-specific implementations MAY introduce
additional actions, but extensions SHOULD preserve the distinction between
informational, portfolio, execution, and governance operations so that
authority can contract selectively.

The hyphenated identifiers in the equations are conceptual action names.
Machine-readable profile artifacts and implementation code serialize the same
identifiers in snake case; for example,
$\operatorname{serialize}(\mathrm{place\mbox{-}order})=
\texttt{place\_order}$ and
$\operatorname{serialize}(\mathrm{cancel\mbox{-}order})=
\texttt{cancel\_order}$. This representation mapping does not define distinct
actions.

\subsection{Declared Trading Authority}

For principal $u$ and actor $g$, a configured TGP instance represents the principal-declared authority ceiling

\begin{equation}
A_{d,u,g}
=
\langle
\Pi,
R,
S,
V,
L,
D,
H,
E,
T,
O,
\tau
\rangle,
\end{equation}

where:

\begin{itemize}[leftmargin=*]
\item $\Pi:\mathsf{Act}_{\mathrm{TGP}}\rightarrow
\{\mathrm{permit},\mathrm{prohibit},\mathrm{unspecified}\}$ is the declared
action-permission component, with unspecified consequential actions denied by default;
\item $R$ denotes governed resources such as accounts, portfolios,
cash, and financial instruments;
\item $S$ denotes trading scope, including asset classes, symbols,
markets, sectors, accounts, or jurisdictions;
\item $V$ denotes financial and risk limits;
\item $L$ denotes the maximum permitted autonomy level;
\item $D$ denotes delegation authority;
\item $H$ denotes approval requirements;
\item $E$ denotes permitted execution environments;
\item $T$ denotes operating constraints, including trading-session conditions
that constrain authority rather than merely current platform feasibility;
\item $O$ denotes obligations associated with authority; and
\item $\tau$ denotes temporal validity.
\end{itemize}

This ordering and typing match the generic effective-authority envelope.
The action- and environment-specific requirement selector
$\mathrm{Req}_{\mathrm{TGP}}$ is an input to assurance-conditioned authority
derivation, not an additional effective-authority-envelope dimension.

The principal-declared authority ceiling represents the maximum authority that the principal is
willing to declare for the actor. \framework{} derives the currently justified
effective-authority envelope, which is no more permissive than that ceiling and may
expand, contract, or recover as the assurance basis changes, but MUST NOT derive
effective authority outside the principal-declared authority ceiling:

\begin{equation}
A_{\mathrm{eff},u,g}^{(t)}
\preceq_d
A_{d,u,g}.
\label{eq:tgp-authority-containment}
\end{equation}

Expansion beyond $A_{d,u,g}$ requires an explicit change to the principal-declared authority ceiling
rather than merely improved assurance or operating conditions.

\subsection{Trading Scope and Resources}

TGP separates an action from the resources and trading scope over which that
action may be exercised. Authorization for \texttt{place\_order}, for
example, does not imply permission to trade every instrument or use every
account.

Scope predicates may constrain authority by

\begin{equation}
\{
\mathrm{account},
\mathrm{asset\mbox{-}class},
\mathrm{instrument},
\mathrm{symbol},
\mathrm{sector},
\mathrm{market},
\mathrm{jurisdiction}
\}.
\end{equation}

An authority statement may therefore permit equity trading in selected
securities while explicitly prohibiting options or futures:

\begin{equation}
\mathrm{place\mbox{-}order}
\land
\mathrm{assetclass}=\mathrm{equity}
\land
s \in \{\mathrm{NVDA},\mathrm{MSFT},\mathrm{AAPL}\}.
\end{equation}

A request outside this scope MUST be denied even if the model, tool, broker,
and account are technically capable of performing it.

\subsection{Financial and Risk Limits}

TGP represents quantitative limits as dimensions of authority rather than as
properties of the trading strategy. Initial dimensions include

\begin{equation}
V =
\langle
v_o,
p_{\max},
e_g,
e_n,
\ell_d,
n_o
\rangle,
\end{equation}

where $v_o$ is maximum order value, $p_{\max}$ maximum position percentage,
$e_g$ maximum gross exposure, $e_n$ maximum net exposure, $\ell_d$ maximum
permitted daily loss, and $n_o$ maximum number of orders within a defined
period.

These limits MAY differ by action, environment, asset class, account, or
assurance state.

\subsection{Autonomy Levels}

TGP defines autonomy independently from action permission. An actor may be
permitted to participate in an action while being restricted in the degree of
independence with which the action can proceed.

TGP v0.1 defines:

\begin{center}
\begin{tabularx}{0.92\textwidth}{@{}lX@{}}
\toprule
\textbf{Level} & \textbf{Meaning} \\
\midrule
A0 & Observe only; no analytical or consequential action. \\
A1 & Analyze, explain, recommend, or formulate bounded action proposals; the
actor cannot itself cause a consequential effect. Proposal and pending-request
constraints such as value limits and human-approval requirements may still
apply. \\
A2 & Proceed with a consequential action only after required human approval or
equivalent external confirmation; no autonomous consequential execution. \\
A3 & Execute consequential actions autonomously within a bounded authority
envelope. \\
A4 & Execute and delegate bounded authority to subordinate actors. \\
\bottomrule
\end{tabularx}
\end{center}

These levels are specific to TGP v0.1 and are not asserted as universal
autonomy levels. At A1, proposal and pending-request constraints such as value
limits and human-approval requirements may still apply even though the governed
actor cannot itself cause the consequential effect.

\subsection{Approval Requirements}

Let

\begin{equation}
H(a,x,t)
\end{equation}

denote the approval requirement for action $a$ with parameters $x$ at time
$t$. TGP v0.1 defines

\begin{equation}
H \in
\{
\mathrm{none},
\mathrm{human},
\mathrm{dual}
\}.
\end{equation}

For example,

\begin{equation}
H(\mathrm{place\mbox{-}order},v,t)
=
\begin{cases}
\mathrm{none}, & v \leq \$20{,}000,\\
\mathrm{human}, & v > \$20{,}000.
\end{cases}
\end{equation}

An action requiring approval SHOULD be placed into a pending-approval state
rather than interpreted as permanently denied.

\subsection{Execution Environment}
TGP v0.1 recognizes

\begin{equation}
E =
\{
\mathrm{simulation},
\mathrm{paper},
\mathrm{live}
\}.
\end{equation}

The same action MAY receive different limits, assurance requirements, or
approval requirements across environments.

\subsection{Composed TGP State Schema}

TGP does not introduce a second state object alongside the generic state in
Eq.~\ref{eq:profile-relative-state}. Instead, its core and trading schema
contributions compile into one unified runtime schema:
\begin{equation}
\boxed{
\Sigma_X^{\mathrm{TGP}}
=
\Sigma_{X,\mathrm{core}}
\oplus
\Sigma_{X,\mathrm{trading}},
\qquad
X_t^{\mathrm{TGP}}
=
\langle S_t^{\mathrm{TGP}},D_t^{\mathrm{TGP}},O_t^{\mathrm{TGP}},
H_t^{\mathrm{TGP}},B_t^{\mathrm{TGP}}\rangle
\in
\operatorname{Val}\!\left(\Sigma_X^{\mathrm{TGP}}\right).
}
\label{eq:tgp-state-schema}
\end{equation}
Core-origin fields remain part of this state. For example, model type, model
version, model digest, harness version, and the active TGP and applicable-policy
identity, version, and digest are represented in $S_t^{\mathrm{TGP}}$. The trading
layer can add strategy identity and version, risk-control configuration, and
broker-adapter configuration to the same system-state component.

Trading-origin fields are placed according to their semantics rather than in an
undifferentiated context set. The external operating-environment contribution
includes
\begin{equation}
\begin{aligned}
\Sigma_{O,\mathrm{trading}}\supseteq\{&
\mathrm{market\mbox{-}normality},
\mathrm{market\mbox{-}regime},
\mathrm{volatility},
\mathrm{liquidity},\\
&
\mathrm{market/session\mbox{-}status},
\mathrm{observed\mbox{-}spread},
\mathrm{trading\mbox{-}halt},\\
&
\mathrm{venue\mbox{-}circuit\mbox{-}breaker},
\mathrm{news\mbox{-}shock}
\},
\end{aligned}
\label{eq:tgp-operating-schema}
\end{equation}
whereas the dependency-state contribution includes
\begin{equation}
\begin{aligned}
\Sigma_{D,\mathrm{trading}}\supseteq\{&
\mathrm{market\mbox{-}data\mbox{-}freshness},
\mathrm{market\mbox{-}data\mbox{-}quality},\\
&
\mathrm{broker\mbox{-}service\mbox{-}health},
\mathrm{exchange\mbox{-}connectivity/health},
\mathrm{monitoring\mbox{-}health}
\}.
\end{aligned}
\label{eq:tgp-dependency-schema}
\end{equation}
An observed bid--ask spread is therefore operating state, while an assessment
that the spread feed is stale or corrupted is dependency state or evidence about
that state. TGP standardizes the governance meaning and placement of these fields
but does not prescribe the detector or estimator that populates them.

The hyphenated labels above denote conceptual schema fields; namespaced dotted
paths denote their serialized representation in profile artifacts and runtime
adapters. In particular,
\begin{equation}
\begin{aligned}
\mathrm{market\mbox{-}normality}
&\longleftrightarrow
\texttt{trading:market.normality},\\
\mathrm{market/session\mbox{-}status}
&\longleftrightarrow
\texttt{trading:market.status}.
\end{aligned}
\label{eq:tgp-field-serialization}
\end{equation}
The conceptual field $\mathrm{market\mbox{-}data\mbox{-}freshness}$ is
evaluated by comparing a runtime data-age value with a configured maximum-age
threshold. Their serialized keys are
\[
\begin{aligned}
k_{\mathrm{age}}
&=\texttt{trading:market-data.age},\\
k_{\mathrm{max\mbox{-}age}}
&=\texttt{trading:market-data.max\_age\_seconds}.
\end{aligned}
\]
These are representation and configuration mappings, not additional state
fields with competing meanings.

\subsection{TGP Assurance-Claim Vocabulary and Action Requirements}

The TGP claim vocabulary is the derivation closure of core claims and
trading-specific claims:
\begin{equation}
\mathcal{C}_{\mathrm{TGP}}
=
\operatorname{Cl}_{\mathrm{derive}}\!\left(
\mathcal{C}_{\mathrm{core}}
\oplus
\mathcal{C}_{\mathrm{trading}}
\right).
\label{eq:tgp-claim-composition}
\end{equation}
An initial base vocabulary is
\begin{equation}
\begin{aligned}
\mathcal{C}_{\mathrm{core}}=\{&
\mathrm{MODEL\_ASSURANCE},
\mathrm{DATA\_ASSURANCE},\\
&
\mathrm{MONITORING\_HEALTH},
\mathrm{SECURITY\_ASSURANCE},
\mathrm{CONTEXT\_INTEGRITY},\\
&
\mathrm{INSTRUCTION\_CONSISTENCY},\\
&
\mathrm{TASK\_FOCUS},
\mathrm{POLICY\_ADHERENCE},\\
&
\mathrm{DECISION\_STABILITY},
\mathrm{TOOL\_USE\_RELIABILITY},\\
&
\mathrm{DELEGATION\_RELIABILITY}
\},
\\[3pt]
\mathcal{C}_{\mathrm{trading}}=\{&
\mathrm{STRATEGY\_ASSURANCE},
\mathrm{RISK\_CONTROL\_HEALTH},\\
&
\mathrm{BROKER\_HEALTH}
\}.
\end{aligned}
\label{eq:tgp-base-claims}
\end{equation}
The core vocabulary includes the seven behavioral-assurance claims
introduced in Section~\ref{sec:behavioral-assurance}; an action-specific
requirement selector need not require all of them for every action or execution
environment.

Claim vocabulary and action requirements are distinct. For action $a$ and
execution environment $\eta$, TGP selects
\begin{equation}
\mathrm{Req}_{\mathrm{TGP}}(a,\eta)
\subseteq
\mathcal{C}_{\mathrm{TGP}}.
\label{eq:tgp-requirement-selection}
\end{equation}
For example,
\begin{equation}
\begin{aligned}
\mathrm{Req}_{\mathrm{live}}
&:=
\mathrm{Req}_{\mathrm{TGP}}^{\mathrm{base}}
(\mathrm{place\mbox{-}order},\mathrm{live})
\\
&=\{
\mathrm{MODEL\_ASSURANCE},
\mathrm{DATA\_ASSURANCE},\\
&
\mathrm{STRATEGY\_ASSURANCE},
\mathrm{RISK\_CONTROL\_HEALTH},
\mathrm{BROKER\_HEALTH},\\
&
\mathrm{MONITORING\_HEALTH},\\
&
\mathrm{CONTEXT\_INTEGRITY},
\mathrm{POLICY\_ADHERENCE},\\
&
\mathrm{TOOL\_USE\_RELIABILITY}
\}.
\end{aligned}
\label{eq:tgp-live-base-requirements}
\end{equation}
Requirements MAY differ by action, actor, and execution environment.

\texttt{SAFE\_AUTONOMOUS\_TRADING} is a derived composite assurance claim
rather than a peer base claim or a stakeholder-specific authorization
decision. Let $C_{t,\mathrm{base}}$ denote the lower-level base claims supported
from $E_t^*$. A configured derivation may, for example, support the composite
claim only when all selected lower-level claims are supported:
\begin{equation}
\mathrm{SAFE\_AUTONOMOUS\_TRADING}\in C_t
\iff
\mathrm{Req}_{\mathrm{live}}\subseteq C_{t,\mathrm{base}}.
\label{eq:safe-autonomous-trading-derived}
\end{equation}
The mnemonic word \texttt{SAFE} names this profile-level assurance claim; it
does not assert that a particular stakeholder has accepted the residual risk
or authorized autonomous execution. Direct contextual predicates and $P_u$
remain downstream inputs to effective-authority derivation and do not
determine membership in $C_t$. The derivation rule and its dependencies are
profile configuration, not direct evidence that bypasses the lower-level
claims.

\subsection{Evidence Applicability and Trading Assumptions}

Let candidate evidence artifact $e\in E_t$ be associated with applicability
predicate

\begin{equation}
\phi_e(X_t^{\mathrm{TGP}})
\equiv
\mathrm{Applicable}(e,X_t^{\mathrm{TGP}}).
\end{equation}

Evidence is eligible to contribute support to a claim at time $t$ only when it
is both valid and applicable:

\begin{equation}
e\in E_t^*
\iff
\mathrm{Valid}(e,t)
\land
\phi_e(X_t^{\mathrm{TGP}}).
\label{eq:tgp-evidence-eligibility}
\end{equation}

For example, one cross-layer applicability predicate can depend jointly on a
trading-origin operating field, a trading-origin dependency field, and a
core-origin system field:

\begin{equation}
\begin{aligned}
\phi_{e_{17}}(X_t^{\mathrm{TGP}})
={}&
\bigl(
O_t^{\mathrm{TGP}}[\texttt{trading:market.normality}]
\in
\{\mathrm{normal},\mathrm{elevated}\}
\bigr)
\\[-1pt]
&{}\land
\bigl(
D_t^{\mathrm{TGP}}[\texttt{trading:market-data.age}]
\leq30\,\mathrm{s}
\bigr)
\\[-1pt]
&{}\land
\bigl(
S_t^{\mathrm{TGP}}[\texttt{core:model.version}]
=M_1
\bigr).
\end{aligned}
\label{eq:tgp-cross-layer-applicability}
\end{equation}

Violation of an evidence assumption does not itself prescribe a universal
authorization response. Instead,

\begin{equation}
\Delta X_t
\leadsto
\Delta E_t^*
\leadsto
\Delta C_t
\xrightarrow{\,P_u\,}
\Delta A_{\mathrm{eff},u,g}^{(t)}.
\end{equation}

\subsection{Direct Contextual Policy}

Not every contextual constraint must be mediated by assurance evidence.

For example,

\begin{equation}
O_t^{\mathrm{TGP}}[\texttt{trading:market.status}]=\mathrm{halted}
\Rightarrow
\mathrm{place\mbox{-}order}
\notin
\operatorname{Allow}\!\left(A_{\mathrm{eff},u,g}^{(t)}\right).
\label{eq:tgp-direct-path}
\end{equation}

TGP therefore distinguishes

\begin{align}
\bigl(X_t^{\mathrm{TGP}},E_t\bigr)
&\rightarrow
\mathrm{valid/applicable\ evidence}\ E_t^*
\rightarrow
\mathrm{claims}\ C_t
\rightarrow
\mathrm{policy}\ P_u
\rightarrow
\mathrm{authority}\ A_{\mathrm{eff},u,g}^{(t)},
\\
X_t^{\mathrm{TGP}}
&\rightarrow
\mathrm{direct\ policy}\ P_u
\rightarrow
\mathrm{authority}\ A_{\mathrm{eff},u,g}^{(t)}.
\end{align}

\subsection{Effective Trading Authority}

At deployment, the configured TGP instance supplies the validated composed
semantics, state schema, claim vocabulary, requirement-selection rules, dependency
relations, and authorization-relevance and authority-material-change rules. Runtime sources populate the unified
$X_t^{\mathrm{TGP}}$ and candidate evidence set $E_t$. The principal or another
authorized governance process separately supplies the principal-declared authority ceiling $A_{d,u,g}$ and applicable
policy $P_u$ expressed over those semantics; TGP does not itself declare the ceiling or allocate effective authority.
Effective trading authority is therefore the TGP instantiation of
Eq.~\ref{eq:effective-authority-derivation}:

\begin{equation}
A_{\mathrm{eff},u,g}^{(t)}
=
F(
A_{d,u,g},
X_t^{\mathrm{TGP}},
E_t^*,
C_t,
P_u
).
\end{equation}

A runtime action request is

\begin{equation}
q =
\langle
g,a,r,x,\eta
\rangle,
\end{equation}

where $g$ is actor, $a$ action, $r$ resource, $x$ material action parameters,
and $\eta$ execution environment. Let $\alpha_q$ denote the corresponding
action-authorization artifact, which binds the request to the authority epoch
$\epsilon_q$ under which it was issued and to any additional version values required for that action or
enforcement path. We write $\mathrm{CurrentVersions}(\alpha_q,t)$ to mean that
each version value whose currentness is required for the action matches or
otherwise satisfies the corresponding authoritative current value. This always
includes $\epsilon_q=\mathrm{Epoch}_t$ and, when configured, includes
$\beta_q=\mathrm{BasisVer}_t$ for a carried or referenced authorization-basis
version $\beta_q$, together with any other required version-vector components.

Formally,

\begin{equation}
\begin{aligned}
\mathrm{PermitEffect}(q,\alpha_q,t)\Rightarrow{}&
\mathrm{IssuedPermit}(q,\alpha_q)\\
&\land\;\mathrm{CurrentVersions}(\alpha_q,t)\\
&\land\;\mathrm{ValidArtifact}(\alpha_q,q,t)\\
&\land\;\mathrm{RequiredConditionsHold}(\alpha_q,t).
\end{aligned}
\end{equation}
Here $\mathrm{IssuedPermit}$ records a prior Permit decision for the bound request;
$\mathrm{ValidArtifact}$ denotes the integrity, expiry, replay, and material-parameter checks required by the artifact;
and $\mathrm{RequiredConditionsHold}$ denotes any conditions whose effect-time satisfaction is required by the enforcement contract.
The PEP need not rerun the full action-decision function at time $t$.
This implication states an intended contract for a protected effect boundary,
not a property experimentally established for external providers. Any
post-issuance change capable of invalidating a required authorization
condition must be covered by an effective version transition, effect-time
condition check, or another explicitly defined protection; untracked changes
are not detected merely by comparing epochs.

\subsection{Continuous Action Authorization}

TGP treats consequential actions as the primary runtime enforcement events.
Continuous authorization therefore does not require every token or internal
reasoning step of an LLM to be interpreted as an access-control event.

Continuity is maintained in two complementary ways:

\begin{enumerate}[leftmargin=*]
\item effective authority is reevaluated when an authorization-relevant change
affects its state, evidence, applicable-policy, or declared-authority inputs; and
\item consequential external actions are evaluated against the current effective
authority and any additional required authorization-state version values that
hold at the relevant enforcement point.
\end{enumerate}

Thus,

\begin{equation}
A_{\mathrm{eff},u,g}^{(t_0)}
\neq
A_{\mathrm{eff},u,g}^{(t_1)}
\end{equation}

may hold even when the underlying model itself is unchanged.

\subsection{Example TGP Profile}

\begingroup
\small
\begin{verbatim}
tgp_version: "0.1"

profile:
  id: conservative-equity-001
  owner: trader-001
  composition:
    core_layer: core-governance-0.1
    domain_layer: trading-0.1

actor:
  id: agent:portfolio-manager
  role: portfolio_manager

authority:
  actions:
    allowed:
      - analyze_security
      - generate_recommendation
      - construct_portfolio
      - propose_order
      - place_order
      - cancel_order

    prohibited:
      - withdraw_cash
      - change_risk_policy
      - modify_credentials

  scope:
    asset_classes:
      allowed: [equity]
    symbols:
      allowed: [NVDA, MSFT, AAPL]

  limits:
    max_order_value: 30000
    max_position_pct: 0.25
    max_gross_exposure: 1.0
    max_daily_loss_pct: 0.03

  autonomy:
    max_level: A3

  environment:
    allowed: [simulation, paper, live]

  approval:
    - when:
        order_value_gt: 20000
      require: human

  delegation:
    allowed: true
    allowed_roles: [execution]
    max_depth: 1

assurance:
  place_order:
    required:
      - MODEL_ASSURANCE
      - DATA_ASSURANCE
      - STRATEGY_ASSURANCE
      - RISK_CONTROL_HEALTH
      - BROKER_HEALTH
      - MONITORING_HEALTH
      - CONTEXT_INTEGRITY
      - POLICY_ADHERENCE
      - TOOL_USE_RELIABILITY

  live_execution:
    required:
      - SAFE_AUTONOMOUS_TRADING

claims:
  derived:
    SAFE_AUTONOMOUS_TRADING:
      all_of:
        - MODEL_ASSURANCE
        - DATA_ASSURANCE
        - STRATEGY_ASSURANCE
        - RISK_CONTROL_HEALTH
        - BROKER_HEALTH
        - MONITORING_HEALTH
        - CONTEXT_INTEGRITY
        - POLICY_ADHERENCE
        - TOOL_USE_RELIABILITY

evidence:
  E-17:
    supports:
      - MODEL_ASSURANCE
      - STRATEGY_ASSURANCE

    assumptions:
      trading:market.normality: [normal, elevated]
      core:model.version: [M1]
      trading:market-data.max_age_seconds: 30

adaptation:
  - when:
      claim:
        SAFE_AUTONOMOUS_TRADING: unassured
    derive:
      max_autonomy: A1
      max_order_value: 1000
      human_approval: human

  - when:
      claim:
        DATA_ASSURANCE: degraded
    derive:
      live_execution: false
      paper_execution: true

  - when:
      claim:
        CONTEXT_INTEGRITY: degraded
    derive:
      live_execution: false
      max_autonomy: A1
      human_approval: human

  - when:
      context:
        trading:market.status: halted
    derive:
      place_order: false
\end{verbatim}
\endgroup

For compactness, this example places three separately governed inputs in one file: the TGP composition reference, the principal-declared authority ceiling (\texttt{authority}), and applicable-policy rules (\texttt{assurance} requirements and \texttt{adaptation} responses). In a deployment these are separately authorized and versioned inputs; the \texttt{adaptation} rules are part of $P_u$, not profile semantics. The \texttt{assurance.place\_order.required} list illustrates the base claims associated with live order placement, while \texttt{assurance.live\_execution.required} represents an additional composite prerequisite for live execution; these entries should not be interpreted as identical unconditional gates for paper and live actions. The policy excerpt does not encode the complete assured-state policy, the S2 prohibition, or the S7 alternative. Accordingly, the equations in Section~6 are illustrative policy outcomes rather than results that can all be derived from this shortened YAML alone. Because autonomy level and delegation rights are separate authority dimensions, the sample delegation entry does not imply that the agent can exercise A4 delegation autonomy; any actual delegated execution remains subject to its own authorization and applicable limits. The underlying TGP specification defines the vocabulary and constraints for these declarations but does not itself declare the ceiling, supply applicable policy, or allocate effective authority. \framework{} derives the no-more-permissive effective-authority envelope that remains justified under the current assurance, behavioral, policy, and operating state.

\bibliographystyle{plainurl}
\bibliography{references_paper1_v55_final}

\end{document}

%% file: preamble_add.tex
\usepackage{tikz}
\usetikzlibrary{arrows.meta,positioning,fit,backgrounds,calc}
\tikzset{
  acca/box/.style={draw, rounded corners=2pt, align=center, font=\footnotesize,
                   inner sep=3pt, minimum height=8mm, fill=white,
                   execute at begin node={\hyphenpenalty=10000\exhyphenpenalty=10000}},
  acca/core/.style={acca/box, fill=gray!12, thick},
  acca/input/.style={acca/box, fill=white, dashed},
  acca/arr/.style={-{Stealth[length=2.2mm]}, semithick},
  acca/darr/.style={-{Stealth[length=2.2mm]}, semithick, dashed},
  acca/lbl/.style={font=\scriptsize, align=center, inner sep=1pt},
  acca/group/.style={draw, rounded corners=4pt, inner sep=6pt, gray!70, thick}
}

%% file: fig_refs.tex
\providecommand{\figEqLifecycle}{\ref{eq:canonical-lifecycle}}
\providecommand{\figEqEligible}{\ref{eq:eligible-evidence-set}}
\providecommand{\figEqDerive}{\ref{eq:effective-authority-derivation}}
\providecommand{\figEqAuthorize}{\ref{eq:action-authorization-basis}}
\providecommand{\figEqBasisVer}{\ref{eq:authority-basis-versioning-cases}}
\providecommand{\figEqEpoch}{\ref{eq:epoch-transition-rule}}
\providecommand{\figEqDeny}{\ref{eq:stale-epoch-deny}}
\providecommand{\figEqDirect}{\ref{eq:tgp-direct-path}}
\providecommand{\figSecDepGraph}{\ref{sec:dependency-graph}}

\providecommand{\figSecTrust}{\ref{sec:governance-plane-integrity}}
\providecommand{\figTabContracts}{\ref{tab:enforcement-strength}}

%% file: fig1_lifecycle.tex
\begin{figure}[t]
\centering
\resizebox{\textwidth}{!}{%
\begin{tikzpicture}[x=1cm,y=1cm]
  \node[acca/box, text width=34mm] (Ec) at (0,0.8)  {Candidate evidence $E_t$\\{\scriptsize from evidence producers}};
  \node[acca/box, text width=34mm] (X)  at (0,-0.8) {Authority-relevant\\state $X_t^d=\langle S,D,O,H,B\rangle$};
  \node[acca/box, text width=29mm] (G)  at (4.3,0) {Eligibility (Eq.~\figEqEligible)\\$\mathrm{Valid}(e,t)\wedge\phi_e(X_t^d)$};
  \node[acca/box, text width=19mm] (Es) at (7.65,0) {Eligible\\evidence $E_t^{*}$};
  \node[acca/box, text width=18mm] (C)  at (10.95,0) {Assurance\\claims $C_t$};
  \node[acca/core, text width=24mm] (F) at (13.85,0) {Authority resolver\\$F(\cdot)$, Eq.~(\figEqDerive)};

  \node[acca/input, text width=18mm] (P) at (12.8,2.0) {Applicable\\policy $P_u$};
  \node[acca/input, text width=18mm] (A) at (14.95,2.0)  {Declared\\ceiling $A_{d,u,g}$};

  \draw[acca/arr] (Ec.east) -- ($(G.west)+(0,0.15)$);
  \draw[acca/arr] (X.east)  -- ($(G.west)+(0,-0.15)$);
  \draw[acca/arr] (G) -- (Es);
  \draw[acca/arr] (Es) -- node[acca/lbl, above]{Support} (C);
  \draw[acca/arr] (C) -- (F);
  \draw[acca/arr] (P.south) -- (P.south |- F.north);
  \draw[acca/arr] (A.south) -- (A.south |- F.north);
  \draw[acca/darr] (X.south) -- ++(0,-0.6) -| node[acca/lbl, pos=0.25, below]{direct policy predicates on $X_t^d$ (e.g., market halted), Eq.~(\figEqDirect): no evidence or claims involved} ($(F.south)+(-0.7,0)$);

  \begin{scope}[on background layer]
    \node[fit=(Ec)(X)(G)(Es)(C), draw=gray!65, dashed, rounded corners=4pt, inner sep=2.5mm,
          label={[font=\scriptsize, text=black!70, anchor=south west, inner sep=1pt]north west:%
          $\mathrm{DGP}_d=\mathcal{G}_{\mathrm{core}}\oplus\mathcal{L}_d$: state schema, evidence classes, $\phi_e$, claim vocabulary, dependencies $G_A$}] (dgp) {};
  \end{scope}

  \node[acca/core, text width=30mm] (Aeff) at (13.85,-3.4) {Effective authority\\$A^{(t)}_{\mathrm{eff},u,g}\preceq_d A_{d,u,g}$\\{\scriptsize (multidimensional envelope)}};
  \node[acca/core, text width=48mm] (Ep) at (8.7,-3.4) {$\mathrm{Epoch}_t$: advances iff\\the envelope changes (Eq.~\figEqEpoch);\\optional $\mathrm{BasisVer}_t$ (Eq.~\figEqBasisVer)};
  \draw[acca/arr] ($(F.south)+(0.6,0)$) -- ($(Aeff.north)+(0.6,0)$);
  \draw[acca/arr] (Aeff) -- (Ep);

  \node[acca/box, text width=29mm] (Q) at (-0.95,-6.0) {Request $q$\\{\scriptsize proposed by the agent}};
  \node[acca/box, text width=40mm] (Art) at (3.95,-6.0) {Action-level decision (Eq.~\figEqAuthorize)\\{\scriptsize $F_{\mathrm{auth}}\big(q,A^{(t)}_{\mathrm{eff}},\pi_q(\mathcal{B}^{\mathrm{auth}}_t)\big)$}\\{\scriptsize if $\mathsf{Permit}$: issue $\alpha_q$ binding $q$,}\\{\scriptsize digest of $x_q$, $\epsilon_q$, any required $\beta_q$}};
  \node[acca/core, text width=48mm] (PEP) at (8.7,-6.0) {PEP at the protected\\effect boundary (Eq.~\figEqDeny)\\{\scriptsize checks $\epsilon_q=\mathrm{Epoch}_t$ and any required}\\{\scriptsize basis or version-vector components;}\\{\scriptsize parameter digest, expiry, approvals}};
  \node[acca/box, text width=33mm] (Cm) at (14.95,-5.3) {Commit via selected\\{\scriptsize enforcement contract (Table~\figTabContracts)}};
  \node[acca/box, text width=33mm] (Dn) at (14.95,-6.7) {Deny; complete\\reauthorization};

  \draw[acca/arr] (Q) -- (Art);
  \draw[acca/arr] (Art) -- (PEP);
  \draw[acca/arr] (Ep.west) -| node[acca/lbl, pos=0.72, left, align=right]{current $A^{(t)}_{\mathrm{eff}}$, $\mathrm{Epoch}_t$\\and basis $\pi_q(\mathcal{B}^{\mathrm{auth}}_t)$} (Art.north);
  \draw[acca/arr] (Ep.south) -- node[acca/lbl, right]{current values at effect time} (PEP.north);
  \draw[acca/arr] ($(PEP.east)+(0,0.3)$) -- ++(0.2,0) |- node[acca/lbl, pos=0.72, above]{all current} (Cm.west);
  \draw[acca/arr] ($(PEP.east)+(0,-0.3)$) -- ++(0.2,0) |- node[acca/lbl, pos=0.72, below]{any stale} (Dn.west);
\end{tikzpicture}}
\caption{The assurance-to-authority lifecycle. Solid arrows in the upper row show the evidence-mediated path of Eq.~(\figEqLifecycle): only evidence that is both valid and applicable to the current state $X_t^d$ supports assurance claims, and the authority resolver maps those claims through applicable policy $P_u$ to an effective-authority envelope bounded by the principal-declared ceiling. The dashed arrow is the direct path (Eq.~\figEqDirect, Appendix~A.11), which bypasses evidence and claims. The authority epoch advances only when the envelope changes. The action-level decision is obtained through the PEP interaction shown in Figure~2; the boxes depict logical functions rather than a mandated message sequence. A request $q$ proposed by the agent receives an action authorization $\alpha_q$ only if the action-level decision of Eq.~(\figEqAuthorize), taken against the current envelope and the action-specific authorization basis, returns $\mathsf{Permit}$; $\alpha_q$ binds the epoch and any required basis version under which it was issued, and the policy enforcement point (PEP) at the protected effect boundary rejects it when the epoch or any other required version value is no longer current. Passing these checks permits the effect only through the commit mechanism of the selected enforcement contract; the strength of the stale-effect guarantee depends on that contract.}
\label{fig:lifecycle}
\end{figure}
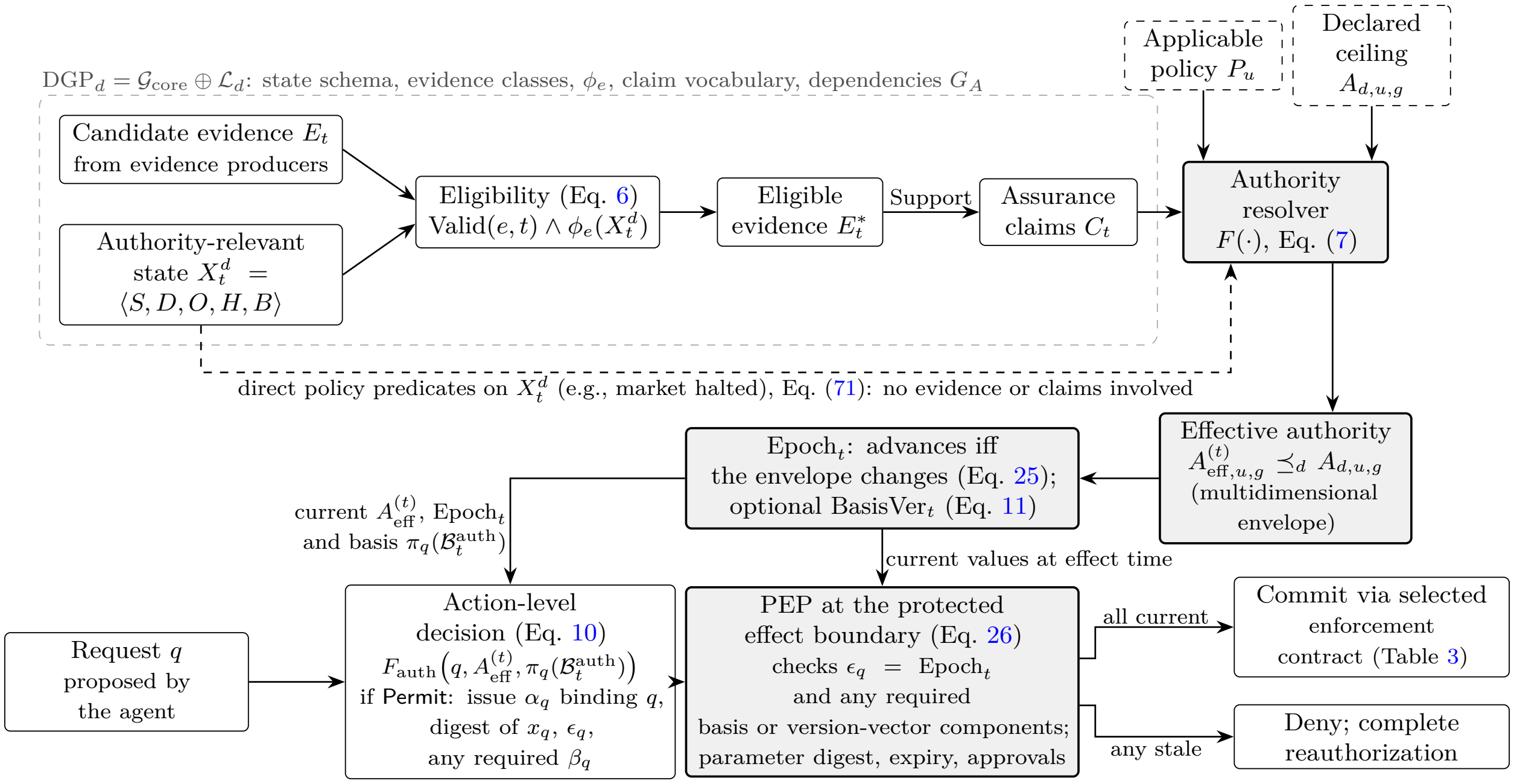

%% file: fig2_architecture.tex
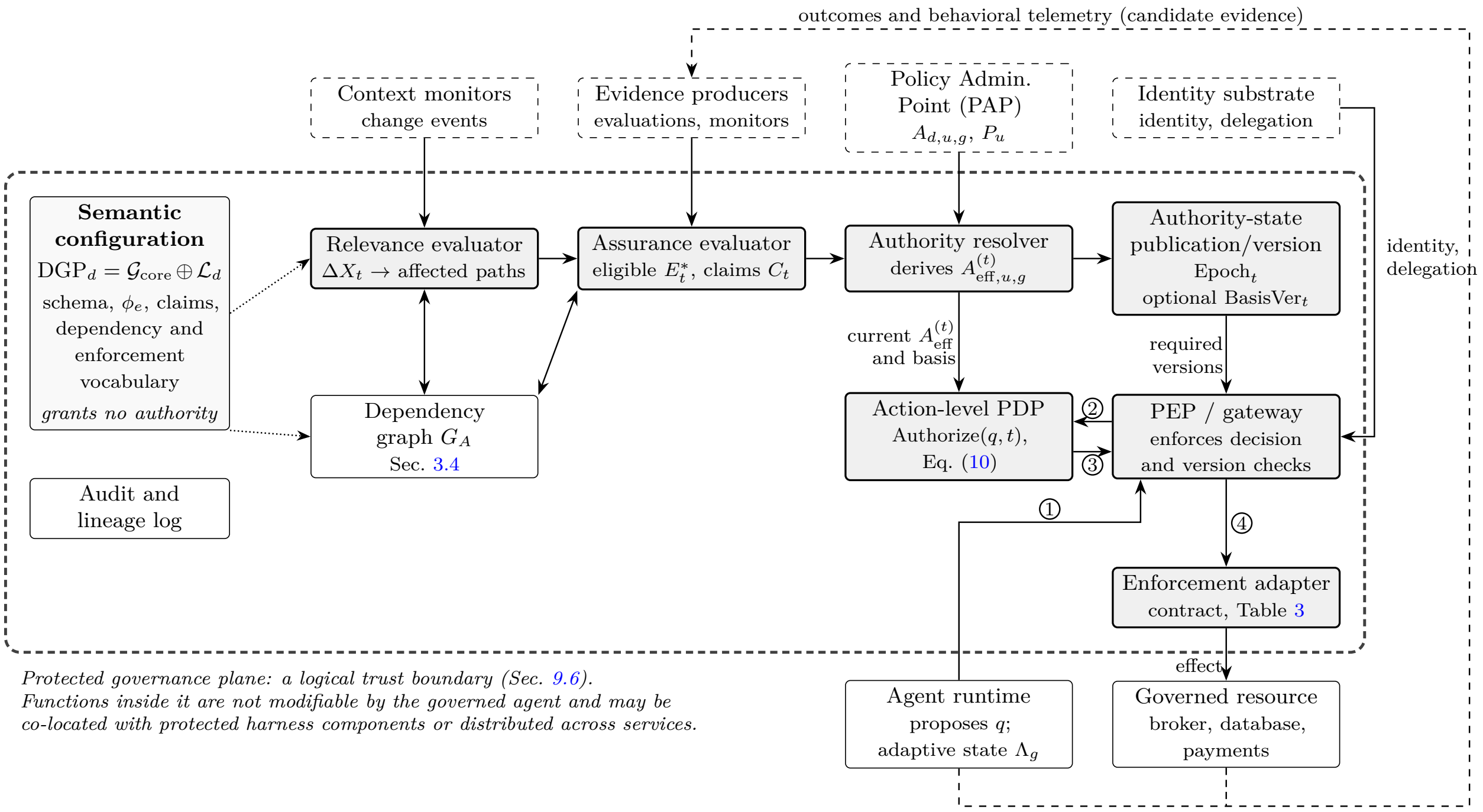
\begin{figure}[p]
\centering
\resizebox{\textwidth}{!}{%
\begin{tikzpicture}[x=1cm,y=1cm]
  \def\cA{3.85}\def\cB{7.75}\def\cC{11.65}\def\cD{15.55}
  \node[acca/input, text width=31mm] (Mon) at (\cA,7.0) {Context monitors\\{\scriptsize change events}};
  \node[acca/input, text width=31mm] (AEP) at (\cB,7.0) {Evidence producers\\{\scriptsize evaluations, monitors}};
  \node[acca/input, text width=31mm] (PAP) at (\cC,7.0) {Policy Admin. Point (PAP)\\{\scriptsize $A_{d,u,g}$, $P_u$}};
  \node[acca/input, text width=31mm] (IdP) at (\cD,7.0) {Identity substrate\\{\scriptsize identity, delegation}};

  \node[draw, rounded corners=2pt, fill=gray!5, align=center, font=\footnotesize, text width=27mm, minimum height=33mm, inner sep=3pt] (DGP) at (-0.45,4.0)
     {\textbf{Semantic configuration}\\[2pt]$\mathrm{DGP}_d=\mathcal{G}_{\mathrm{core}}\oplus\mathcal{L}_d$\\[2pt]{\scriptsize schema, $\phi_e$, claims, dependency and enforcement vocabulary}\\[2pt]{\scriptsize\itshape grants no authority}};
  \node[acca/box, text width=27mm] (Log) at (-0.45,1.15) {Audit and\\lineage log};

  \node[acca/core, text width=31mm] (Rel) at (\cA,4.8) {Relevance evaluator\\{\scriptsize $\Delta X_t\to$ affected paths}};
  \node[acca/core, text width=31mm] (Ass) at (\cB,4.8) {Assurance evaluator\\{\scriptsize eligible $E_t^{*}$, claims $C_t$}};
  \node[acca/core, text width=31mm] (Res) at (\cC,4.8) {Authority resolver\\{\scriptsize derives $A^{(t)}_{\mathrm{eff},u,g}$}};
  \node[acca/core, text width=31mm] (Eps) at (\cD,4.8) {Authority-state\\publication/version\\{\scriptsize $\mathrm{Epoch}_t$}\\{\scriptsize optional $\mathrm{BasisVer}_t$}};

  \node[acca/box,  text width=31mm] (GA)  at (\cA,2.2) {Dependency graph $G_A$\\{\scriptsize Sec.~\figSecDepGraph}};
  \node[acca/core, text width=31mm] (PDP) at (\cC,2.2) {Action-level PDP\\{\scriptsize $\mathrm{Authorize}(q,t)$, Eq.~(\figEqAuthorize)}};
  \node[acca/core, text width=31mm] (PEP) at (\cD,2.2) {PEP / gateway\\{\scriptsize enforces decision}\\{\scriptsize and version checks}};
  \node[acca/core, text width=31mm] (Ad)  at (\cD,-0.15) {Enforcement adapter\\{\scriptsize contract, Table~\figTabContracts}};

  \draw[acca/arr] (Mon) -- (Rel);
  \draw[acca/arr] (AEP) -- (Ass);
  \draw[acca/arr] (PAP) -- (Res);
  \draw[acca/arr] (IdP.east) -- ++(0.5,0) |- (PEP.east);
  \node[acca/lbl, anchor=west, align=left] at ($(\cD,4.8)+(2.3,0)$) {identity,\\delegation};

  \draw[acca/arr] (Rel) -- (Ass);
  \draw[acca/arr] (Ass) -- (Res);
  \draw[acca/arr] (Res) -- (Eps);
  \draw[{Stealth[length=2.2mm]}-{Stealth[length=2.2mm]}, semithick] (GA) -- (Rel);
  \draw[{Stealth[length=2.2mm]}-{Stealth[length=2.2mm]}, semithick] (GA.north east) -- (Ass.south west);
  \draw[-{Stealth[length=2mm]}, densely dotted, semithick] (DGP.east) -- (Rel.west);
  \draw[-{Stealth[length=2mm]}, densely dotted, semithick] (DGP.south east) -- (GA.west);

  \draw[acca/arr] (Res) -- node[acca/lbl, left, align=right]{current $A^{(t)}_{\mathrm{eff}}$\\and basis} (PDP);
  \draw[acca/arr] (Eps) -- node[acca/lbl, left, align=right]{required\\versions} (PEP);
  \draw[acca/arr] ($(PEP.west)+(0,0.22)$) -- node[circle, draw, inner sep=0.6pt, font=\scriptsize, fill=white, above=1pt]{2} ($(PDP.east)+(0,0.22)$);
  \draw[acca/arr] ($(PDP.east)+(0,-0.22)$) -- node[circle, draw, inner sep=0.6pt, font=\scriptsize, fill=white, below=1pt]{3} ($(PEP.west)+(0,-0.22)$);
  \draw[acca/arr] (PEP) -- node[circle, draw, inner sep=0.6pt, font=\scriptsize, fill=white, right=2pt]{4} (Ad);

  \begin{scope}[on background layer]
    \node[fit=(DGP)(Log)(Rel)(Eps)(GA)(Ad), draw=black!75, densely dashed, very thick, rounded corners=6pt, inner sep=3.5mm] (tb) {};
  \end{scope}
  \node[font=\scriptsize\itshape, anchor=north west, align=left] at ($(tb.south west)+(0.1,-0.1)$)
     {Protected governance plane: a logical trust boundary (Sec.~\figSecTrust).\\Functions inside it are not modifiable by the governed agent and may be\\co-located with protected harness components or distributed across services.};

  \node[acca/box, text width=31mm] (Agt) at (\cC,-2.0) {Agent runtime\\{\scriptsize proposes $q$; adaptive state $\Lambda_g$}};
  \node[acca/box, text width=31mm] (Rsc) at (\cD,-2.0) {Governed resource\\{\scriptsize broker, database, payments}};
  \draw[acca/arr] (Agt.north) |- ($(\cD,0.95)+(-1.25,0)$) node[pos=0.75, above=1pt, circle, draw, inner sep=0.6pt, font=\scriptsize, fill=white]{1} -- ($(PEP.south)+(-1.25,0)$);
  \draw[acca/arr] (Ad.south) -- node[acca/lbl, left, pos=0.7]{effect} (Rsc.north);

  \draw[acca/darr] (Agt.south) -- ++(0,-0.55) -| ($(\cD,8.15)+(3.55,0)$) -| (AEP.north);
  \draw[densely dashed, semithick] (Rsc.south) -- ++(0,-0.55);
  \node[acca/lbl, anchor=south] at (13.0,8.15) {outcomes and behavioral telemetry (candidate evidence)};
\end{tikzpicture}}
\caption{Logical reference architecture. External sources enter the components that consume them: change events to the relevance evaluator, evidence to the assurance evaluator, PAP-administered principal-declared authority and applicable policy to the authority resolver, and authenticated identity and delegation context to the enforcement path (and, where relevant, as authority-relevant state). The validated Domain Governance Profile is a semantic configuration layer: it supplies schemas, applicability conditions, claim vocabulary and dependency semantics to the indicated control-plane functions (dotted arrows); other functions consume the same validated semantics as needed. The DGP grants no authority and does not itself supply runtime evidence. The authority resolver derives effective authority; a separate action-level policy decision point (PDP) decides a particular request against that authority and the action-specific basis (Eq.~\figEqAuthorize); the policy enforcement point (PEP) enforces it. The numbered arrows give the action sequence: (1)~the agent submits request $q$ to the PEP; (2)~the PEP sends an authorization request to the PDP; (3)~the PDP returns its decision; (4)~if the decision permits and the required version checks pass, the PEP forwards the operation to the enforcement adapter, whose enforcement contract determines how far the stale-effect guarantee extends. PDP and PEP may be physically integrated. The dashed boundary is a logical trust boundary, not a required deployment. MCP-gateway and isolation integrations are architectural and not part of the reported evaluation.}
\label{fig:architecture}
\end{figure}

%% file: fig3_epoch_timeline.tex
\begin{figure}[p]
\centering
\resizebox{\textwidth}{!}{%
\begin{tikzpicture}[x=1cm,y=1.2cm,
    life/.style={densely dashed, gray!70},
    msg/.style={-{Stealth[length=2.2mm]}, semithick},
    act/.style={draw, fill=gray!12, rounded corners=1.5pt, align=center, font=\scriptsize, inner sep=2.5pt}]
  \foreach \x/\n/\t in {0/ag/{Agent},3/mo/{Context\\monitor},6/ac/{ACCA control\\plane},12/pe/{Execution\\gateway (PEP)},15/br/{Simulated broker\\(effect boundary)}}{
    \node[acca/box, text width=26mm, minimum height=13mm] (h\n) at (\x,0) {\t};
    \draw[life] (h\n.south) -- (\x,-9.3);
  }
  \node[acca/input, text width=26mm, minimum height=13mm] (hqu) at (9,0) {Queue / worker\\{\scriptsize\itshape illustrative only}};
  \draw[life, dotted] (hqu.south) -- (9,-9.3);
  \node[font=\scriptsize\itshape, anchor=east] at (-1.6,-1.2) {$t_0$};
  \node[font=\scriptsize\itshape, anchor=east] at (-1.6,-3.8) {$t_1$};
  \node[font=\scriptsize\itshape, anchor=east] at (-1.6,-6.45) {$t_2$};

  \draw[msg] (0,-1.2) -- node[acca/lbl, above]{proposes $q$ (\texttt{place\_order})} (6,-1.2);
  \node[act, text width=36mm] at (6,-1.9) {$\mathrm{Authorize}(q,t_0)=\mathsf{Permit}$\\under $\mathrm{Epoch}=1$};
  \draw[msg, densely dotted] (6,-2.7) -- (9,-2.7);
  \node[acca/lbl, fill=white, inner sep=1pt, anchor=south] at (9.65,-2.54) {illustrative queue: $\alpha_q$, $\epsilon_q=1$};
  \node[act, text width=24mm, densely dotted, fill=white] at (9,-3.2) {$\alpha_q$ waits\\(queued, delayed)};

  \draw[msg] (3,-3.8) -- node[acca/lbl, above, pos=0.5, yshift=1.5mm, fill=white, inner sep=1pt]{regime shift} (6,-3.8);
  \node[act, text width=58mm] at (6,-4.75) {$e_{17}$ still valid, no longer applicable;\\claim unsupported; $A^{(t)}_{\mathrm{eff}}$ contracts;\\$\mathrm{Epoch}$: $1\to 2$ (Eq.~\figEqEpoch)};
  \draw[msg] (6,-5.75) -- node[acca/lbl, above]{publishes $\mathrm{Epoch}=2$} (12,-5.75);

  \draw[msg, densely dotted] (9,-6.45) -- node[acca/lbl, above]{presents $\alpha_q$ ($\epsilon_q=1$)} (12,-6.45);
  \node[act, text width=34mm] at (12,-7.25) {$\epsilon_q=1\neq\mathrm{Epoch}=2$\\$\Rightarrow\mathrm{Deny}(q)$ (Eq.~\figEqDeny)\\\texttt{AuthorityPreempted}};
  \draw[semithick, dashed] (12,-8.15) -- (14.7,-8.15);
  \node[font=\small] at (14.75,-8.15) {$\times$};
  \node[acca/lbl, below, yshift=-2mm, fill=white, inner sep=1.5pt] at (13.4,-8.15) {no broker call};
  \draw[msg, dashed] (12,-8.85) -- node[acca/lbl, above, pos=0.5, fill=white, inner sep=1pt]{reauthorize at epoch 2} (6,-8.85);

\end{tikzpicture}}
\caption{Authority-epoch stale-authorization rejection (S10).
An authorization issued under epoch~1 is presented after an assurance-driven
authority contraction advances the epoch to~2. The local governed-execution
gateway rejects the stale authorization before invoking the simulated broker.
The dotted queue path is illustrative and was not exercised: the S10 test
presented the stale authorization directly to the gateway after the epoch
transition. Any subsequent reauthorization under epoch~2 occurs only if applicable
policy permits reconsideration. The result does not establish provider-side
fencing or atomic external commitment.}
\label{fig:epoch-timeline}
\end{figure}
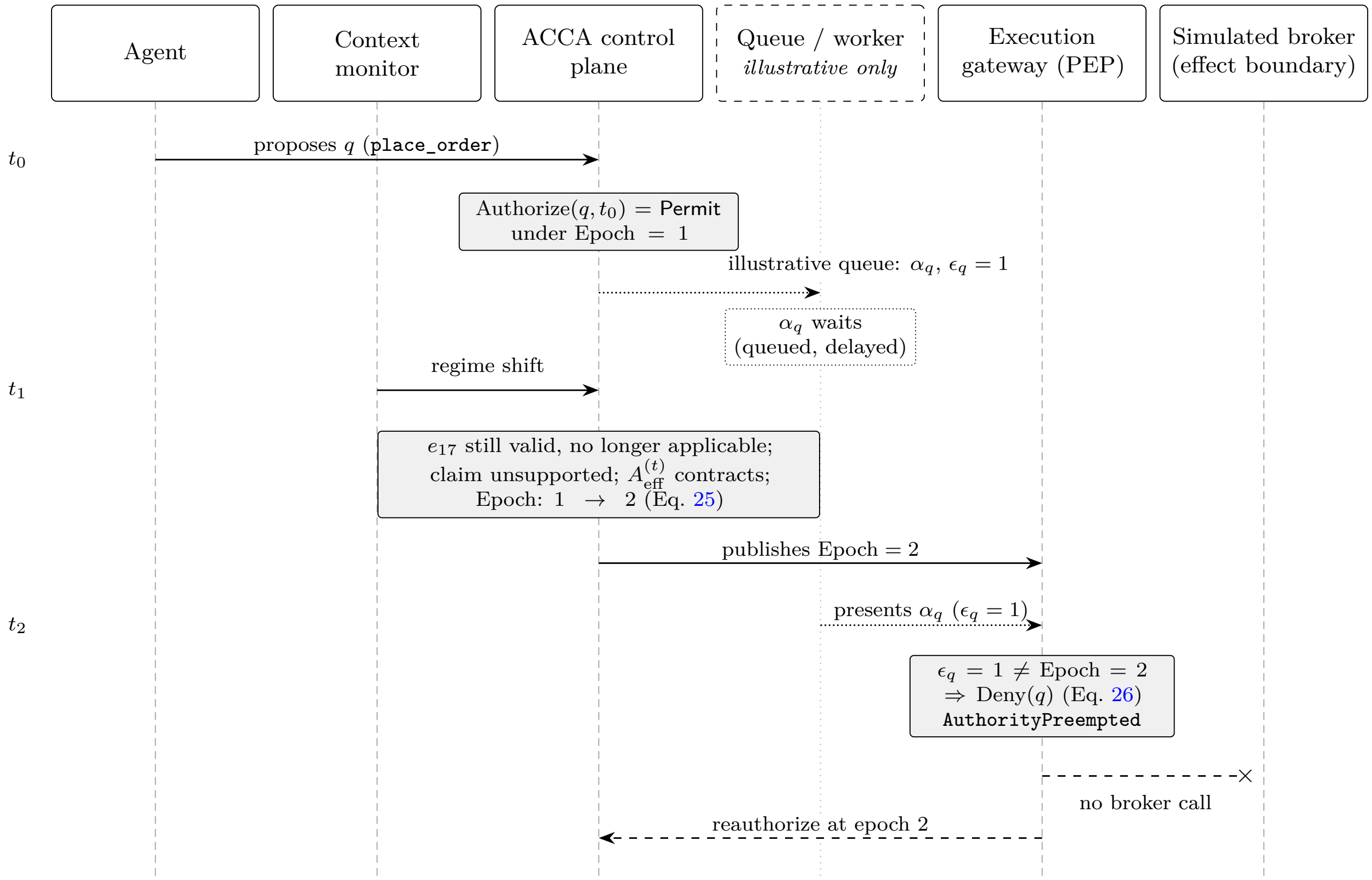